\documentclass{article}

\usepackage{arxiv}
\usepackage[dvipsnames]{xcolor}
\usepackage{annotate-eq}
\usepackage{pifont}
\usepackage{bm}

\usepackage{boxedminipage}
\usepackage{arydshln}
\usepackage{bbm}
\usepackage[flushleft]{threeparttable}

\usepackage{caption}
\usepackage{subcaption}
\usepackage{enumitem}
\usepackage{tabularx}
\usepackage{booktabs}
\usepackage{multirow}
\usepackage{amsmath, amsfonts, amsthm, amssymb}

\usepackage{hyperref}

\newtheorem{theorem}{Theorem}[section]
\newtheorem{lemma}{Lemma}[section]
\newtheorem{assumption}{Assumption}[section]
\theoremstyle{definition}
\newtheorem{definition}{Definition}[section]
\newtheorem*{remark}{Remark}

\title{Imputation Strategies Under Clinical Presence: Impact on Algorithmic Fairness}

\author{%
    Vincent Jeanselme\\
    MRC Biostatistics Unit\\
    University of Cambridge, UK\\
    The Alan Turing Institute\\
    \texttt{vincent.jeanselme@mrc-bsu.cam.ac.uk} 
\AND
    Maria De-Arteaga\thanks{Equal contribution}\\
    McCombs School of Business\\
    University of Texas at Austin, USA
\And
    Zhe Zhang$^*$\\
    Rady School of Management\\
    University of California, San Diego, USA
\AND
    Jessica Barrett\\
    MRC Biostatistics Unit\\
    University of Cambridge, UK
\And
    Brian Tom\\
    MRC Biostatistics Unit\\
    University of Cambridge, UK
} 
\begin{document}
\maketitle
\begin{abstract}
Machine learning risks reinforcing biases present in data and, as we argue in this work, in what is absent from data. In healthcare, societal and decision biases shape patterns in missing data, yet the algorithmic fairness implications of group-specific missingness are poorly understood. The way we address missingness in healthcare can have detrimental impacts on downstream algorithmic fairness. Our work questions current recommendations and practices aimed at handling missing data with a focus on their effect on algorithmic fairness, and offers a path forward. Specifically, we consider the theoretical underpinnings of existing recommendations as well as their empirical predictive performance and corresponding algorithmic fairness measured through subgroup performances.
Our results show that current practices for handling missingness lack principled foundations, are disconnected from the realities of missingness mechanisms in healthcare, and can be counterproductive. For example, we show that favouring group-specific imputation strategy can be misguided and exacerbate prediction disparities. We then build on our findings to propose a framework for empirically guiding imputation choices, and an accompanying reporting framework. 
Our work constitutes an important contribution to recent efforts by regulators and practitioners to grapple with the realities of real-world data, and to foster the responsible and transparent deployment of machine learning systems. We demonstrate the practical utility of the proposed framework through experimentation on widely used datasets, where we show how the proposed framework can guide the selection of imputation strategies, allowing us to choose among strategies that yield equal overall predictive performance but present different algorithmic fairness properties.
\end{abstract}

\newpage
\section{Introduction} 
Healthcare is increasingly leveraging machine learning to improve patient care. This often occurs through machine learning models for risk prediction, prioritisation, or even treatment response modelling. These machine learning models typically rely on observational data. The data generation process that shapes this data involves a complex interaction between patients and the healthcare system, which is referred to as clinical presence~\cite{jeanselme2022deep}. Each observation, from orders of laboratory tests to treatment decisions, depends on access to medical care, patients' medical states, and practitioners' expert decisions. As a result, the collected medical records suffer from missing observations~\cite{sharafoddini2019new}. We refer to the missingness that stems from these clinical interactions as \emph{clinical missingness}.

Clinical missingness is widespread in medical observational data~\cite{little2012prevention, weiskopf2013defining, weiskopf2013methods}. Medical records reflect and inform treatment, and are not primarily gathered for scientific discovery and analysis. The prevalence of missing data is an issue because statistical analysis and machine learning often require complete data. Practitioners, therefore, routinely rely on preprocessing strategies, such as imputation, to address missingness in their medical datasets. However, the importance of this step is often overlooked. In their literature review,~\cite{nijman2022missing} note that 65\% of machine learning papers on clinical applications mention the problem of missingness, among which, less than 10\% report their assumptions about missing data, and only 3\% analyse how their choice of handling missing data impacts their conclusions.

Overlooking clinical missingness may have consequential repercussions on algorithmic fairness. This algorithmic fairness connection arises because clinical missingness patterns are often group-specific. In other words, the patterns and causes of missing data can vary between different population subgroups. Group-specific patterns of missingness are particularly notable in medical datasets. They can occur due to historical healthcare biases or disparities, which subsequently influence healthcare access, treatment, and outcomes~\cite{chen2021algorithm, freeman2000racial, jeanselme2021sex, kim2016sex, norris2008race}. For instance, limited access to healthcare resources can translate into group disparities in available testing procedures.
Additionally, medical guidelines and practice can also reinforce existing group inequalities by focusing primarily on populations considered high-risk. Consequently, these differences in medical interactions translate into group-specific missingness in testing. For instance, this is evidenced by~\cite{lin2018racial}, who show increased missingness in Black patients' family history records.

In this work, we tackle the question of how to evaluate and compare imputation strategies in a manner that accounts for algorithmic fairness, considering the realities of missingness patterns in healthcare. Several works have called for more attention to the impact that historical medical biases reflected in missing data may have on algorithmic fairness~\cite{ahmad2019challenge,ghassemi2020review,gianfrancesco2018potential,mitra2023learning,rajkomar2018ensuring}. Previous research has shown that from an algorithmic fairness perspective, imputation is preferable over complete case analysis~\cite{martinez2019fairness,fricke2020missing,fernando2021missing}.  
As a result of these growing concerns over missingness handling, there have been attempts to provide recommendations to foster best practices that mitigate predictive biases. For example, studies often use a single strategy with all likely confounders included to ensure the plausibility of the missingness assumption~\cite{haukoos2007advanced,li2021imputation}. Because group missingness differences are a concern, such an approach can motivate imputation strategies that control for or stratify by group membership, e.g.~\cite{howard2011disparities}. Notably, while previous works have noted that there is no universally best imputation strategy~\cite{caton2022impact,jeong2022fairness}, they have either not considered whether group-specific imputation is still preferable over its population variant~\cite{caton2022impact}, or explicitly recommended group-specific strategies on the basis of theoretical results that make strong assumptions about the missingness process~\cite{jeong2022fairness}. 
We question the validity of current imputation practices that aim to reduce inequities and offer a path forward. 

\paragraph{Theoretical contribution.} We analyse the theoretical underpinnings of current recommendations for imputation. First, we provide a structured view of the relationship between clinical presence mechanisms and group-specific missingness patterns, highlighting how traditional missingness assumptions fail to capture the complexity of clinical missingness. Then, we consider the recommendation of favouring group-specific imputation strategies and show that there are no principled foundations to justify it. Specifically, we demonstrate that group-specific strategies can increase reconstruction error compared to their population variants, and in particular, they can reduce data quality for marginalised groups and widen the reconstruction error gap between groups. 

While the current literature on imputation has focused on minimising reconstruction error, these errors cannot be assessed without knowledge of the missingness process. We recognise the importance of minimising reconstruction errors as an aim in itself, as the quality of available data has critical implications on inference and, consequently, our understanding of health. To this point, our theoretical results show that the choice of imputation to optimally reduce reconstruction error disparities is most often underdetermined under unknown missingness processes. In other words, if practitioners' central concern is data quality, then group-specific imputation strategies do not guarantee better results.

In the context of machine learning, practitioners are not always concerned with reconstruction error but rather with improving quality and algorithmic fairness of downstream predictive performance. In this context, our results show that improving reconstruction error is neither necessary nor sufficient to improve downstream predictive properties. Through simulations, we empirically demonstrate that the imputation strategy that leads to better reconstruction error or smaller error gaps across groups does not necessarily yield the best predictive performance or smaller gaps in predictions, and reciprocally. Furthermore, even in simple cases, the imputation strategy with the best performance in terms of algorithmic fairness varies widely depending on the missingness mechanism, and group-specific strategies are not always better. 

\paragraph{Practical contribution.} Together, these results lead to a crucial conclusion. While there is no universally best imputation strategy based on algorithmic fairness, empirical evaluation of downstream predictive properties following different imputation choices can inform this choice. Based on these findings, we introduce a framework for evaluating and guiding the selection of imputation strategies and provide a Python operationalisation to facilitate its application. We offer a theoretical grounding for when and why the framework will yield reliable results. The proposed work constitutes an important contribution to recent efforts by regulators and practitioners to grapple with real-world data realities and to foster the responsible and transparent deployment of machine learning systems.

Through experimentation on widely used datasets, we demonstrate how the proposed framework can guide the selection of imputation strategies, allowing us to choose among imputation strategies that yield equal overall predictive performance but present different algorithmic fairness properties. Particularly, on both MIMIC-III and SUPPORT datasets, the framework shows that the choice between two popular imputation methods can invert the directionality of fairness between two demographic groups (Black vs Non-Black). If this thorough evaluation of imputation choice was not conducted, the choice of either of these imputation methods could result in substantively different outcomes and algorithmic fairness implications at deployment. Through this assessment, we demonstrate how practitioners can inform imputation choices. 

\paragraph{Managerial contribution.} While the impact of imputation on algorithmic fairness has gained attention in the literature, our work challenges existing imputation practices and offers a path forward. In addressing the question of how to choose imputation strategies when concerned with algorithmic fairness, we demonstrate that current imputation practices do not improve predictive performance nor fairness gap. 
This misalignment has an important managerial implication: practitioners in healthcare must change their imputation practices. Our proposed framework provides a data-driven tool for informing imputation choice through systematic evaluation of imputation's impact on downstream performance and algorithmic fairness.

\paragraph{Outline.} Section~\ref{sec:lit} reviews the literature associated with missingness, fairness, and their intersection. Section~\ref{sec:missingness} introduces and formalises common, historical clinical missingness scenarios. Section~\ref{sec:reconstruction} theoretically shows how recommendations for group-specific imputation to improve reconstruction performance are not well-founded and are sensitive to the missingness process. 
Section~\ref{sec:synthetic} demonstrates that, while assessing reconstruction performance is difficult, predictive performance and fairness can be evaluated, and, importantly, that better reconstruction methods are not necessarily the ones that provide the best downstream fairness outcomes. Section \ref{sec:framework} ties together our findings by introducing a framework to assess the impact of missingness handling on fairness outcomes, as well as how to report it appropriately for deployment. Finally, we apply the proposed framework in Section~\ref{sec:mimic} on the widely used MIMIC III dataset, demonstrating how the framework reveals how real-world study conclusions can be meaningfully impacted under different imputation strategies. We conclude in Section \ref{sec:conclusion}.

\section{Related work}
\label{sec:lit}
This work explores the link between missingness and algorithmic fairness in machine learning for healthcare. In this section, we review the related literature across domains. 

\subsection{Clinical missingness}
Missingness naturally occurs in medical studies in which information is recorded for clinical decision-making~\cite{haukoos2007advanced}. Missing data may, therefore, present informative patterns. Current clinical understanding of missingness relies on the three well-studied patterns~\cite{little2019statistical}: \emph{Missing Completely At Random} (MCAR) --- random subsets of covariates are missing; \emph{Missing At Random} (MAR) --- missing data patterns are a function of observed variables; and \emph{Missing Not At Random} (MNAR) --- missing patterns depend on unobserved variables, potentially on the missing values themselves. 

Missing data prohibits the use of traditional statistical models that require complete data. Ignoring patients with missing data, also known as complete case analysis, lowers statistical power~\cite{little1989analysis}. Thus, practitioners often replace missing data, selecting from a wide range of available imputation strategies. These include single imputation strategies, which replace missing data with a single value such as mean, median, or nearest neighbour value~\cite{batista2002study, bertsimas2021imputation}, or multiple imputation strategies, which propose multiple possible values for each missing one~\cite{newgard2015missing, rubin2004multiple, white2011multiple} as a way to quantify the uncertainty associated with the missingness process. Typically, both types of imputation strategies assume MCAR and/or MAR patterns, and all associated theoretical guarantees depend on these assumptions. 

These common imputation strategies may be ill-adapted to handle clinical missingness reflective of more complex patterns. Crucially, missingness patterns are non-identifiable from observational data alone and require knowledge of the missingness process and domain expertise for adequate modelling~\cite{beaulieu2018characterizing}. The recommended strategy to tackle this non-identifiability issue is to condition the imputation strategy on potential confounders to render these missingness assumptions more plausible~\cite{haukoos2007advanced}. 

Previous literature has studied the consequences of making incorrect assumptions about the missingness process, with a focus on potential parameter misestimation, e.g.~treatment effect or odds ratios~\cite{bennett2001can}. 
In this work, we first show that the traditional categorisation of missingness patterns used in the literature do not account for the realities of historical disparities that may be reflected in clinical missingness. 
We then show potential shortcomings of imputation practices when there are such clinical missingness patterns. We provide theoretical and empirical evidence showing that the recommendation of controlling or stratifying on group membership can be counter-productive.

\subsection{Algorithmic fairness in medicine}
\label{subsection:metric}
Historically, medical research and practice have been marked by biases against marginalised groups~\cite{kartoun2022prediction, van2002research}. For instance, developing risk score assessment tools using data from populations with low ethnic diversity may detrimentally impact risk management for minority groups~\cite{kartoun2022prediction}. Our work is grounded on a detailed characterisation of the different ways in which disparities in the healthcare system and in the development of healthcare technologies have shaped missingness patterns in data.

Machine learning has the risk of reinforcing biases present in the data~\cite{chouldechova2020snapshot}, and, as we argue in this paper, in what is absent from the data. Measuring and mitigating the risk of inequitable real-world deployment is the aim of algorithmic fairness. Given that reinforcing biases is of critical concern in medicine, where data marked by inequalities can influence life-threatening decisions, algorithmic fairness has become a central concern in machine learning for healthcare~\cite{chen2020ethical}. 

When quantifying algorithmic fairness, three families of definitions emerge from the multiple definitions proposed in the literature~\cite{mitchell2021algorithmic, verma2018fairness}. \emph{Individual fairness}~\cite{dwork2012fairness} deems an algorithm fair if similar individuals (according to a relevant metric) are treated similarly. \emph{Causal fairness} deems an algorithm fair if the prediction would remain unchanged if an individual's group membership changed~\cite{kusner2017counterfactual}, or if group membership does not affect the prediction through inadmissible pathways~\cite{nabi2018fair}. \emph{Group fairness} defines fairness in terms of equal performance across groups, where the performance metric of interest may vary~\cite{hardt2016equality,chouldechova2017fair}. Individual fairness requires access to a relevant, task-specific, distance metric to assess who is ``similar", and notions of causal fairness require knowledge of the causal graph between all covariates and target labels. In practice, it is rare to have access to such distance metrics or causal graphs. As a result, group fairness definitions are the most widely used in practice.

In healthcare,~\cite{rajkomar2018ensuring} proposes to quantify group fairness as the difference in (i)~observed outcomes, (ii)~model performance or (iii)~care allocation. In this paper, we focus on model performance, and in particular the ``\emph{equal performance}" definition of algorithmic fairness~\cite{rajkomar2018ensuring}. This definition evaluates if the model performs comparably across groups~\cite{chouldechova2018case, flores2016false, noriega2019active} by comparing group-level metrics, for relevant metrics of interest. This definition has been used to quantify if marginalised groups would be impacted differently by medical models' deployment~\cite{chen2018my, chen2019can, pfohl2019creating, seyyed2020chexclusion, zhang2020hurtful}. For instance,~\cite{seyyed2020chexclusion} demonstrates X-ray classifiers' performance gap between groups and highlights the detrimental misdiagnosis for marginalised groups if the models were deployed.
In this paper, we focus on group fairness as the fairness metric of interest to contribute a study on how missingness patterns may shape algorithmic biases, a problem that has been understudied in the algorithmic fairness literature.

\subsection{Algorithmic fairness and missingness}

A central thrust of research on algorithmic fairness has focused on developing methods to mitigate disparities, such as resampling~\cite{kamiran2010classification}, loss regularisation~\cite{kamishima2011fairness} or post-processing adjustment~\cite{hardt2016equality}. Such approaches, however, assume that data is complete. Furthermore, characterisations of sources of algorithmic bias~\cite{mitchell2021algorithmic,chouldechova2020snapshot,barocas2016big} rarely focus on the potential impact of missing data. 

At the intersection of algorithmic fairness, missingness, and machine learning for health, \cite{ahmad2019challenge, ghassemi2020review, gianfrancesco2018potential, mitra2023learning, rajkomar2018ensuring} describe multiple challenges linked to medical data, among which they state that historical biases may lead to missingness patterns that could impact fairness, and call for more research in this realm. An emerging body of research has begun to study this interplay between algorithmic fairness and missing data. On the medical side, \cite{ganju2020role} encourages using clinical decision support systems to improve data collection, as the authors show that unfair medical decisions can emerge from missing standardised testing in marginalised groups. On the statistical side, \cite{martinez2019fairness, fricke2020missing, fernando2021missing} show that mean imputation presents better algorithmic fairness properties compared to complete case analysis. In an effort to improve downstream performance and algorithmic fairness when performing complete case analysis,
\cite{zhang2021assessing} introduces a weighting scheme to correct for unobserved data. Similarly, \cite{wang2021analyzing} proposes a group-specific adjustment as a function of the group's observation rate to reduce the accuracy gap between groups.

Closer to our work, \cite{caton2022impact, jeong2022fairness} show that the choice of imputation strategy may lead to distinct performance gaps. \cite{caton2022impact} compare different imputation strategies using simulation on different datasets. They empirically show that no imputation strategy is consistently best across datasets, using gaps in group performance as their fairness metric. In their assessment, no group-specific imputation methods are considered. \cite{jeong2022fairness} theoretically
show that no imputation strategy can be consistently best---defined in terms of group fairness---across datasets or modelling strategies. Of note, despite their overall finding noting that no strategy is \emph{best},  \cite{jeong2022fairness} still recommends group-specific imputation with the goal of improving reconstruction error.

Our work contributes to this line of research by anchoring our analysis on group-specific missingness patterns that are common to clinical settings. In particular, we show that recommendations favouring group-specific imputation are flawed and potentially detrimental to fairness. Through empirical and theoretical results, we show that group-specific imputation may yield worse reconstruction errors for the marginalised group, as well as worse reconstruction error gaps. This finding directly challenges common recommendations, anchored on theoretical assumptions that do not consider the nature of clinical missingness patterns. Furthermore, we provide a path forward by noting that under assumptions of stability in data generation, one can empirically select an imputation strategy given a dataset and a predictive task of interest, and we provide a framework to guide the choice of imputation strategy and to report missingness handling.

\section{Clinical missingness}
\label{sec:missingness}
The central motivation of this paper is that the underlying missingness process can reflect disparities and can have unanticipated impacts on group-specific performance. To better understand how missingness occurs in clinical settings and formalise these missingness patterns, we review the clinical literature and identify three clinical missingness patterns:

\begin{enumerate}[label=(S{{\arabic*}})]
    \item \textbf{Limited access to quality care}. Some groups do not have access to the same health services and quality care as others, which may result in more missing covariates for disadvantaged groups.
    \item \textbf{(Mis)-informed collection}. Often, medical research has focused on a subset of the population. The resulting guidelines may be ill-adapted to other groups, and relevant covariates may be missing due to standard recommendations of when to collect such covariates.
    \item \textbf{Confirmation bias}. The collection of certain types of data depends on practitioners' unobserved prognoses, which may be affected by group-specific expectations.
\end{enumerate}

\begin{figure*}[t!]
    \makebox[\textwidth][c]{\includegraphics[width=0.9\textwidth]{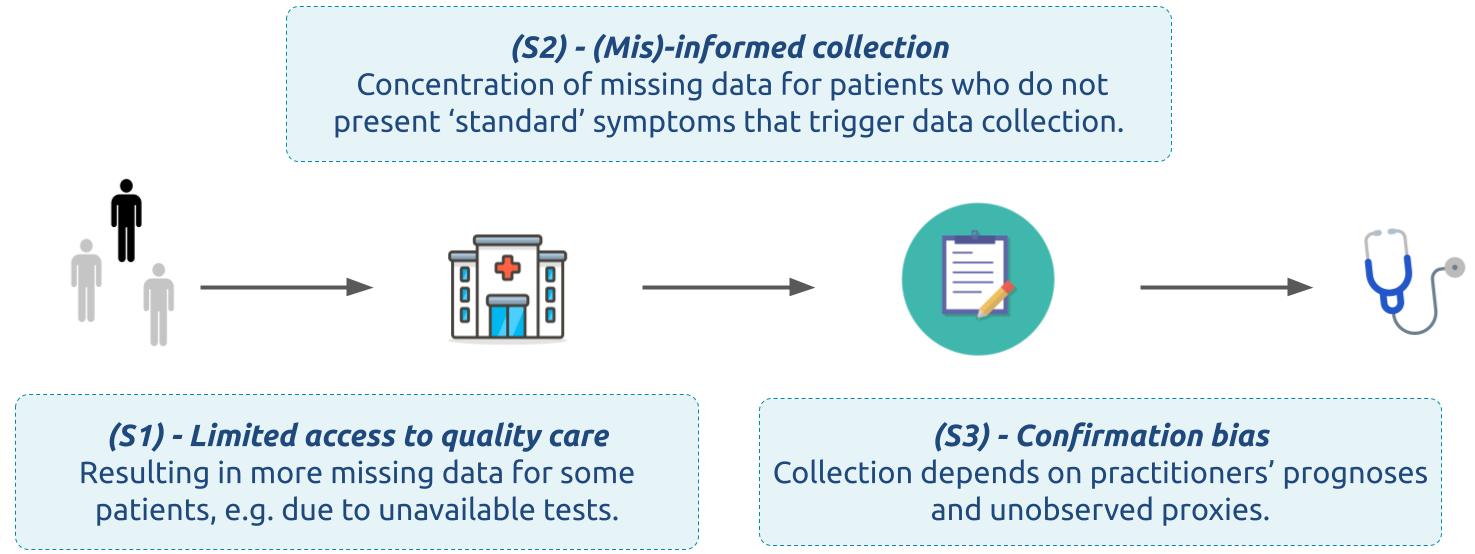}}
    \caption{Examples of group-specific clinical presence mechanisms.}
    \label{fig:scenarios}
\end{figure*}

These scenarios have a substantial history of medical evidence, which we summarise in Section~\ref{sub:evidence}. Furthermore, these three scenarios are mathematically distinct, as shown by the formalisation in Section~\ref{sub:form}. 

\subsection{Clinical evidence}
\label{sub:evidence}

\paragraph{Historical evidence of limited access to quality care.} Socioeconomic factors impact access to care and, consequently, missingness. For instance, education~\cite{barik2015issues}, urban residence~\cite{barik2015issues}, insurance~\cite{hoffman2008health}, distance to hospitals~\cite{barik2015issues} or mobility~\cite{szczepura2005access} have been shown to impact patients' interactions with the healthcare system. These differences in medical interactions may translate into inconsistent medical history~\cite{gianfrancesco2018potential}, limited access to advanced diagnostic tools~\cite{lin2019electronic}, and may also impact behaviours such as additional waiting time before seeking care~\cite{weissman1991delayed}, and avoidance of preventive care~\cite{smith2018access}. All these realities can result in group-specific missingness patterns. For instance, the lack of medical history is in itself a problem of missing data. 
Avoidance or lack of access to care may translate into less frequent check-up data, and, therefore, a sparse record of patients' health evolution. Finally, limited access to advanced diagnostic tools also leads to absent tests in medical records.

\paragraph{Historical evidence of (mis)-informed collection.} Historically, studies focused on perceived highest-risk groups and were constrained by the available and willing-to-participate patient population. Cardiovascular diseases have been predominantly studied in men~\cite{vogel2021lancet}, breast cancer in women~\cite{arnould2006breast, giordano2018breast}, skin cancers in whiter skins~\cite{gloster2006skin}, and autism in men~\cite{gould2011missed}. Resultant medical practices and guidelines target these groups. However, evidence shows the prevalence of these conditions among a more diverse population: 1 out of 3 women dies from cardiovascular disease~\cite{garcia2016cardiovascular}, men experience 1\% of breast cancers~\cite{yalaza2016male}, neoplasm can be cancerous in non-white populations~\cite{gloster2006skin}, and autism has an estimated male-female ratio of 4.2:1~\cite{zeidan2022global}. Stemming from social constructs and biological differences, distinct groups may present and express different symptoms for the same condition. \cite{mauvais2020sex} shows how biological sex influences condition manifestation through genetics and how the associated social construct of gender has epigenetic effects across a large set of cardiac conditions. 
Thus, there is a misalignment between condition manifestation in these groups and existing tests and guidelines. This can result in missing covariates that are helpful in identifying the condition for the under-studied group. Rather, guidelines recommend testing on the observation of ``standard" symptoms. If the symptoms considered do not include the symptoms exhibited by a marginalised subgroup, this will result in more missing tests for this group. For instance, women with heart failure are more likely to experience fatigue while men are more likely to report chest pain~\cite{heo2019sex}. As guidelines have focused on male patients~\cite{stromberg2003gender}, this difference in the expressed symptoms may not trigger further testing for women, as suggested by the rate difference in diagnostic tests offered to men and women~\cite{stromberg2003gender}.

\paragraph{Historical evidence of confirmation bias.} Practitioners may choose to perform a test only if they suspect that it will be informative. Research shows that the presence of tests in medical records is more predictive of the outcome than the actual values of the tests~\cite{agniel2018biases, sisk2020informative}. \cite{wells2013strategies} also suggests that missing laboratory tests correspond to healthier patients. Conversely,~\cite{rusanov2014hidden, sharafoddini2019new, weiskopf2013sick} show that sicker patients present more complete data. Under this missingness process, the way a condition manifests for different groups may result in group-specific testing patterns. For instance, general practitioners are more likely to record weights for underweight or overweight patients~\cite{nicholson2019determinants}. However, weights' distributions differ depending on socioeconomic characteristics~\cite{nicholson2019determinants} and, consequently, can result in group-specific patterns of observation.

\subsection{Formalisation}
\label{sub:form}
Each of the three common scenarios above has different dependence structures between the missingness processes and the resulting data. For further analysis, we formalise the causal representation of these scenarios.

Consider a set of covariates, $X$, and a medical condition of interest, $Y$, both influenced by the group membership, $G$. Here, note that the condition may refer to a \emph{future} condition, i.e. an outcome of interest, or a \emph{current} condition, i.e. a diagnosis. The direction of the association between $Y$ and $X$ depends upon a given problem formalisation. For instance, a measured symptom may reflect a condition, e.g. chest pain may be a result of a cardiac condition; meanwhile, an environmental exposure may render an outcome more likely, e.g. smoking may increase the risk of lung cancer. We choose not to restrict the problem type and maintain both directions in our formalisation. Following the notations from~\cite{mohan2021graphical}, let $O_i$ be the indicator of observation of $X_i$ such that the observed value is defined as:
\begin{align*}
    X_i^* = 
    \begin{cases}
        \varnothing &\text{ if } O_i = 0\\
        X_i &\text{ otherwise}
    \end{cases}
\end{align*}

\begin{figure*}[ht!]
    \renewcommand*\thesubfigure{S\arabic{subfigure}}
    \centering
    \begin{subfigure}[b]{0.3\textwidth}
         \centering
         \includegraphics[width=\textwidth]{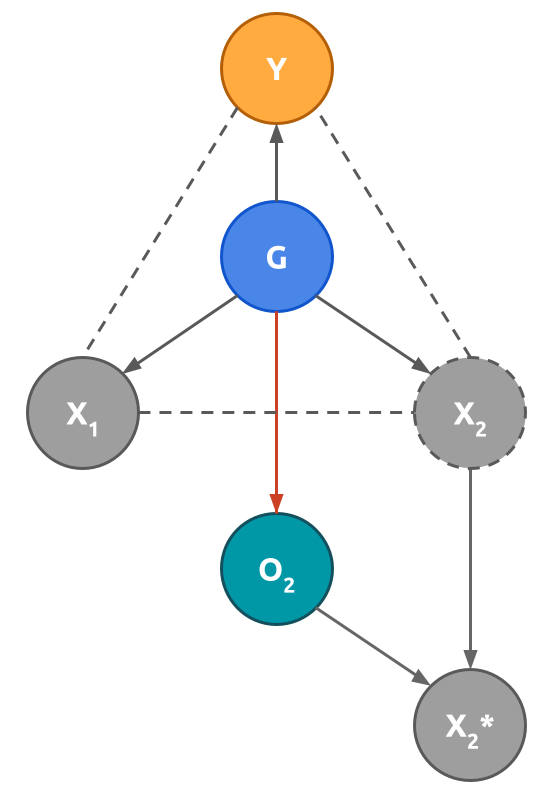}
         \caption[size=14pt]{ }
     \end{subfigure}
     \begin{subfigure}[b]{0.3\textwidth}
         \centering
         \includegraphics[width=\textwidth]{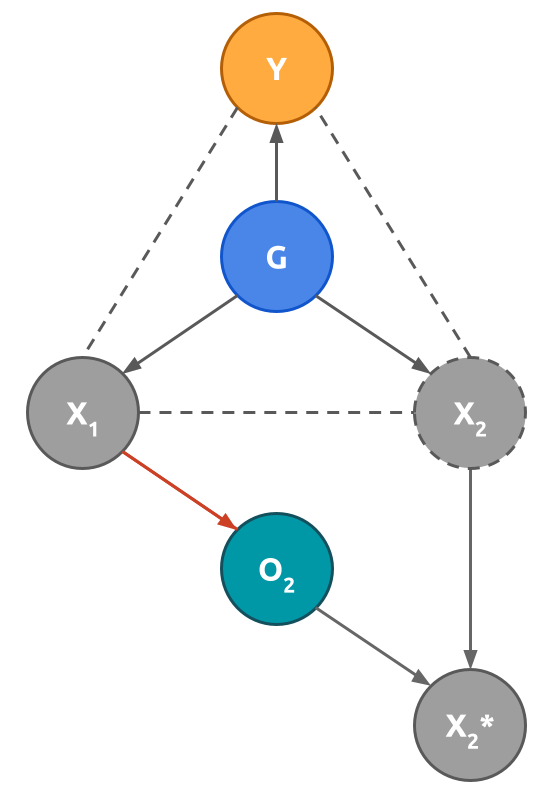}
         \caption[size=14pt]{ }
     \end{subfigure}
     \begin{subfigure}[b]{0.3\textwidth}
         \centering
         \includegraphics[width=\textwidth]{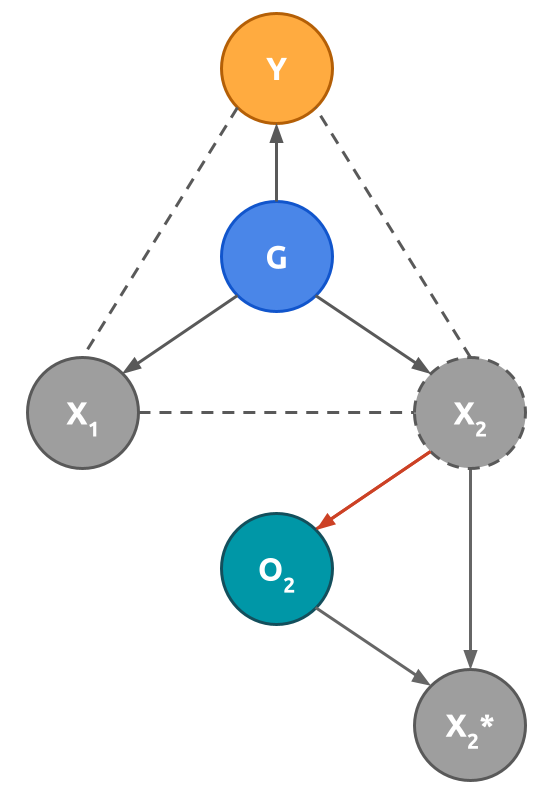}
         \caption[size=14pt]{ }
     \end{subfigure}
     \caption{Graphs associated with the identified clinical missingness scenarios. {Full circled covariates are observed, dashed ones are unobserved. $Y$ is the condition, $G$ is the group membership, $X_1$ and $X_2$ are the two covariates. $O_2$ is the decision to observe the associated $X_2$. Red arrows underline the dependency differences across scenarios. Undirected arrows represent problem-specific directed dependencies.}}
    \label{fig:dags}
\end{figure*}

We formalise the proposed scenarios in the bi-dimensional case: $X$ is the concatenation of two, potentially correlated, covariates ($X_1$, $X_2$). One covariate $X_1$ is always observed, while $X_2$ is potentially missing. 
Note that this can be generalised to higher dimensions by considering $X_1$ and $X_2$ to be sets of covariates rather than single covariates.
Moreover, while 
real-world missingness patterns likely result from a mixture of these processes over multiple covariates, illustrating the difference between these scenarios using two covariates allows us to study how different clinical presence results in distinct missingness patterns. Following these notations, Figure~\ref{fig:dags} displays the graphs associated with each scenario. Each graph shows the dependencies between missingness, group membership, covariates, and condition. 
 
The graphs in Figure~\ref{fig:dags} illustrate the differences between these three clinical missingness scenarios. Following the traditional missingness framework, Scenarios 1 (S1) and 2 (S2) both represent Missing at Random (MAR) missingness processes, with key differences between them that we outline below. Scenario 3 (S3) represents a Missing Not at Random (MNAR) missingness process, where the missingness depends on the missing covariate itself. This latter occurs, for instance, when physicians measure weight only if expected to be outside of a given range.

In (S1), missingness is directly dependent on group membership. An example of this scenario is 
increased missingness in family medical history records for Black patients~\cite{lin2018racial}. Meanwhile, in (S2), missingness is \emph{indirectly} linked to group membership, through group-specific symptom manifestation in $X_1$. For example, consider $Y$ as a heart failure condition. Women and men present a similar risk of heart failure~\cite{lam2019sex}, however, symptom expression differs. 
For instance, women present symptoms of fatigue, while men have chest pain~\cite{heo2019sex}. 
If $X_1$ is a measure of chest pain and those with chest pain are more likely to receive an echocardiogram test $X_2$, then the diagram for (S2) shows that women with heart failure will be less likely to have echocardiogram tests observed. More generally, (S2) represents scenarios where $X_2$ is observed only when $X_1$ is within a given range, and the values of $X_1$ depend on group membership.

This distinction between (S1) and (S2) highlights the connection between clinical missingness and group-specific patterns. Notably, these characterisations also serve to show that the traditional categorisation of missingness patterns is underspecified and thus does not capture these phenomena. 
As shown in this section, both of these patterns fall under the MAR assumption, even though the underlying causal graph and the underlying societal patterns are different. 
This underspecification may have consequential impacts because recommendations for choosing imputation strategies often rely on theoretical guarantees upon the traditional categorisation of missingness patterns. 
Particularly, many imputation strategies make such assumption upon the missingness process, and multiple recommendations encourage controlling or stratifying for covariates such as group membership to improve the plausibility of these assumptions~\cite{haukoos2007advanced}.

Subsequently, in the next two sections, we explore the potential consequences of relying on the current categorisation and resulting recommendations, building on the characterisation provided in this section. Specifically, in Section~\ref{sec:reconstruction} we present a theoretical analysis of the implications of controlling for group membership, which is a central recommendation stemming from the traditional view on imputation. Then, in Section~\ref{sec:synthetic} we provide an empirical analysis of the clinical presence scenarios introduced in this section, demonstrating how the type of missingness can dramatically impact the reconstruction error and downstream predictive performance of different imputation strategies.

\section{Theoretical analysis of imputation and group fairness}
\label{sec:reconstruction}

As shown in Section~\ref{sec:missingness}, clinical presence patterns can lead to complex group-specific missingness patterns. In this section, we explore how the common practice of group-specific imputation, i.e. stratifying imputation per group, impacts reconstruction error. We demonstrate how this practice may counter-intuitively amplify group disparities under clinical missingness patterns. 
We do so by focusing on a commonly used imputation strategy: mean imputation, considering its population and group-specific variants. This commonly used imputation strategy is amenable to tractable analyses and allows us to demonstrate why the group-specific variant can be harmful to reconstruction error. 

Our focus on reconstruction error in this section is anchored on two elements. First, reconstruction error can be theoretically quantified. Second, reconstruction error minimisation is typically the objective of imputation strategies. Hence, it ensures that we are assessing the performance of the strategies with respect to their stated objectives.

Population mean imputation, which is commonly used due to its ease of implementation, imputes missing values with the mean values of the population. This approach assumes (i)~a Missing Completely at Random (MCAR) process and (ii)~a homogeneous population. As these assumptions are understood to be unlikely to hold in medical data, intuition often leads researchers and practitioners to use mean imputation in subgroups of the population, with the aim of better capturing group heterogeneity and improving reconstruction quality for groups. In the group-specific variant, missing values are imputed with the mean of the group to which the data point belongs. By stratifying by subgroups, this strategy assumes a MAR process in which each subgroup presents MCAR patterns aligned with the mean imputation's assumption. This group-specific mean imputation is a common imputation strategy~\cite{burt2016structural, chun2021utility, nielsen2020predictors}, recommended when fairness is a primary consideration~\cite{jeong2022fairness}.

As a summary of our theoretical findings below: we show that under different missingness patterns, group-specific mean imputation can, counter-intuitively, deteriorate the reconstruction quality in the marginalised group that one aims to protect (Theorem~\ref{theorem:lessbiaspop}). Additionally, it may further increase the gap in reconstruction error between groups (Theorem~\ref{theorem:lessunfair}). These findings result from the novel expression of reconstruction error as a function of group-specific properties of the missingness process.

\subsection{Problem setting}
Population mean imputation replaces any missing value in a covariate with that covariate's mean, observed at the population level. The group-specific extension of this strategy replaces each missing value with the mean computed in the group to which the data point belongs. Formally, population mean imputation and group-specific mean imputation associate the imputed value $\Tilde{x}_i$ to the point $i$ as follows:
\begin{equation*}
    \Tilde{x}^{pop}_{i} = \left\{
    \begin{array}{ll}
        \mu^O  & \mbox{if point $i$ is missing, i.e., } o_i = 0 \\
        x_i & \mbox{if } o_i = 1
    \end{array}
    \right.,
    \hskip 2em
    \Tilde{x}^{group}_{i} = \left\{
        \begin{array}{ll}
            \mu^O_{g_i}  &\mbox{if } o_i = 0 \\
            x_i & \mbox{if }o_i = 1
        \end{array}
    \right.
\end{equation*}
where $\mu^O$ is the observed population mean,~$\mu^O = \frac{\sum_i o_i x_i}{\sum_i o_i}$, and $\mu^O_g$ is the observed group-specific mean,~$\mu^O_g = \frac{\sum_{i \in P_g} o_i x_i}{\sum_{i \in P_g} o_i}$. $P_g = \{i \mid g_i = g\}$ are the indices of $g$-members, with $o_i$ indicating if point $i$ is observed and $g_i$ indicating group membership. Note that the dimension associated with covariates with missing data ---in the graphical formalisation $X_2$---are omitted in this notation as these imputation strategies treat each covariate independently.

Given any group $g$, we consider two metrics of interest. First, the reconstruction error for group $g$ under imputation strategy $\mathcal{I}$, $L^\mathcal{I}_g$. Second, the reconstruction \emph{error gap} between this group's error and the error in the rest of the population, $\Delta_g^\mathcal{I}$. Group-specific reconstruction error $\mathcal{L}_g^\mathcal{I}$ is valuable to improve each group's performance; whereas the reconstruction error gap between groups, $\Delta_g^\mathcal{I}$, is also valuable to consider the fairness of performance between groups. Formally, these quantities are defined as follows:
\begin{definition}[Reconstruction error]
\label{def:rec}
 The reconstruction error of an imputation $\mathcal{I}$ in a group $g$ is the average distance between the underlying true $x_i$ and imputed values $\Tilde{x}^\mathcal{I}_i$ over all missing data for that group:
    \begin{equation}
    \label{eq:def}
        L^\mathcal{I}_g = \mathbb{E}_{x_i}\left[||\Tilde{x}^\mathcal{I}_i - x_i||_2^2 \mid g_i = g, o_i = 0\right]
    \end{equation}
\end{definition}
An optimal imputation strategy assigns the true value to any missing value, resulting in $L_g = 0$. In practice, estimating the reconstruction error requires knowing $x$, i.e. the underlying distribution and missingness process, which is rarely achievable in real-world settings.

To quantify algorithmic fairness, we utilise the commonly-used equal performance definition~\cite{rajkomar2018ensuring}, in the context of reconstruction error:
\begin{definition}[Equal Performance]
\label{def:equal}
An imputation $\mathcal{I}$ is fairer than another $\mathcal{J}$ with regard to group $g$ if its absolute performance gap with respect to the rest of the population is the smaller under $\mathcal{I}$ than under $\mathcal{J}$, i.e. $|\Delta_g^\mathcal{I}| < |\Delta_g^\mathcal{J}|$, where
$\Delta_g^\mathcal{I} := d(\mathcal{I}(\{X_i\}_{G_i = g})) - d(\mathcal{I}(\{X_i\}_{G_i \neq g}))$ for some performance metric $d$, and $(X_i, G_i)$ the associated covariates and group for patient $i$.
\end{definition}

Connecting this generic measure of algorithmic fairness (Definition~\ref{def:equal}) with reconstruction error as the performance metric of interest (Definition~\ref{def:rec}) leads to the measure of the reconstruction error gap, defined by the difference in a group's reconstruction error compared with the rest of the population: $\Delta^\mathcal{I}_{g} = L^\mathcal{I}_{g} - L^\mathcal{I}_{\neg g}$.

Under this definition, imputation $\mathcal{I}$ is considered fairer than another $\mathcal{J}$ if its reconstruction gap is smaller, i.e., $|\Delta^\mathcal{I}_{g}| < |\Delta^\mathcal{J}_{g}|$. A null gap reflects equal error across groups. If $\Delta^\mathcal{I}_{g}>0$, then this means group $g$ has a larger reconstruction error than the rest of the population, a group fairness concern. Throughout our analysis, we compare the reconstruction errors of each group under different strategies, as well as the resulting performance gaps.

\subsection{Imputation's impact on reconstruction error and gap} \label{sub:reconstruction}

We now investigate the impacts of the imputation strategy $\mathcal{I}$ and the missingness process on these measures of algorithmic fairness. In Lemma~\ref{theorem:expression}, we express each reconstruction error, $L^{group}_g$ and $L^{pop}_g$, as a function of (i)~the underlying distribution of the covariate, and (ii)~the missingness process. The missingness process influences the reconstruction error through: $\alpha_g$, the observation rate, and $\rho_{g}$, the correlation between observation indicators and the covariate values. Both of these can differ across groups, but note that while $\alpha_g$ is observed, estimation of $\rho_g$ requires an assumption upon the missingness process. To the best of our knowledge, the use of $\rho_{g}$ to reflect the group-specific missingness process in expressing reconstruction error is a novel formulation that is key to the findings of this section.

\begin{lemma}[Group and population mean imputations' reconstruction error]
\label{theorem:expression}
    Assuming i.i.d. data points $\{x_i\}$, one can express the reconstruction error in group $g$ resulting from group mean imputation as:
        \begin{equation}
        \label{eq:bias:group}
            L^{group}_{g} = \Bigg(\overbrace{\eqnmarkbox[NavyBlue]{missingness}{-\frac{1}{\sqrt{\alpha_g(1-\alpha_g)}} \cdot \rho_g} \cdot
                                \eqnmarkbox[OliveGreen]{comp}{\sigma_{X \mid G = g}}}^{\textcolor{Mulberry}{\bm{B^{group}_{g}}}}\Bigg)^2 + \eqnmarkbox[BurntOrange]{error}{\sigma^2_{X \mid O = 0, G = g}}
        \end{equation}
        \vspace{1em}
        \annotate[yshift=-0.8em]{below}{missingness}{Missingness process}
        \annotate[yshift=-0.4em]{below}{comp}{Standard deviation}
        \annotate[yshift=0.4em]{above}{error}{Variance of unobserved data}

    where the missingness process is represented through (i) $\rho_g = \text{Corr}(O, X \mid G = g)$, the unobserved correlation between the observation indicator and the ground truth covariate values, and (ii) $\alpha_g = \mathbb{E}[O \mid G = g]$, the observation rate in group $g$, which is observable. Other values impacting the reconstruction error are reflective of the underlying covariate distribution. This includes $\sigma^2_{X \mid G = g} = \text{Var}(X \mid G = g)$, the ground truth variance of the covariate in the group $g$; and $\sigma^2_{X \mid O = 0, G = g}$, the variance of the unobserved values of this same group.

    Under the same assumptions, one can compute the group-specific reconstruction error of population mean imputation as a function of $B^{group}_{g}$, noted in Equation~\ref{eq:bias:group} as the term in parenthesis (labelled in magenta):    
    \vspace{1.3em}
    \begin{equation}
    \label{eq:bias:pop}
        L^{pop}_{g} = \left(\eqnmarkbox[Mulberry]{group}{B^{group}_{g}} + \eqnmarkbox[Maroon]{penalty}{\mu_{g}^{O} - \mu^{O}}\right)^2 + \eqnmarkbox[BurntOrange]{error}{\sigma^2_{X \mid O = 0, G = g}}
    \end{equation}
    \vspace{0.4em}
    \annotate[yshift=0.8em]{above}{group}{Group imputation bias}
    \annotate[yshift=-0.5em]{below}{penalty}{Difference between group- and population- observed means}
\end{lemma}
Proofs of all lemmas and theorems are provided in Appendix~\ref{app:proofs}.

These reconstruction error expressions highlight the impact that the missingness process has on the reconstruction error under different imputation strategies. 
In both expressions, the reconstruction error from group-imputation ($L_g^{group}$) and the reconstruction error from population-imputation ($L_g^{pop}$), the $B_g^{group}$ term reflects the missingness process. 
These expressions also show that while the variance of the covariate influences the reconstruction error (see $\sigma^2_{X|O=0,G=g}$ and $\sigma_{X|G=g}$), they equally increase reconstruction error under the population and group-specific strategies.

Further, this theorem shows that the population mean reconstruction error is a function of group mean reconstruction error. This expression enables us to directly compare the performance of these two imputation strategies on a given group. We can see that the difference between $L^{group}_{g}$ and $L^{pop}_{g}$ depends on ($\mu_g^O - \mu^O$), which is the difference between the two possible values used for imputation. 

In the following theorem, we compare these two imputation strategies and show that the correlation $\rho_g$ plays a central role in determining which imputation strategy is better. As a reminder, $\rho_g$ is the unobservable correlation between the observation indicators and the ground truth covariate values of $X$, capturing an important dimension of the missingness process.

\begin{theorem}[Comparison of group and population mean imputations' reconstruction error]
    \label{theorem:lessbiaspop}
    The group reconstruction error resulting from group mean imputation is larger than the one resulting from population mean imputation, i.e. $L^{group}_{g} > L^{pop}_{g}$, iff one of the following conditions holds:
    \vspace{1.5em}
    \begin{equation}
         \label{eq:theorem:lessbiaspop}
        \eqnmarkbox[NavyBlue]{miss1}{\rho_{g} \cdot \frac{1}{\sqrt{\alpha_g (1 - \alpha_g)}}} < \eqnmarkbox[OliveGreen]{obs1}{\frac{\mu_g^O - \mu^O}{2\sigma_{X | G}}} < 0 \text{ or } 0 < \eqnmarkbox[OliveGreen]{obs2}{\frac{\mu_g^O - \mu^O}{2\sigma_{X | G}}} < \eqnmarkbox[NavyBlue]{miss2}{\rho_g \cdot \frac{1}{\sqrt{\alpha_g (1 - \alpha_g)}}}
    \end{equation}
    \vspace{1em}
    \annotatetwo[yshift=0.5em]{above}{miss1}{miss2}{Missingness process}
    \annotatetwo[yshift=-0.5em]{below, label below}{obs1}{obs2}{Distribution characteristics}
\end{theorem}

Note that this is equivalent to $B^{group}_{g} < \frac{1}{2}  (\mu^O - \mu_g^O) < 0$, and $0 < \frac{1}{2} (\mu^O -\mu_g^O) < B^{group}_{g}$, respectively. In other words, $L^{group}_{g} > L^{pop}_{g}$ if $|B^{group}_{g}| > \frac{1}{2}|\mu^O_g - \mu^O|$.

Theorem~\ref{theorem:lessbiaspop} explicitly identifies the conditions under which group imputation worsens the reconstruction error for a given group $g$ in comparison to population imputation. While $\alpha_g$,  $\mu_g^O$, and $\mu^O$  are observable, Theorem~\ref{theorem:lessbiaspop} shows that whether group imputation is better or worse than population imputation depends on the unobservable $\rho_g$.

To understand this, consider the two MAR scenarios (S1) and (S2), introduced in Section~\ref{sec:missingness}. If the missingness process follows (S1), this corresponds to a scenario in which each subgroup follows a MCAR pattern. The rate of missingness differs for each group, but the missingness is MCAR conditional on group, meaning that $\rho_g=0$. This scenario aligns with the group-specific mean imputation assumption, and consequently, benefits from the use of the group-variant imputation.

However, as we show in the scenario formalisation, (S2) also reflects a MAR pattern but $\rho_g = 0$ is not a reasonable assumption. Consider an example following (S2) in which general practitioners only weigh patients based on a recorded family history of obesity, and lower-income patients ($g$) are more likely to have a family history of obesity. If the patients with such a family history weigh more than those who do not have a family history of obesity, this would lead to a positive $\rho_g > 0$. 

Further, consider a setting where practitioners only record the weights of underweight patients from a group $g$. This represents (S3), where the missingness depends on the missing values themselves (MNAR). In this setting, $\rho_g < 0$. Following Theorem~\ref{theorem:lessbiaspop}, this subsequently means that  $B^{group}_{g}$ is positive. Moreover, in this example the underweight group presents a \textit{smaller} mean weight than the overall population,~i.e., $\mu^O_g - \mu^O < 0$. Thus, $|B^{group}_{g}| > \frac{1}{2}|\mu^O_g - \mu^O|$, meaning that population mean imputation outperforms the group mean imputation in this example.

More generally, consider the case when the observable group mean is larger than the population ($\mu^O_{g} > \mu^O$). In this setting, only the inequality on the right side of Theorem~\ref{theorem:lessbiaspop} can be satisfied. If the missingness process, characterised by $\rho_g$ and scaled by $\alpha_g$, is sufficiently positive and pronounced relative to the normalised difference in means, then population mean imputation results in smaller reconstruction errors than group imputation. This is because the missing values are, on average, smaller than the observed group mean when the correlation is positive. Thus, the missing values are closer to $\mu^O$ than $\mu^O_{g}$, resulting in the superiority of population mean imputation ($L^{group}_{g} > L^{pop}_{g}$). Interestingly, the more positively pronounced the quantity linked with the missingness process, i.e. $B^{group}_{g}$, is, the bigger difference there can be between the two imputation strategies, with population imputation still being superior. On the other hand, if the correlation $\rho_g$ is negative, the missing values are, on average, larger than both the observed group and population means and, therefore, consistently closer to the group mean; group imputation would then always be better in the case of negative correlation. (Similar reasoning follows when $\mu^O_{g} < \mu^O$.) 

Thus, Theorem~\ref{theorem:lessbiaspop} shows that group mean imputation can actually lead to a larger reconstruction error than the population mean imputation for group $g$, i.e. $L^{group}_{g} > L^{pop}_{g}$. In the next theorem, we investigate a follow-up question that considers the fairness gap between groups, and whether group imputation may worsen this gap, i.e., $\Delta^{group}_g > \Delta^{pop}_g > 0$. 

\newpage
\begin{theorem}[Comparison of group and population mean imputations' fairness gaps]
    \label{theorem:lessunfair}
    Under the simplifying assumptions $\sigma^2_{X \mid  \neg O, G} = \sigma^2_{X \mid  \neg O, \neg G}$, and $\mu^O_{g} > \mu^O$, both imputation strategies penalise the marginalised group and the reconstruction gap is larger for the group imputation than the population one (i.e., $\Delta^{group}_g > \Delta^{pop}_g > 0$) iff: 
    \begin{align*}
        &\begin{cases}
            \rho_g \cdot \sigma_{X | G} \cdot f(\alpha_g, r_g, \alpha_{\neg g}) + \rho_{\neg g} \cdot \sigma_{X | \neg G} \cdot f(\alpha_{\neg g}, 1 - r_g, \alpha_{g}) > ((1 - r_g) \alpha_{\neg g} - r_g \alpha_g) (\mu_{g} - \mu_{\neg g})\\
            \rho_g \cdot \sigma_{X | G} \cdot e(\alpha_g) - \rho_{\neg g} \cdot \sigma_{X | \neg G} \cdot e(\alpha_{\neg g}) > \mu_{g} - \mu_{\neg g}\\
            \rho_g \cdot \sigma_{X | G} \cdot h(\alpha_g, r_g, \alpha_{\neg g}) + \rho_{\neg g} \cdot \sigma_{X | \neg G} \cdot h(\alpha_{\neg g}, 1 - r_g, \alpha_{g}) > ((1 - r_g) \alpha_{\neg g} - r_g \alpha_g) (\mu_{g} - \mu_{\neg g})\\
        \end{cases}\\  
        \text{or}&\\
        &\begin{cases}
            \rho_g \cdot \sigma_{X | G} \cdot f(\alpha_g, r_g, \alpha_{\neg g}) + \rho_{\neg g} \cdot \sigma_{X | \neg G} \cdot f(\alpha_{\neg g}, 1 - r_g, \alpha_{g}) > ((1 - r_g) \alpha_{\neg g} - r_g \alpha_g) (\mu_{g} - \mu_{\neg g})\\
            \rho_g \cdot \sigma_{X | G} \cdot e(\alpha_g) - \rho_{\neg g} \cdot \sigma_{X | \neg G} \cdot e(\alpha_{\neg g}) < \mu_{g} - \mu_{\neg g}\\
            \rho_g \cdot \sigma_{X | G} \cdot h(\alpha_g, r_g, \alpha_{\neg g}) + \rho_{\neg g} \cdot \sigma_{X | \neg G} \cdot h(\alpha_{\neg g}, 1 - r_g, \alpha_{g}) < ((1 - r_g) \alpha_{\neg g} - r_g \alpha_g) (\mu_{g} - \mu_{\neg g})
        \end{cases} 
    \end{align*}        
    with $r_g = \mathbb{P}[G = g]$, the ratio of the population belonging to group $g$, $f(\alpha_g, r_g, \alpha_{\neg g}) = \frac{2\alpha_{\neg g} (1-r_g)}{\sqrt{\alpha_{g}(1-\alpha_{g})}}  - \sqrt{\frac{1 - \alpha_{g}}{\alpha_{g}}} \cdot (\alpha_{\neg g} (1 - r_{g}) - \alpha_g r_g)$, $e(\alpha_g) = \sqrt{\frac{\alpha_{g}}{1-\alpha_{g}}}$, and $h(\alpha_g, r_g, \alpha_{\neg g}) = \frac{\alpha_{g}r_g + \alpha_{\neg g} (1 - r_g)}{\sqrt{\alpha_{g}(1-\alpha_{g})}} - \sqrt{\frac{1 - \alpha_{g}}{\alpha_{g}}} \cdot (\alpha_{\neg g} (1 - r_{g}) - \alpha_g r_g)$.
\end{theorem}

This result demonstrates that there exist settings in which standard group-specific imputation strategies, recommended to practitioners, can harm rather than improve algorithmic fairness. Consider a group $g$ negatively impacted by imputation, where imputation increases its reconstruction error in comparison to the rest of the population ($\Delta_g > 0$). Under this consideration, Theorem~\ref{theorem:lessunfair} proves that there exist settings where using population imputation reduces the absolute fairness gap $|\Delta_g|$ compared to group imputation. Figure~\ref{app:fig:satisfy} in the Appendix visualizes areas in which these inequalities are satisfied.

Importantly, this theorem illustrates that population imputation can present better fairness properties under non-MCAR missingness processes. By considering a more general setting, one that incorporates the possibility of non-zero values of $\rho_g$, this result contradicts previous research by \cite{jeong2022fairness} recommending the use of group-specific imputation when the differences between group means are large.

Taken together, the findings in this section question the assumed benefits of group-specific imputation on algorithmic fairness under complex missingness patterns. While practitioners often aim to reduce the reconstruction error gap by controlling on group membership, Theorems~\ref{theorem:lessbiaspop} and~\ref{theorem:lessunfair} show there are settings where this practice may either increase the fairness gap or further harm the population they try to protect.

\section{Empirical evidence of the impact of imputation on algorithmic fairness}
\label{sec:synthetic}

\begin{figure}[t!]
     \centering
     \includegraphics[width=\textwidth]{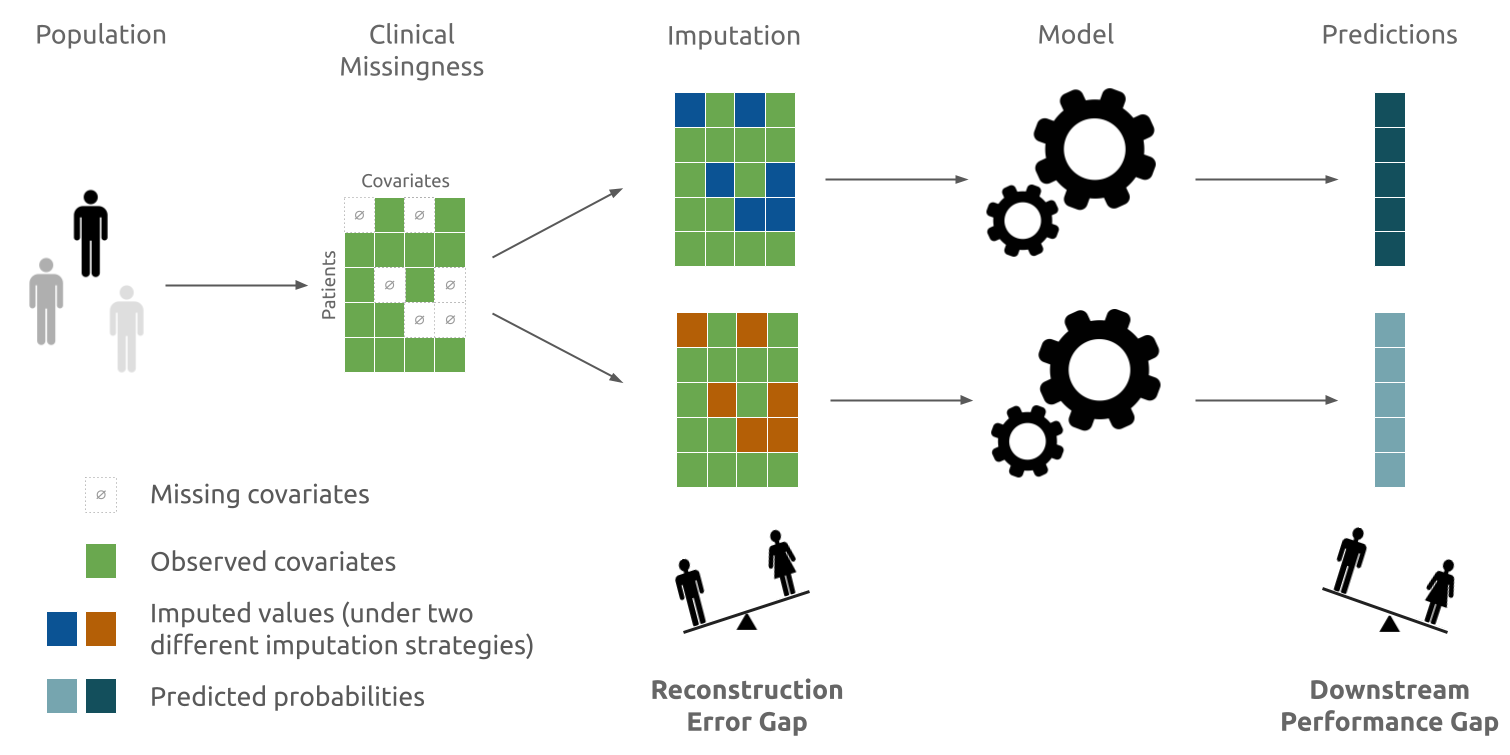}
     \caption{Impact of different imputation strategies on algorithmic fairness, given a population marked by group-specific missingness patterns. This paper measures algorithmic fairness at two levels: (i)~imputation, i.e., how different imputation strategies impact the quality of the reconstructed data for different groups, (ii)~prediction, i.e., how different imputation strategies impact the downstream gap in performance.}
     \label{fig:pipeline}
 \end{figure}

The previous sections identify group-specific patterns of clinical missingness and show that these may translate into disparities in reconstruction errors. Furthermore, the theoretical analysis shows that when selecting between population mean imputation and group mean imputation, the strategy that minimises reconstruction error gaps depends on the missingness process. This process is typically unknown in practice, rendering blanket recommendations counterproductive. In this section, we use simulations---which offer control over the missingness process while allowing us to study more complex imputation strategies---to assess whether the key takeaways from our theoretical analysis also hold true when we consider other commonly used, more advanced imputation strategies. Furthermore, practitioners developing ML models often aim to improve downstream predictive performance, and in such cases improving data quality is only a means to an end. While it is typically assumed that improving data quality is a sufficient and necessary condition to improve downstream predictive performance, these are two distinct types of errors, as illustrated in Figure~\ref{fig:pipeline}. Hence, we also use the simulations to investigate the impact of imputation on downstream predictive performance, and its interplay with reconstruction error. 

\subsection{Data generation}
\label{sec:simulation}
\begin{figure*}[!t]
    \centering
    \includegraphics[width =0.8\linewidth]{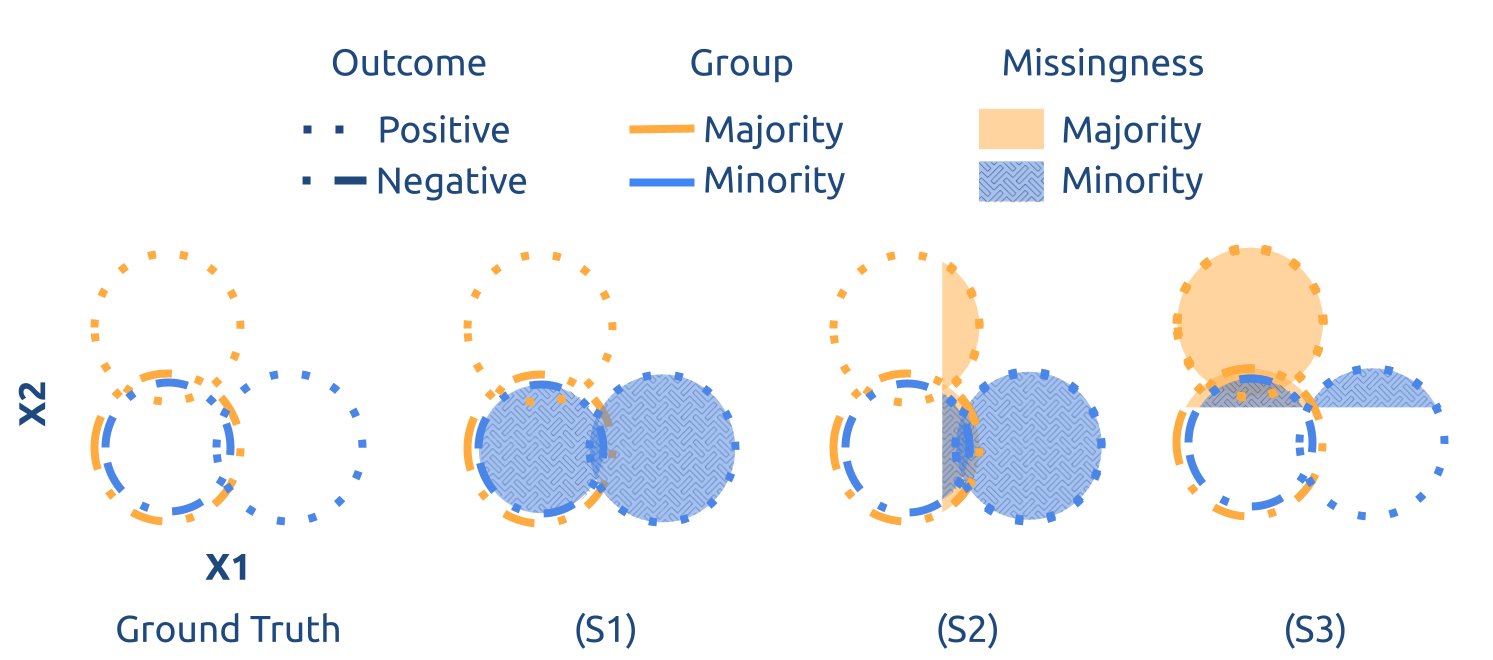}
    \caption{Graphical summary of clinical missingness in the simulation experiments. Missingness is enforced on $X_2$, affecting 50\% of the shaded regions for the indicated group.}
    \label{fig:summary}%
\end{figure*}

Our experiments rely on a population of $N$ patients with associated covariates $X$, marginalised group membership $G$, and a target label $Y$, corresponding to a medical condition. Assume a simulated population consisting of a bi-dimensional covariate set ($X \in \mathbbm{R}^{2}$), with $N = 101,000$ individuals divided into two groups ($G \in \{0, 1\}$), and consider the marginalised group ($G = 1$) is a minority in the population with $1,000$ patients, while the majority is comprised of $100,000$ patients. We assume the two groups differ in condition manifestation, i.e., positive cases across groups differ in how they express the condition in the covariates $X$. Both groups present the same condition prevalence, with $2/3$ of the population presenting the condition. To enforce a difference in condition manifestation, negatives ($Y = 0$) are drawn from a shared bi-dimensional normal distribution, while patients affected by the condition are sampled from two different normal distributions, depending on their group membership. This simulation therefore consists of three clusters illustrated in Figure~\ref{fig:summary} (Ground Truth), and the associated predictive task is to classify between positive and negative cases.  

We then enforce the clinical missingness patterns introduced in Section~\ref{sec:missingness}, by masking covariates on the second dimension $X_2$. The proposed missingness processes reflect the causal graphs introduced in Figure~\ref{fig:dags}. In (S1), the missingness process has a direct dependence on $G$, while in (S2) and (S3) the dependence is indirect, mediated by different condition manifestations across groups. Additionally, we consider a mixture of the previous scenarios to reflect real-world settings in which the missingness process is likely a combination of the identified missingness patterns. Figure~\ref{fig:summary} provides a visual summary of the missingness patterns enforced on the synthetic data. Note how the different scenarios result in group-specific missingness patterns affecting group-specific clusters. 

Formally, the missingness process for each simulation scenario is detailed below. We assume that 50\% of the data in the second dimension is removed in a given subgroup to enforce the three clinical presence scenarios. In the Mixture (M) simulation, the $\land$ stands for the logical `and'---covariate $X_2$ is observed in (M) only if the three other missingness patterns would lead to its observation.
\begin{itemize}
    \item Limited access to quality care (S1):  $O_2^{(S1)} \mid [G = 1] \sim \text{Bernoulli}(0.5)$
    \item (Mis)-informed collection (S2): $O_2^{(S2)} \mid [X_1 > 0.5] \sim \text{Bernoulli}(0.5)$
    \item Confirmation bias (S3): $O_2^{(S3)} \mid [X_2 > 0.5] \sim \text{Bernoulli}(0.5)$
    \item Mixture (M): $O_2^{(M)} = O_2^{(S1)} \land O_2^{(S2)} \land O_2^{(S3)}$
\end{itemize}

 See Appendix~\ref{annex:data_simulation} for full data generation protocol. As an additional robustness check on our findings, in Appendix~\ref{annex:sensitivity}, we explore several adjustments to our synthetic experiments. This includes a lower condition prevalence in the population, correlated covariates $X_1$ and $X_2$, increased noise in the covariate distributions, a larger-sized marginalised group relative to the majority group, and lastly, the condition having the same symptom manifestation across groups but different prevalence between groups. In each of these robustness analyses, we find results that corroborate our core insights.

\subsection{Handling Missingness}
\label{subsec:handling}

We consider an array of imputation strategies widespread in practice. For each approach, we consider its population-level variant and its group-specific counterpart, for a total of eight different strategies considered. 
\paragraph{Mean imputation.} Missing data are replaced by the mean of each covariate, as reviewed in~\cite{donders2006gentle}. The \emph{population} variant (Mean) computes the mean over the whole population, while the \emph{group} variant (Group Mean) computes it over the members of the same group. 

\paragraph{Hot Deck.} Missing data are replaced with values of the most similar patient~\cite{roth1994missing}, defined as the one with the smallest euclidean distance computed on the observed covariates. The \emph{population} variant (Hot Deck) considers the most similar patient over the entire population, while the \emph{group} variant (Group Hot Deck) only considers members of the same group.

\paragraph{Multiple Imputation using Chained Equation (MICE).} Missing data are iteratively drawn from a regression model built over other available covariates after median initialisation. This approach is repeated $I$ times with an associated predictive model for each imputed draw. At test time, the same imputation generates $I$ imputed points for which models' predictions are averaged. MICE is recommended in the literature~\cite{janssen2010missing, newgard2007advanced, wood2004missing, zhou2001multiple, white2011multiple}, based on the argument that it quantifies the uncertainty associated with the missingness process. In the experiments, we used 10 iterations repeated 10 times, resulting in $I = 10$ datasets with associated predictive models. This process describes the \emph{population} variant (MICE). 
Because this strategy assumes a MAR mechanism, \cite{haukoos2007advanced} recommend the addition of potentially informative covariates to make such assumption more plausible. In our experiment, we therefore adjust the regression to include group membership as a covariate, yielding the group variant (Group MICE).

\paragraph{MICE Missing.} Encoding missingness has been shown to improve performance when the patterns of missingness are informative~\cite{groenwold2020informative, lipton2016directly, saar2007handling, sperrin2020missing}. As clinical missingness can contain informative patterns~\cite{jeanselme2022deep, lipton2016directly}, we concatenate missingness indicators to the imputed data. Applying such an approach to the output of MICE yields the population variant (MICE Miss), while applying it to the output of Group MICE yields the group-specific variant (Group MICE Miss). 

\subsection{Results}
\label{sec:results}

We assess the impact of imputation on reconstruction error as well as downstream predictive performance by considering 100 repetitions of each of the proposed simulations.

\begin{figure*}[ht!]
    \centering
    \includegraphics[height = 140px]{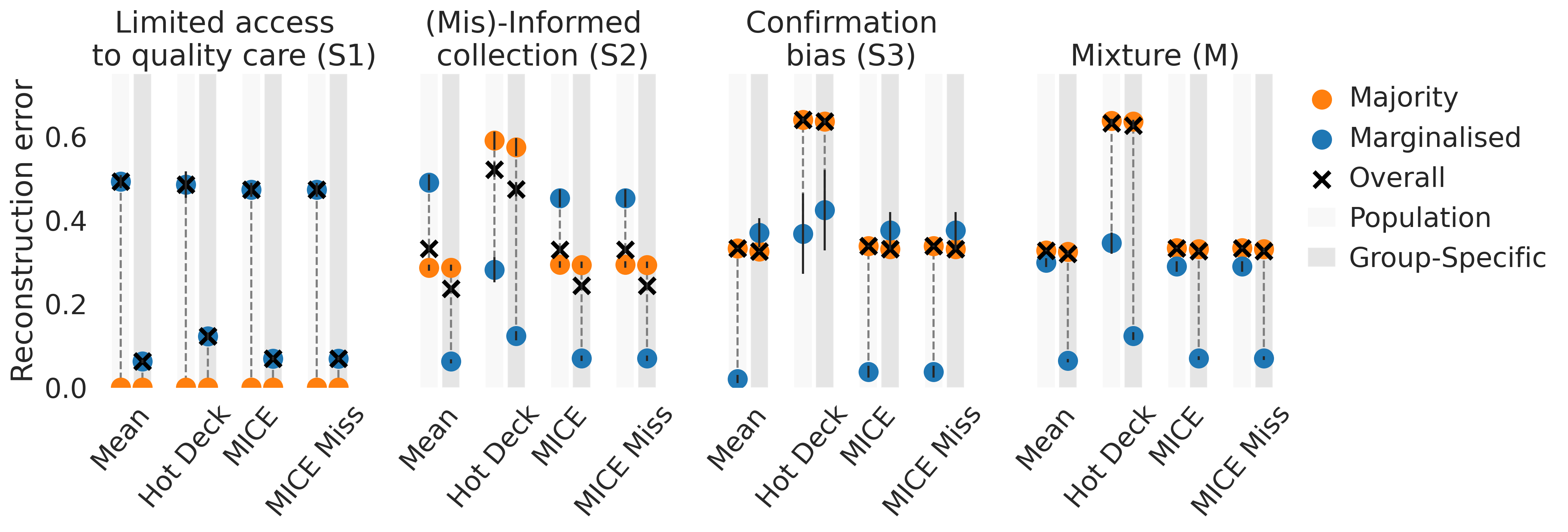}
    \caption{Impact on reconstruction error: Group-specific reconstruction errors across scenarios on 100 synthetic experiments for each missingness pattern. Lower reconstruction error is better.}
    \label{fig:reconstruction_error}
\end{figure*}

As the underlying distribution and missingness process are known, we can compute the reconstruction error gaps. Figure~\ref{fig:reconstruction_error} presents the overall reconstruction errors as well as the reconstruction errors for each group. A larger gap between orange and blue points, representing different groups, indicates a larger bias in reconstruction error. The results are presented for each imputation strategy across the different clinical missingness scenarios, with the group alternatives marked by the darker shade of grey. We discuss these reconstruction results alongside the downstream performance results in the following subsection \ref{sec:results:simulation}.

To evaluate the impact of imputation strategies on downstream predictive performance, we consider the prediction task of differentiating between positive and negative cases, and train a logistic regression model. Note that we explore a single model since our goal is not to quantify how different prediction models may mitigate disparities in data quality; instead, we want to assess the downstream impact of imputation strategies on prediction. The choice of logistic regression is driven by the low dimensionality of our simulations---in our real-world experiments in Section~\ref{sec:mimic} we also consider more complex models.

\begin{figure*}[!ht]
    \centering
    \includegraphics[height = 140px]{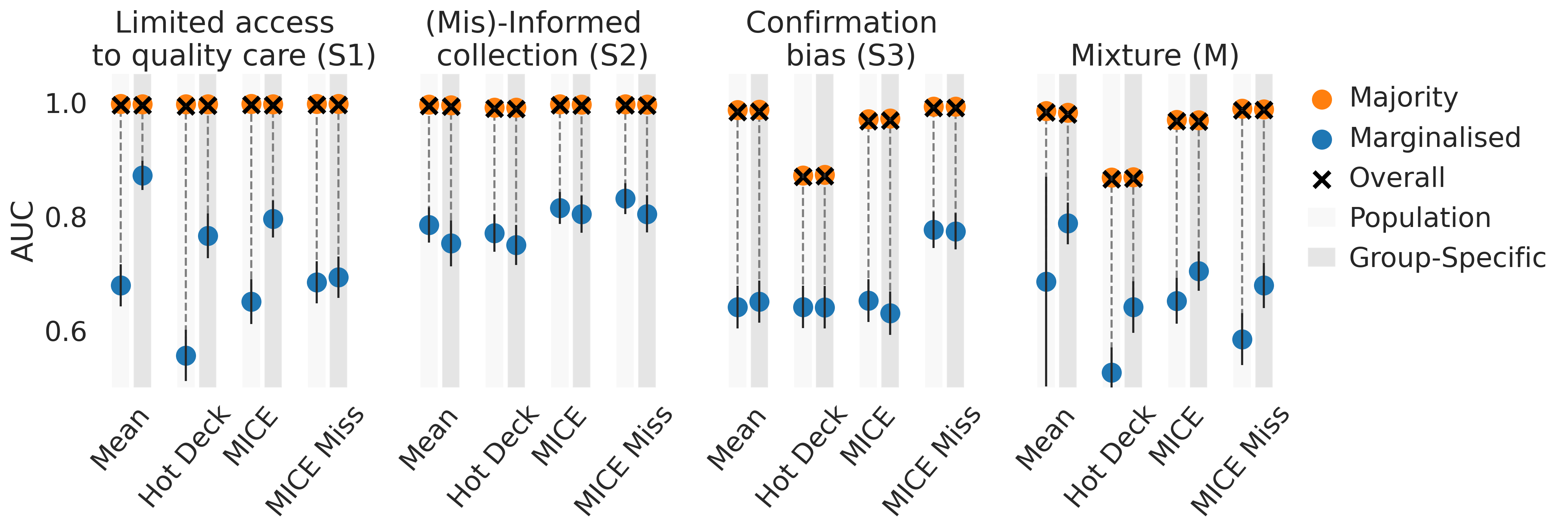}
    \caption{Impact on downstream predictive performance: Group-specific AUC across scenarios on 100 synthetic experiments for each missingness pattern. A higher AUC is better.}
    \label{fig:summary:synth}
\end{figure*}

Analogous to the quantification of algorithmic fairness in reconstruction error, we adopt the \emph{equal predictive performance across groups} definition of algorithmic fairness (\cite{rajkomar2018ensuring}, see Definition~\ref{def:equal}). We use the Area Under the Curve for the Receiver Operating Characteristic curve (AUC), i.e., metric $d$ in Definition~\ref{def:equal} from~\cite{rajkomar2018ensuring}. The AUC measures each group's discriminative performance and is commonly used as a measure of algorithmic fairness in machine learning for healthcare~\cite{larrazabal2020gender, roosli2022peeking, pmlr-v174-zhang22a}. Figure~\ref{fig:summary:synth} presents the group-specific AUCs. A smaller distance between group-specific discriminative performance corresponds to a fairer model. These results are computed on a 20\% test set and averaged over the 100 simulations.  

\subsection{Analysis}  
\label{sec:results:simulation}

Together, the simulation results illustrate the impact of different imputation strategies on reconstruction error and downstream predictive performance, and the connections between the two. The following practical insights emerge. 

\paragraph{Consider both population and group-specific imputation variants, because group-specific can harm marginalised groups despite the aim to improve performance for them.} 
In our theoretical results, we demonstrated that group mean imputation does not necessarily yield better reconstruction error, nor smaller reconstruction error gaps across groups, when compared to population mean imputation. Our simulations extend these insights in ways that have important managerial implications. First, the simulations show that these insights hold when considering more advanced imputation strategies. Consider the reconstruction errors produced by MICE or MICE Miss, shown in Figure~\ref{fig:reconstruction_error}. Under (M), the population variant yields almost equal errors across groups whereas the group-variant significantly reduces errors for the marginalised group. Under (S3), however, this pattern is inverted: the group-specific variants yield almost identical reconstruction errors across groups, whereas the population variant significantly reduces the reconstruction error for the marginalised group. These results show that the relationship between population and group variants is not consistent across clinical missingness patterns. In other words, in some cases, a group-variant may be desirable, while in other cases, the population-wide approach may be preferable. Second, our simulations complement the theoretical results by showing that analogous insights also hold true when considering downstream predictive performance as the measure of interest. 
Figure~\ref{fig:summary:synth} shows that group imputation strategies present \emph{larger} performance gaps than their population variants in (S2). Additionally, whether they yield better or worse performance for the marginalised group varies across scenarios. Together, these results show that blanket recommendations favouring group-specific imputation are misguided.

\paragraph{To improve algorithmic fairness, compare properties of downstream predictive performance.}
It may be tempting to assume, for example, that relying on a more advanced imputation strategy is a sensible choice. Or relying on one with strong theoretical guarantees. Whatever the rationale underlying a choice is, it is very often the case that practitioners rely on a single imputation strategy without exploring the impact of this choice~\cite{nijman2022missing}. Our simulations show this practice to be suboptimal with respect to both reconstruction error and downstream predictive performance. 
For instance, consider a practitioner relying on the more advanced and recommended MICE Miss imputation strategy, following practical recommendations~\cite{wood2004missing}. Figure~\ref{fig:summary:synth} shows that which imputation strategy is better in terms of downstream predictive performance varies widely across missingness patterns. For example, under (S3), MICE Miss provides the best AUC for both groups and the smaller gap in performance; however, under (S1) and (M), Group Mean yields equal AUC for the majority, better AUC for the marginalised group and smaller gaps in performance. Hence, while MICE Miss is the appropriate choice in one setting, Group Mean is preferable in the others. 
Crucially, these specific patterns cannot be expected to remain stable across datasets or other missingness patterns. Given a dataset, the optimal imputation strategy depends on the nature of clinical missingness, which may be unknown in practice.
Hence, there is no imputation strategy that should be consistently preferred.

Moreover, improved reconstruction errors are neither necessary nor sufficient to improve downstream predictive performance and its associated fairness properties. This is demonstrated by comparing reconstruction error in Figure~\ref{fig:reconstruction_error} and downstream predictive performance in Figure~\ref{fig:summary:synth}. 
For instance, in missingness pattern (S2), MICE and MICE Miss yield the best downstream predictive performance, both in terms of group-specific AUC and in terms of gap in AUC across groups. However, looking at reconstruction error presents a very different picture of performance; when we compare MICE and MICE Miss to their group-specific counterparts, these approaches yield significantly worse reconstruction errors for the marginalised group, going as far as to invert the directionality of the gap in reconstruction error. The results for (M) also echo this finding; while Mean, MICE and MICE Miss have comparable performance in terms of reconstruction error, their downstream predictive performance is notably different, with fluctuations affecting the marginalised group.  
Consequently, optimisation of accuracy or algorithmic fairness at the reconstruction error level does not necessarily translate to downstream benefits. \\

Together, these two recommendations indicate that practitioners should not guide imputation choices based on blanket recommendations. Moreover, the results highlight that improving reconstruction error---which is the central aim of methodological developments in imputation---does not translate to better downstream fairness performance. Intuitively, it easy to see why this may happen, as predictive performance and fairness are influenced by multiple factors. For instance, two methods yielding the same reconstruction error may result in different distributions, and their corresponding separability for a condition of interest may differ. Additionally, individuals in a marginalised group may indirectly benefit from errors in imputation if the predictive relationships that an algorithm relies on are driven by the majority population. The insights deriving from our theoretical and empirical evidence motivate the framework we propose in the next section.

\section{Fairness-Aware Imputation Evaluation and Reporting}
\label{sec:framework}
Our theoretical and empirical results demonstrate that common imputation practices are inadequate for the improvement of reconstruction error, as well as for downstream performance and fairness across groups. No strategy consistently outperforms others across different clinical missingness processes. The reliance on a single imputation strategy may hamper performance and result in avoidable unfairness. Further, our results show that improving reconstruction error is neither sufficient nor necessary to improve downstream predictive performance. In prediction settings where data quality is only a means to an end, focusing on minimising reconstruction error can be distracting.

Thus, while these findings question current practices, they also highlight a path forward. In this section, based on our initial findings, we propose a reporting and empirical evaluation framework for imputation selection. Under the outlined theoretical assumption, which can be tested empirically, the proposed empirical comparison of imputation strategy can better inform imputation choice.
Moreover, transparent reporting is central to communicating the risk and the expected settings associated with a model to practitioners who deploy it. Reporting of machine learning transparency and fairness is increasingly being recommended and used in high-profile settings. However, there is a lack of reporting guidelines on handling missingness, despite its critical impact on performance and algorithmic fairness. We address this gap by proposing a reporting framework for imputation.

\subsection{Empirical testability for imputation selection}
\label{sec:compare}

Current machine learning practice typically involves model selection---a process in which a predictive model is chosen among a set of possibilities by empirically comparing their performance. The validity of this selection procedure relies on an assumption about the data distribution: it is assumed that the data encountered during deployment will be sampled from the same distribution as the data used to train and evaluate the models. We extend this assumption to explicitly account for the missingness process.
In other words, we assume that the data on which a model is deployed follows the same data distribution, including the missingness distribution, as the data used for development. Formally, this assumption can be stated as follows:

\begin{assumption}[Stable data generation]\label{assumption:stability}
     The joint distribution of covariates, condition, and missingness processes between development and deployment remains the same. For the random variables associated with the covariates $X$, the condition $Y$, and the observation mask $O$: $$P_{dev}(O, X, Y) = P_{dep}(O, X, Y)$$ with $P_{dev}$ and $P_{dep}$, the development and deployment distributions.
\end{assumption}

With this assumption in place, it is possible to empirically compare downstream predictive performance and associated fairness properties of different imputation strategies. Through this approach, the preferable imputation strategy for a given prediction task can be chosen based on its downstream properties. This imputation selection procedure derives the same advantages as model selection, enabling a data-driven selection of the best imputation strategy. 

Furthermore, this assumption on the joint distribution can be empirically validated, as the distribution of the development data and deployment data can be studied and compared. The literature on distribution shifts offers multiple strategies to detect shifts in this distribution~\cite{rabanser2019failing} to validate the alignment between development and deployment distributions. In healthcare settings, the standardisation of diagnosis and treatment protocols ensures consistency in patient management and, consequently, renders this assumption particularly plausible within a given population and healthcare system.

Formally, we propose the systematic evaluation and comparison of different imputation strategies on fairness metrics. For predictive model $\mathcal{M}$, consider a series of imputation strategies $\mathcal{I} \in \Omega$ to compare, resulting in trained models $\mathcal{M} \circ \mathcal{I}$. The evaluation metric $d$ is being used to evaluate the downstream predictive performance. Lastly, consider different groups being compared for fairness concerns $g \in G$, where $G$ defines the set of considered groups. 
Then, under Assumption \ref{assumption:stability}, practitioners can calculate the overall performance $d(\mathcal{M} \circ \mathcal{I}(X), Y)$, the group-specific downstream performances $d(\mathcal{M}\circ \mathcal{I}(X_{G = g}), Y_{G = g})$, and the fairness gaps between groups. These metrics of interest can then be compared across different imputation strategies in $\Omega$. 

Based on the criteria relevant to a certain domain, which may include considerations such as tolerance of certain types of errors, fairness-performance trade-offs, and bounds on the worst-performing group, practitioners can then choose the imputation strategy that yields the most desirable predictive properties. 
Of note, this approach does not require any assumptions about the missingness process and, as noted above, has theoretical guarantees under data distribution assumptions that are empirically testable. This approach is intuitive and easy to implement, but it is not part of current practice, as evidenced by reviews of the literature that note that imputation strategies are rarely reported and their downstream impact on prediction tasks is seldom compared~\cite{nijman2022missing}. In addition to the theoretical justification for empirical evaluation we have introduced in this section, the next section introduces Imputation Cards to facilitate its use in practice and foster rigorous imputation reporting.

\subsection{Imputation Cards}
\label{sec:card}

Transparent reporting is critical to ensure the alignment between development and deployment settings. Multiple private and governmental organisations have called for such documentation, aiming to mitigate risks. Many of these calls and proposals have been specifically motivated by a societal need to curb the risks of algorithmic bias and discrimination. For instance, \cite{ntia2023} invites practitioners to adopt reporting frameworks to simplify comparison and quantify risks across different demographic groups. Similarly, the widely implemented framework Datasheets for Datasets notes that a central role of such documentation is to mitigate risks of amplifying unwanted societal biases~\cite{gebru2021datasheets}. Further, the AI Bill of Rights in the US~\cite{whitehouse2022} and the Algorithmic Transparency Recording Standard in the UK~\cite{uk2024algorithmic} underscore the critical role of reporting in communicating risk to the public and decision-makers.

However, missingness is largely overlooked in existing reporting frameworks. Widely adopted frameworks, such as Model Cards~\cite{mitchell2019model}---referenced in the AI Bill of Rights~\cite{whitehouse2022} and adopted in industry~\cite{google_model_cards}---do not include guidance on missingness. While the TRIPOD framework~\cite{collins2015transparent} encourages practitioners to ``describe how missing data were handled", scant guidance is provided on how to do this, precluding standardisation. The AI Risk Management Framework Playbook~\cite{NIST_AI_RMF}, a multi-year effort by the National Institute of Standards and Technology (NIST) in the US, has only a single mention of missing data in its 142-page manuscript, noting that missing handling should be documented, without further discussion. 
The lack of guidelines is of particular concern in machine learning for health, where there is very limited reporting of missingness handling. For instance, in their review of machine learning papers focusing on clinical applications,~\cite{nijman2022missing} highlights that, among the papers mentioning missingness in the data, only 3\% analyse the impact of imputation choice on their conclusions. Other reviews of the literature in healthcare~\cite{bellou2019prognostic} and information systems~\cite{peng2023handling} have also noted a lack of adequate reporting of how missing data is handled.

To address this gap, we propose \emph{Imputation Cards}, a reporting framework to document the handling of missing data and its implications. We provide guidance for practitioners to explicitly consider the missingness process and report the impact of imputation strategies on downstream performance and fairness. To facilitate its use in conjunction with existing modelling reporting frameworks, we adopt the structure of the widely used Model Cards~\cite{mitchell2019model}. The core axes of reporting, which we further discuss below, are summarised in Figure~\ref{fig:card}. An application of the reporting framework is then presented in Section~\ref{sec:mimic}.

\begin{figure}[!ht]
    \centering
    \begin{boxedminipage}{3in}
        \begin{itemize}
            \item Key Factors.
                \begin{itemize}
                  \item[\labelitemiv] Demographic groups
                  \item[\labelitemiv] Data collection setting
                \end{itemize}
            \item Missingness Process.
                \begin{itemize}
                    \item[\labelitemiv] Known mechanisms
                    \item[\labelitemiv] Potential mechanisms
                \end{itemize}
            \item Descriptive Statistics.
            \item Methods and Metrics.
                \begin{itemize}
                    \item[\labelitemiv] Imputation methods.
                    \item[\labelitemiv] Predictive models.
                    \item[\labelitemiv] Evaluation metrics.
                \end{itemize}
            \item Empirical Evaluation of Imputation.
            \item Recommendations and Caveats.
        \end{itemize}
    \end{boxedminipage}
    \caption{Summary of Imputation Card Reports.}
    \label{fig:card}
\end{figure}

\textbf{Key Factors.} Identify and describe the groups considered, and the setting in which the data was collected. As both of these factors shape the missingness process, and may vary between development and deployment, they should be explicitly reported. 
We recommend considering demographic groups, such as those defined by ethnicity, gender, insurance, or income, as available in the studied dataset. Practitioners should document where the data was collected and through which means; for instance, this documentation should clarify whether data are collected for insurance claims, or for a study with specific inclusion criteria.

\textbf{Missingness Process.} Identify and describe the mechanisms that may influence the missingness process, and how these may differ across groups.
Acknowledge any epistemic uncertainty in the missingness process and differentiate between known and uncertain mechanisms that may impact the missingness patterns. The missingness mechanisms are often hard to establish, but it is valuable to explicitly consider them. The characterisation of missingness patterns presented in Section~\ref{sec:missingness} can guide this step. Additionally, if there are available institutional details that provide complete knowledge on how the missingness process occurs, it may be possible to theoretically establish which mean imputation strategy may be preferable in terms of reconstruction error by applying the theoretical results presented in Section~\ref{sec:reconstruction}. Notably, as established in this paper, this does not circumvent the need to empirically report impact on downstream predictive performance, or empirically evaluate the performance of other imputation methods. 

\textbf{Descriptive Statistics.} Provide descriptive statistics of the dataset to document missing data. For each covariate, report summary statistics such as the mean, standard deviation, range, and associated missingness rates, disaggregated by groups. This quantification is an effective tool to detect any misalignment between development and deployment environments. 

\textbf{Methods and Metrics.} Detail the imputation strategies and predictive models considered at development, and document any rationale informing their selection. In particular, the description of imputation strategies should be thorough for improved reproducibility. As a default recommended set of imputation strategies in clinical settings, we recommend using the commonly used strategies---mean imputation, median imputation, MICE---as well as the non-linear alternative of Hot Deck imputation. Each of these is detailed in Section~\ref{sec:simulation}.

Additionally, specify the metrics used to evaluate performance and algorithmic fairness. Justify the selection of these metrics in relation to the application. For example, in the absence of task-specific metrics or real-world constraints, we recommend using a practitioner's preferred metric, such as AUC for classification tasks or mean squared error for regression tasks, and a measure of calibration. However, in the healthcare setting, tasks often offer more relevant and valuable metrics. For instance, in healthcare settings, we note that the prioritisation of limited resources is often a key consideration for prediction. Thus, the false negative rate (FNR) at a given resource constraint is a more relevant metric. This metric quantified a facility's capability of identifying the patients in need of prioritisation among those with positive states. The corresponding fairness metric of gaps in FNR then measures disparities in prioritisation errors across groups. Alternatively, if over-diagnosis is a key practitioner concern, with false positives having larger harms than false negatives, then the false positive rate (FPR) quantifies this.

\textbf{Empirical Evaluation of Imputation.} Apply the chosen imputation and predictive models, and use the relevant metrics to report performance under the considered strategies. Report disaggregated performance across groups. 
This evaluation reveals the impact of imputation strategies across subgroups. Critically, based on the findings of this paper and Assumption \ref{assumption:stability}, this provides a valuable tool for practitioners to quantify the differences (if any) that different imputation choices have on downstream performance.

\textbf{Recommendations and Caveats.} Summarise key findings of empirical results, and provide recommendations for deployment. Document which imputation strategy was selected and the reasons justifying this choice. For instance, a combination of imputation strategy and model may offer a preferable trade-off between overall performance and algorithmic fairness properties for the considered application. This trade-off is difficult to weigh exactly, however, the purpose of reporting is to reveal the choices more clearly to the stakeholders deploying a system or impacted by it.\\

The proposed Imputation Cards constitutes a reporting framework that addresses the lack of guidance and practical solutions in reporting how missing data is handled. 
To facilitate its usage in practice, 
we provide a Python toolkit\footnote{\url{https://github.com/Jeanselme/ClinicalPresenceFairness}}.
This toolkit guides users though the selection of key elements such as groups, imputation strategies, metrics, performs the empirical evaluation, and produces an Imputation Card.
Next, we illustrate the proposed framework in the commonly studied MIMIC-III dataset.

\subsection{Case study: Short Term Survival Prediction in MIMIC III}
\label{sec:mimic}
This section illustrates how to inform imputation choice using the proposed evaluation and reporting framework, applied on a widely used observational dataset: MIMIC-III. Appendix~\ref{annex:support} presents an additional analysis on the SUPPORT dataset. These analyses confirm that the impact of imputation is more than a theoretical concern and has real-world consequences on downstream performance and algorithmic fairness.

\subsubsection{Task description.}
The machine learning task is to model short-term survival using a patient's initial 24-hours of laboratory tests in an intensive care unit. We use data from the widely studied Medical Information Mart for Intensive Care (MIMIC III) dataset~\cite{johnson2016mimic}. Following data harmonisation ~(as in \cite{wang2020mimic}), we select adults who survived 24 hours or more after admission to the hospital, resulting in a set of 36,296 patients sharing the 67 most-frequent laboratory tests. The goal is to predict short-term survival (7 days after the initial 24-hour observation period --- $Y$) using the most recent value of each laboratory test observed in the first 24 hours of observation ($X$). Short-term survival is a standard task in the machine learning literature~\cite{pmlr-v146-nagpal21a, tsiklidis2022predicting, xu2019association}. In short-term survival prediction, the observed labels are less likely to suffer from group-specific misdiagnosis, and, therefore, disentangles our analysis from potential biases in labelling.

\subsubsection{Informing imputation choice.}
\begin{figure}[hp]
    \centering
    \captionsetup{font=small, labelformat=empty}
    \begin{boxedminipage}{\textwidth}
        \centerline{\textbf{Short-term survival prediction}}
        \vspace{5mm}
        \begin{minipage}[t]{0.475\textwidth}
            \paragraph{\textbf{Key Factors}: }
            \begin{itemize}
                \item Demographic groups: Sex (43.3\% female), ethnicity (Black (7.7\%) vs non-Black) and insurance (public (66.6\%) vs rest).
                \item Data collection setting: All data from Intensive Care Units (ICU) in a teaching hospital in Boston, Massachusetts, USA.
            \end{itemize}
            \paragraph{\textbf{Missingness Process}: }
            \begin{itemize}
                \item Known mechanisms: Standardised procedures following ICU guidelines.
                \item Potential influences: limited access to care, misinformed collection, and confirmation biases.
            \end{itemize}
            \paragraph{\textbf{Descriptive Statistics}: } 
            \begin{itemize}
                \item Range of missingness rate at the end of 24 hours of observation across the different covariates [1.90 - 99.98] with average 60.53\%. The following table presents the results stratified for the different groups.
                \item Percentage of patients with more than 50\% of tests observed: 85.56\%
            \end{itemize}
            \vspace{5mm}
            \footnotesize
            \centerline{
                \begin{tabular}{c | c | c }
                   Groups & Marginalised & Majority \\ \toprule
                   Sex& 60.82 [2.15 - 99.98] & 60.31 [1.70 - 99.99] \\ 
                   Ethnicity& 61.10 [2.62 - 100] & 60.48 [1.83 - 99.98] \\ 
                   Insurance& 60.93 [1.88 - 99.98] & 60.81 [1.93 - 99.99] \\ 
                \end{tabular}}
            \captionof*{table}{Missingness percentage (mean [min - max]) stratified per group. }
        \end{minipage}
        \hfill
        \begin{minipage}[t]{0.475\textwidth}
            \paragraph{\textbf{Methods and Metrics}: }
            \begin{itemize}
                \item Imputation: Mean imputation, Hot Deck, MICE, MICE Missing (using a missingness indicator as input to the model); and their group-specific variants.
                \item Modelling: Logistic regression with l2 penalty on the imputed data.
                \item Metrics: Use of False Negative Rate (FNR) at a 30\% capacity (current threshold of prioritisation) to reflect the percentage of patients that would not be prioritised despite being at risk, both at the population level and stratified by groups.
            \end{itemize}
            \paragraph{\textbf{Empirical Evaluation of Imputation}: } 
            The following figure describes the performance stratified by groups. Overall performance ranges from 28.4\% to 38.0\% FNR, highlighting a large impact of imputation on performance.

            {\footnotesize
            \centerline{
                \begin{tabular}{c | c | c }
                   Groups & Gap Range & Best \\ \toprule
                   Sex&  [-6.42 - 1.87]& -3.73\\ 
                   Ethnicity& [-9.52 - 10.83]& 0.01\\ 
                   Insurance& [-28.6 - -0.81] & -0.81\\  
                \end{tabular}}
            \captionof*{table}{Range FNR performance gaps (in percent) stratified per group.}}
            
            \paragraph{\textbf{Recommendations and Caveats}:} 
            Assuming a stable missingness process and population distribution at deployment, MICE with missingness indicators minimises the number of patients missed across and within each group.
        \end{minipage}
        \vspace{5mm}
        \begin{minipage}{\textwidth}
        \centering
            \includegraphics[height = 140px]{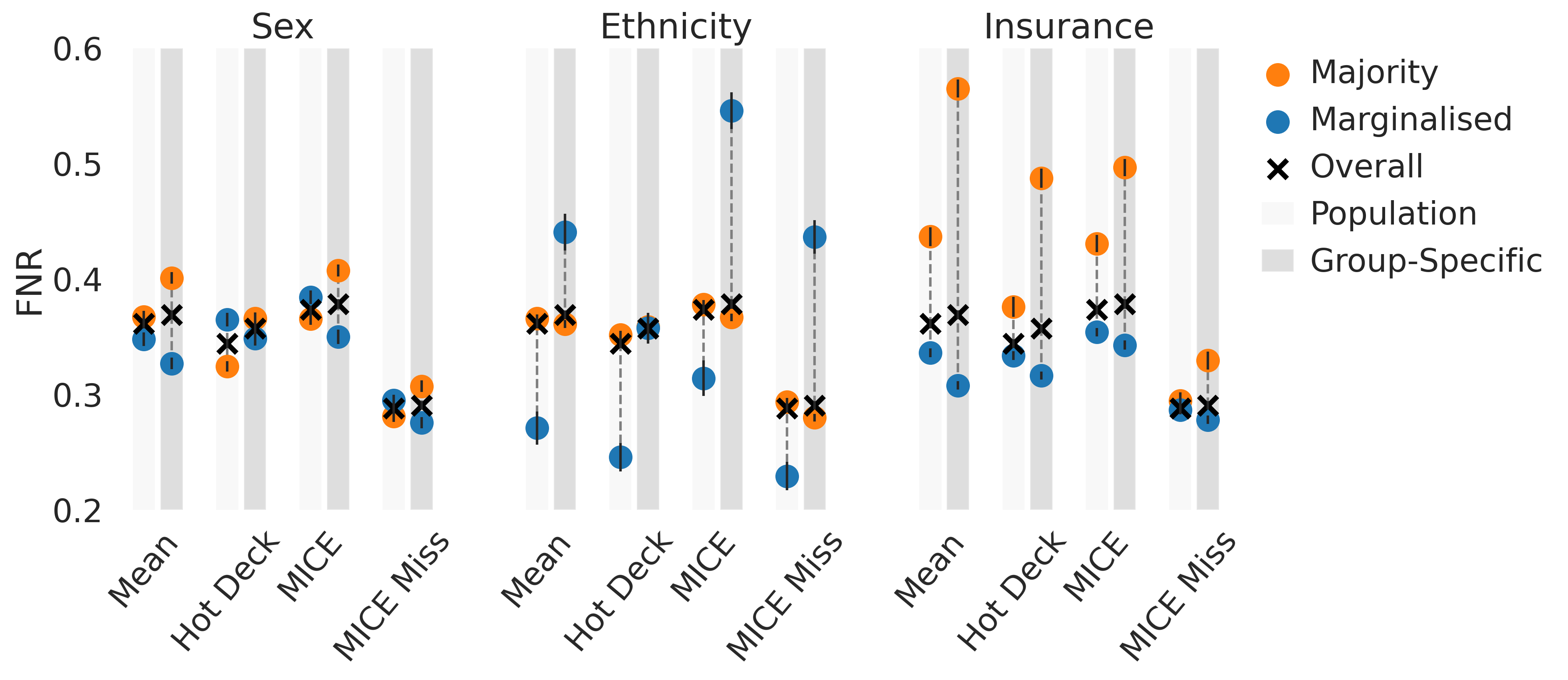}
            \captionof*{figure}{Model performance stratified per group and imputation strategies.}
        \end{minipage}
    \end{boxedminipage}
    \captionsetup{font=normalsize, labelformat=default}
    \caption{Imputation Card for short-term prediction in the MIMIC dataset.}
    \label{fig:card:mimic}
\end{figure}

The following applies the proposed evaluation framework and produces the corresponding Imputation Card to illustrate how practitioners can inform their choice of imputation.
We provide a visual summary of the Imputation Card in Figure~\ref{fig:card:mimic}. 

\paragraph{Key factors.} Based on algorithmic fairness practices and the availability of demographic attributes in the dataset, we focus on groups defined by the following attributes: ethnicity (Black vs non-Black)\footnote{MIMIC's reported ethnicity contains 40 different categories, several referring to Black ethnicities such as Black African American, Black Haitian or Black African; for our analysis, we considered all Black ethnicities as one group.}, sex (female vs male), and insurance (publicly vs privately insured)\footnote{For our analysis, we considered private insurance versus all other types of reported insurance.}. These socio-medical groups may differ in their interactions with the healthcare system. 

\paragraph{Missingness process.} MIMIC data contain observational data from intensive care units in a single teaching hospital in Boston. As a result, the missingness patterns likely follow structured guidelines~\cite{ezzie2007laboratory} and prioritisation rules used in intensive care units. However, evidence exists that these guidelines differ between different groups, e.g. sex-based guidelines~\cite{merdji2023sex} that may result in group-specific patterns. In addition to known guidelines, the ordering of laboratory tests results from experts' intuition and potential biases, which may affect groups differently. Further, historically marginalised patients may differ in their access to care, and consequently have higher rate of missing data. As all these patterns may reflect the Boston hospital's and associated patients' characteristics, practitioners should carefully test for the validity of Assumption~\ref{assumption:stability} at deployment by comparing the joint distribution in the MIMIC and deployment datasets.

\begin{table}[!ht]
    \centering
    \caption{Mean (std) number of orders and observed tests performed during the first post-admission stratified by marginalised groups and outcomes.}
    \label{table:difference}
    \begin{minipage}{.49\textwidth}
        \centering
        \begin{threeparttable}
            \begin{tabular}{rcccc}
                 & \multicolumn{2}{c}{Orders} & \multicolumn{2}{c}{Distinct Tests}\\
                \toprule
                Survived: Yes$^+$                & 5.68 (4.64)  &\multirow{2}{*}{*} & 40.80 (6.73) &\multirow{2}{*}{*}\\
                Survived: ~No$^+$                 & 7.57 (5.44) & & 37.22 (7.50)& \\
                \midrule
                Black                & 5.24 (4.08) &\multirow{2}{*}{*}  & 40.94 (6.94) &\multirow{2}{*}{*}\\
                Other            & 5.86 (4.77)  & & 40.52 (6.84) &\\
            \end{tabular}
            \begin{tablenotes}
              \small
              \item $^+$ By the 8$^{th}$ day after admission.
            \end{tablenotes}
        \end{threeparttable}
    \end{minipage}
    \begin{minipage}{.49\textwidth}
        \centering
        \begin{threeparttable}
            \begin{tabular}{rcccc}
             & \multicolumn{2}{c}{Orders} & \multicolumn{2}{c}{Distinct Tests}\\
            \toprule
            Female               & 5.54 (4.45)  &\multirow{2}{*}{*} & 40.75 (6.89) &\multirow{2}{*}{*}\\
            Male                 & 6.03 (4.91)  & & 40.41 (6.80) &\\
            \midrule
            Public               & 5.67 (4.57) &\multirow{2}{*}{*}  & 40.46 (6.76) &\multirow{2}{*}{*}\\
            Private              & 6.11 (5.01) &  & 40.75 (7.01)&
            \end{tabular}
            \begin{tablenotes}
              \small
              \item $*$ Significant t-test p-value ($<$ 0.001).
            \end{tablenotes}
        \end{threeparttable}
    \end{minipage}
\end{table}

\paragraph{Descriptive statistics.} 
We examine the missing data and identify evidence of group-specific patterns. First, there is 60.5\% of missing data in MIMIC III data, as all possible laboratory tests are not performed during the first 24 hours of observation. Table~\ref{table:difference} details the number of orders and the number of distinct laboratory tests (out of the 67 possible tests) performed during the first 24-hours for different subgroups. We disaggregate the results between patients who died during their stay vs. those who survived, and observe that there is a larger number of orders for patients who die during their stay compared with the ones who survived. 
Doctors may monitor sicker patients, or patients with conditions with higher mortality, more closely. 

When analysing missingness for the groups of interest, we note that there are fewer test orders for female, Black, and publicly insured patients, but little difference in the distinct number of prescribed tests. This difference may be explained by the underlying conditions or other medically relevant factors, which may be different across groups, or it could also be explained by other mechanisms, such as limited access to quality care (S1), mis-informed collection (S2), confirmation bias (S3), or a complex combination of multiple patterns. Importantly, the cause of these differences cannot be established from the observational data. The combination of having similar tests but less frequent observations of these tests results in less up-to-date data on patient health status available for modelling. 
Thus, even though the cause of performance differences is unclear, these observations show the connection between group membership, testing patterns, and outcomes. This real-world evidence of non-random missingness patterns among subgroups of patients raises concerns about increasing inequities if the algorithmic fairness implications of imputation strategies are not considered.

\paragraph{Methods and metrics.} We do not have deployment constraints with respect to the choice of imputation, so we consider all the imputation strategies presented in Section~\ref{subsec:handling}: Mean, Hot Deck, MICE, MICE Missing and their group-specific variants. The group-specific variants control for membership to all considered groups. Specifically, group-specific strategies replace missing values based on the patient's ethnicity, sex, and insurance. If membership to all groups is unaccounted for, the imputation choice at deployment is underdetermined as a patient is member of multiple of these groups. Naturally, one could consider one group at a time. This additional analysis is presented in Appendix~\ref{annex:mimic:onegroup} and results in the same conclusions and similar findings as presented below. 

Patients are split into three sets: 80\% for training, 10\% for hyper-parameter tuning and 10\% for testing. Imputation strategies rely on the training set to impute all missing data. Regression for a missing covariate relies on all other covariates using the same procedures as described in Section~\ref{subsec:handling}. Using these strategies, we impute missing data, resulting in $\Tilde{X}$. Then, our analysis consists of a logistic regression model---a commonly relied upon predictive model in the medical literature~\cite{Nick2007, goldstein2017opportunities}---to discriminate between positive and negative cases ($\text{logit}(Y) \sim \Tilde{X}$). To avoid overfitting, we use a logistic regression with an $l2$ penalty selected among $[0.1, 1, 10, 100]$ on the held-out tuning set. Appendix~\ref{annex:mimic:models} explores the use of alternative predictive modelling strategies. These additional analyses using different predictive models corroborate the robustness of the insights from our case study. Appendix~\ref{annex:mimic:models} also shows that, using a neural network or a decision tree as a predictive model on the MIMIC case study data also reveals the limitations of relying on group-specific imputation and the importance of careful empirical evaluation of imputation choice with a lens on fairness.

Models that predict short-term survival are often considered for the purpose of prioritisation of care~\cite{pirracchio2015mortality}. Deploying such a model could have important consequences on patients' life. The model could inform the care prioritisation of patients with predicted elevated risks. Thus, ensuring equitable prioritisation is central to this problem. As AUC does not directly quantify how deployment can hurt marginalised groups if a hospital were to use a fixed allocation of resources to treat patients (such as beds or staff limits), we evaluate the False Negative Rate (FNR) given the assumed availability of priority care for 30\% of the population. A higher FNR in this setting is worse, meaning more high-risk patients are incorrectly deprioritised. A gap in FNR between groups illustrates differences in the rate of missed patients between groups. (In Appendix~\ref{annex:mimic:threshold}, we provide results under different prioritisation thresholds. In practice, the threshold would be determined by resource constraints).

\begin{figure*}[!ht]
\centering
    \includegraphics[height = 140px]{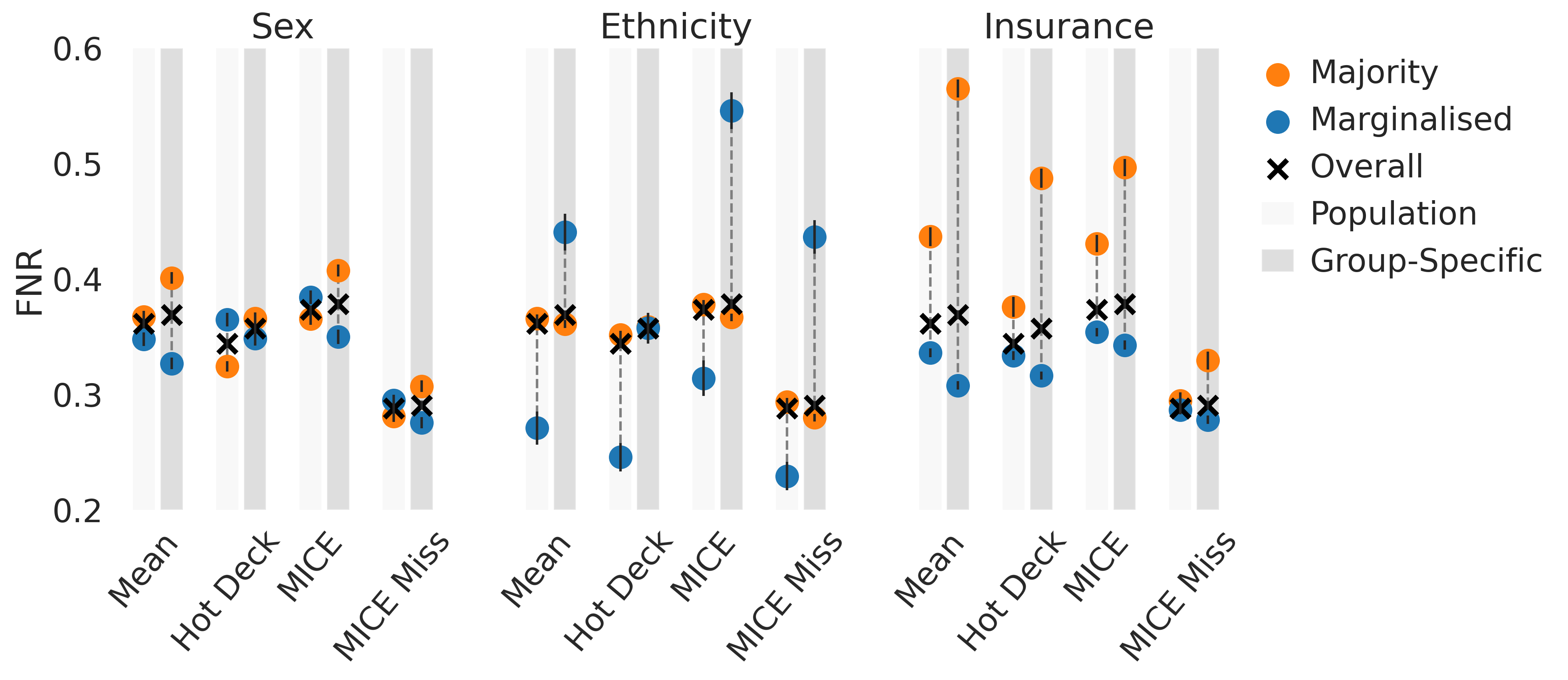}
    \caption{Percentage of non-prioritised high-risk patients (FNR) across marginalised groups in MIMIC III experiment, bootstrapped on the test set over 100 iterations.}
    \label{fig:mimic:fnr}
\end{figure*}

\paragraph{Empirical evaluation of imputation.}
Figure~\ref{fig:mimic:fnr} summarises the impact of each imputation strategy on downstream predictive performance.
First, we see that the MICE Miss group and population variants present the best FNR performance overall in the data. However, while both variants present similar overall FNR, they differ substantially in their group-specific performances. 

The group-specific MICE Miss imputation leads to a gap twice as large as the population variant in the ethnicity split. Furthermore, this gap is not only larger, but group imputation actually \emph{flips} the direction of the fairness gap, relative to the population variant, in a direction that harms the marginalised ethnicity group. This finding from this case study illustrates the surprisingly negative effect that relying on group-specific imputation may have. It also underscores the importance of a careful empirical evaluation of imputation choice with a lens on fairness. 

Additionally, when considering mean imputation in the case of MIMIC-III data, Figure~\ref{fig:mimic:fnr}  shows the group variant has larger (worse) FNR gaps than the population variant when considering sex or insurance. 
Similar to MICE Miss, the two mean imputation variants also surprisingly present opposite algorithmic fairness consequences when considering groups split by ethnicity, with the group-specific mean imputation strategy harming the marginalised ethnicity group.
Group mean imputation results in a larger FNR for Black patients than for non-Black patients. In contrast, population mean imputation yields a smaller FNR for Black patients than for non-Black patients. Our findings remain robust if we consider an alternative predictive model as well. For example, Appendix~\ref{annex:mimic:models} uses a neural network predictive model instead of a logistic regression. Here, the group MICE and group MICE Miss imputation strategies both continue to have larger and opposite-direction fairness gaps than their population equivalents, in directions that harm the marginalised ethnicity group.

These case study observations in real-world data highlight how two imputation strategies, solely differing in their handling of group membership, can either harm or favour the marginalised group's performance relative to the rest of the population. This choice of imputation thus can drastically shift a model's algorithmic fairness properties. 
It also further demonstrates how relying on a single imputation strategy can be harmful, illustrating the value of the Imputation Cards framework.

\paragraph{Recommendations and caveats.} Equipped with the previous analysis and assuming the same data-generation and missingness processes at deployment---a plausible assumption in the ICU context as ordering practices are slow to change as patients are assumed to benefit from current practices~\cite{cismondi2013reducing}---we aim to minimise the number of patients not prioritised despite a critical condition (FNR). In this context, minimisation of group-specific performance prevails over equalisation of performance, provided the worse-off group is not suffering from a further drop in performance. Thus, the population variant of MICE Missing best achieves this goal with the minimal FNR for all, yielding an outcome in which all groups benefit from a performance that is better or equal to what they would have under alternative imputation strategies.

One potential limitation of our analysis is the possible impact of temporal patterns and inconsistencies in laboratory records. For example, delays in recording a lab can occur due to inefficiencies in the healthcare system, leading to observational data that do not fully reflect a patient's condition at a given time. However, note that the data at deployment may also suffer from similar limitations. Thus, under the assumption of a consistent temporal observation process at deployment, our recommendations remain the same. While these limitations are not within the scope of our study, given this potential limitation and to provide a second case study application of our framework, we analyse the real-world SUPPORT dataset in Appendix~\ref{annex:support}. In this data, we explore a similar imputation choice for mortality prediction, but the data is collected as part of a controlled study.  The controlled nature of this study ensures a higher quality of measurements compared to the previous observational study, rendering the problem of timestamp consistency a lesser concern. The results in this dataset corroborate the robustness of our core insights, illustrating, for example, that the choice of imputation strategy between MICE and MICE Missing can flip the direction of the fairness gap.

\clearpage

\section{Discussion}\label{sec:conclusion}
In this section, we provide a summary of our contributions and its implications for practice, and outline promising directions for future research. 

\subsection{Contributions and implications}
The fairness literature has explored how machine learning can reinforce biases present in data. Our work demonstrates how biases may be reflected, not only in what is present, \textit{but in what is absent from the data}. Learning from medical data without sufficient attention to the potential disparities present in clinical missingness can reinforce and automatise inequities, and further harm historically marginalised groups. Consequently, missingness handling should not be considered a disconnected problem but rather an integral part of improving algorithmic fairness. 

In the context of machine learning for health, interactions between patients and the healthcare system can result in group-specific missingness patterns. These patterns may then impact downstream algorithmic fairness under different imputation strategies.
Our review of historical examples from the literature reveals three distinct missingness processes leading to group-specific patterns. 
%
Our theoretical and empirical results challenge the foundations of current imputation practices when considering these realistic patterns. 
Overlooking the importance of imputation on algorithmic fairness has resulted in the current practice of relying on a single strategy, often based on theoretical reduction of error under unrealistic missingness assumptions, which rarely capture the complexity of real-world data.
Our theoretical results demonstrate how the choice between two imputation strategies is underdetermined when faced with an unknown missingness process.
In particular, we demonstrate that the common and recommended practice of controlling for, or stratifying by, group membership can counter-intuitively increase the gap between group-specific performances. 
Finally, we show that minimising reconstruction errors is neither necessary nor sufficient to improve downstream performance and algorithmic fairness properties, rendering current recommendations counterproductive for fairer deployment.

Our case studies using the SUPPORT and MIMIC III datasets demonstrate the relevance of the identified problem as more than a theoretical concern, showing that group-specific missingness patterns are present in widely used medical datasets and that the choice of imputation strategy can drastically impact algorithmic fairness properties of downstream prediction tasks. Critically, two different imputation strategies may result in opposite fairness gaps while presenting similar overall performance. Relying on common imputation practices when developing machine learning models on these datasets could reinforce inequities in the healthcare system that an alternative imputation strategy could have mitigated.

These findings result in a critical managerial recommendation for fairer machine learning deployment: practitioners can and must consider the impact of different imputation strategies on algorithmic fairness. To this end, we introduce a framework to inform and report imputation choices when implementing predictive models in the presence of clinical missingness. The proposed toolkit and its operationalisation in Python allow practitioners to measure the impact of different imputation strategies on downstream performance and algorithmic fairness. By comparing performance following different imputation strategies, practitioners can inform their imputation choice based on performance and algorithmic fairness considerations relevant for their considered task. Finally, to foster transparency and mitigate risks stemming from misalignment between development and deployment settings, we propose a framework for standardised reporting of missingness handling. 

\subsection{Future work}
In future work, there are two axes of potential research that could significantly impact machine learning for health. First, clinical missingness is only one dimension of how clinical presence shapes the data-generating process. The interaction between patients and the healthcare system not only imprints missingness, but it may also shape aspects such as the temporality of medical time series, which may similarly convey group-specific disparities that current machine learning models may amplify.  
Second, our proposed framework is appropriate when the data-generation process remains stable between development and deployment. As noted in the paper, this is a testable assumption. However, if the outcome of the test is that the distribution has shifted, there is a need for guidance that can inform imputation choice under distribution shift while accounting for algorithmic fairness.

\newpage
\section*{Acknowledgements}
The authors acknowledge the partial support of the UKRI Medical Research Council (programme numbers MC\_UU\_00002/5 and MC\_UU\_00002/2 and theme number MC\_UU\_00040/02 – Precision Medicine). Vincent Jeanselme acknowledges the support of The Alan Turing Institute’s Enrichment Scheme. Maria De-Arteaga acknowledges the support of NIH through grant R01NS124642.

\bibliographystyle{unsrt}
\bibliography{bibliography.bib}

\newpage
\appendix

\section{Proofs}
\label{app:proofs}

In this section, we demonstrate the theoretical results introduced in Section~\ref{sec:reconstruction}. 

\paragraph{Notations.}
Consider an imputation strategy $\mathcal{I}$ that replaces missing data with a constant value $c^\mathcal{I}_g$ for the group $g$. Our work analyses $L^\mathcal{I}_g$, the group-specific reconstruction error for imputation $\mathcal{I}$, and $\Delta^\mathcal{I}_g$, the gap in reconstruction error between group $g$ and the rest of the population.

All capital letters stand for random variables: $G$ for group membership, $O$ for the observation process and $X$ for the covariates. Throughout the proofs, we characterise the missingness process with $\rho_g =  \text{Corr}_{P_g}(O, X)$, the unobserved correlation between the observation indicator and the ground truth covariate values, and $\alpha_g = \mathbb{E}[O \mid G = g]$, the observation rate in the group $g$. The covariate $X$ is described with $\mu$ for its mean, $\sigma$ for its variance. Exponent $O$ expresses an observed quantity. Subscript $g$ indicates subgroup characteristics. 

Finally, the population is defined by the ratio of members in group $g$ in comparison to the rest of the population, i.e., $r_g = \mathbb{E}[G = g]$.

\subsection{Lemma \ref{theorem:expression}}
\label{proof:expression}

\paragraph{Lemma.}
    Assuming i.i.d. data points $\{x_i\}$, one can express the reconstruction error in group $g$ resulting from group mean imputation as:
        \begin{equation}
            L^{group}_{g} = \Bigg(\overbrace{\eqnmarkbox[NavyBlue]{missingness}{-\frac{1}{\sqrt{\alpha_g(1-\alpha_g)}} \cdot \rho_g} \cdot
                                \eqnmarkbox[OliveGreen]{comp}{\sigma_{X \mid G = g}}}^{\textcolor{Mulberry}{\bm{B^{group}_{g}}}}\Bigg)^2 + \eqnmarkbox[BurntOrange]{error}{\sigma^2_{X \mid O = 0, G = g}}
        \end{equation}
        \vspace{1em}
        \annotate[yshift=-0.8em]{below}{missingness}{Missingness process}
        \annotate[yshift=-0.4em]{below}{comp}{Standard deviation}
        \annotate[yshift=0.4em]{above}{error}{Variance of unobserved data}

    where the missingness process is represented through (i) $\rho_g = \text{Corr}(O, X \mid G = g)$, the unobserved correlation between the observation indicator and the ground truth covariate values, and (ii) $\alpha_g = \mathbb{E}[O \mid G = g]$, the observation rate in group $g$, which is observable. Other values impacting the reconstruction error are reflective of the underlying covariate distribution. This includes $\sigma^2_{X \mid G = g} = \text{Var}(X \mid G = g)$, the ground truth variance of the covariate in the group $g$; and $\sigma^2_{X \mid O = 0, G = g}$, the variance of the unobserved values of this same group.

    Under the same assumptions, one can compute the group-specific reconstruction error of population mean imputation as a function of $B^{group}_{g}$, noted in Equation~\ref{eq:bias:group} as the term in parenthesis (labelled in magenta):    
    \vspace{1.3em}
    \begin{equation}
        L^{pop}_{g} = \left(\eqnmarkbox[Mulberry]{group}{B^{group}_{g}} + \eqnmarkbox[Maroon]{penalty}{\mu_{g}^{O} - \mu^{O}}\right)^2 + \eqnmarkbox[BurntOrange]{error}{\sigma^2_{X \mid O = 0, G = g}}
    \end{equation}
    \vspace{0.4em}
    \annotate[yshift=0.8em]{above}{group}{Group imputation bias}
    \annotate[yshift=-0.5em]{below}{penalty}{Difference between group- and population- observed means}

\proof{Proof.}
    First, we express the reconstruction error of a constant imputation strategy considering each covariate independently. For clarity, we denote by $\neg O$ if $O = 0$, $\neg G$ if $G \neq g$. 

    \begin{align}
        L^{\mathcal{I}}_{g} :=& \mathbb{E}\left[||c^\mathcal{I}_g - X||_2^2 \Bigm|  \neg O, G \right]\tag*{\textit{(Using Definition~\ref{eq:def})}}\nonumber\\
                  =& \mathbb{E}\left[(c^\mathcal{I}_g - X)^2 \Bigm|  \neg O, G \right]\nonumber\\
                  =& \mathbb{E}\left[X^2\Bigm|  \neg O, G \right] + c^{\mathcal{I}^2} - 2c^\mathcal{I}_g \mathbb{E}\left[X \Bigm|  \neg O, G \right]\nonumber\\
                  =& \mathbb{E}\left[X \Bigm|  \neg O, G \right]^2 + \sigma^2_{X \mid  \neg O, G} + c^{\mathcal{I}^2} - 2c^\mathcal{I}_g \mathbb{E}\left[X \Bigm|  \neg O, G \right] \tag*{\textit{(By definition of variance)}}\nonumber\\
                  =& \left(\mathbb{E}\left[X \mid  \neg O, G\right] - c^\mathcal{I}_g\right)^2 + \sigma^2_{X \mid  \neg O, G} \label{equation:decomposition}
    \end{align}
    
    \begin{remark}
        This first expression demonstrates that the reconstruction error of \emph{any} constant imputation is lower bounded by the variance of the unobserved data: these constant strategies do not capture any of this variance.
    \end{remark}

    Using this decomposition, we further explore how group mean imputation impacts the reconstruction error. As a reminder, group mean imputation replaces missing values in group $g$ with the observed group mean, i.e., $c^{group}_g = \mathbb{E}[X \mid O, G]$. The square root of the first term in Equation~\eqref{equation:decomposition}, that we refer as $B^{group}_{g}$, therefore simplifies to:
    \begin{align*}
        B^{group}_{g} :&= \mathbb{E}[X \mid  \neg O, G] - \mathbb{E}[X \mid O, G] \\
        &= \frac{\mathbb{E}[(1-O)X \mid G]}{\mathbb{E}[(1-O) \mid G]} - \frac{\mathbb{E}[OX \mid G]}{\mathbb{E}[O \mid G]}\\
        &= \frac{\mathbb{E}[X \mid G] - \mathbb{E}[OX \mid G]}{1 - \mathbb{E}[O \mid G]} - \frac{\mathbb{E}[OX \mid G]}{\mathbb{E}[O \mid G]}\\
        &= \frac{-\text{Corr}(O, X \mid G) \cdot \sigma_{O \mid G} \cdot \sigma_{X \mid G}}{(1 - \mathbb{E}[O \mid G])\mathbb{E}[O \mid G]}\tag*{\textit{(By definition of covariance)}}\\
        &= -\rho_g \cdot \sqrt{\frac{1}{\alpha_g (1 - \alpha_g)}} \cdot \sigma_{X \mid G}
    \end{align*}
    
    Similarly, in the context of population mean imputation, which replaces the missing values with the observed mean, i.e., $c^{pop} = \mathbb{E}[X \mid O]$, results in the simplification of the first term as:
    \begin{align*}
        B^{pop}_{g} :=& \mathbb{E}[X \mid \neg O, G] - \mathbb{E}[X \mid O] \\
        =& \mathbb{E}[X \mid \neg O, G] - \mathbb{E}[X \mid O, G] + \mathbb{E}[X \mid O, G] - \mathbb{E}[X \mid O]\\
        =&  B^{group}_{g} + \mu_{g}^{O} - \mu^{O}
    \end{align*}
    
    One can further decompose this equality to disentangle distributions' properties from missingness processes:
    \begin{align}
        B^{pop}_{g} =&B^{group}_{g} + \mathbb{E}[X \mid O, G] - \mathbb{E}[G \mid O]\mathbb{E}[X \mid O, G] - \mathbb{E}[\neg G \mid O]\mathbb{E}[X \mid O, \neg G]\nonumber\\
        =& B^{group}_{g} + (1 - \frac{\alpha_gr_g}{\alpha})\mu_g^O -  \frac{\alpha_{\neg g}(1-r_g)}{\alpha}\mu_{\neg g}^O\tag*{\textit{(By Bayes rule)}}\\
        =& B^{group}_{g} + \frac{\alpha_{\neg g}(1-r_g)}{\alpha} [\mu_g^O - \mu_g + \mu_g - \mu_{\neg g} + \mu_{\neg g} - \mu_{\neg g}^O]\nonumber\\
        =& B^{group}_{g} + \frac{\alpha_{\neg g}(1-r_g)}{\alpha} \left[\rho_g \cdot \sqrt{\frac{1 - \alpha_g}{\alpha_g}} \cdot \sigma_{X \mid G} + \mu_g - \mu_{\neg g} - \rho_{\neg g} \cdot \sqrt{\frac{1 - \alpha_{\neg g}}{\alpha_{\neg g}}} \cdot \sigma_{X \mid \neg G}\right] \label{eq:extendedpop} 
    \end{align}
    
    with $r_g = \frac{|P_g|}{|P|}$, the proportion of patients member of group $g$, and $\alpha$, the overall observation rate, i.e. $\alpha = \alpha_g r_g + \alpha_{\neg g}(1-r_g)$. This last expression is obtained by a decomposition of $\mu_g^O - \mu_g$ similar to the one proposed for the computation of $B^{group}_{g}$.
\endproof

\subsection{Theorem \ref{theorem:lessbiaspop}}
\label{proof:lessbiaspop}

\paragraph{Theorem.}
    The group reconstruction error resulting from group mean imputation is larger than the one resulting from population mean imputation, i.e. $L^{group}_{g} > L^{pop}_{g}$, iff one of the following conditions holds:
    \vspace{1.5em}
    \begin{equation}
        \eqnmarkbox[NavyBlue]{miss1}{\rho_{g} \cdot \frac{1}{\sqrt{\alpha_g (1 - \alpha_g)}}} < \eqnmarkbox[OliveGreen]{obs1}{\frac{\mu_g^O - \mu^O}{2\sigma_{X | G}}} < 0 \text{ or } 0 < \eqnmarkbox[OliveGreen]{obs2}{\frac{\mu_g^O - \mu^O}{2\sigma_{X | G}}} < \eqnmarkbox[NavyBlue]{miss2}{\rho_g \cdot \frac{1}{\sqrt{\alpha_g (1 - \alpha_g)}}}
    \end{equation}
    \vspace{1em}
    \annotatetwo[yshift=0.5em]{above}{miss1}{miss2}{Missingness process}
    \annotatetwo[yshift=-0.5em]{below, label below}{obs1}{obs2}{Distribution characteristics}

\proof{Proof.}
    This inequality can be rewritten as:
    \begin{align*}
        L^{group}_{g} > L^{pop}_{g} 
        \Leftrightarrow& |B^{group}_{g}| > \left|B^{group}_{g} + \mu_g^O - \mu^O\right|
    \end{align*}
    To demonstrate this inequality, we explore the four different cases.
    
    Case 1: $B^{group}_{g} > 0$ and $B^{group}_{g} + \mu_g^O - \mu^O > 0$
    \begin{align*}
        \begin{cases}
            B^{group}_{g} > 0\\
            B^{group}_{g} + \mu_{g}^{O} - \mu^{O} > 0\\
            B^{group}_{g} > B^{group}_{g} + \mu_{g}^{O} - \mu^{O}
        \end{cases}
        \Leftrightarrow\begin{cases}
            \rho_g < 0\\
            \rho_g \cdot \sqrt{\frac{1}{\alpha_g (1 - \alpha_g)}} < \frac{\mu_g^O - \mu^O}{\sigma_{X | G}}\\
            \mu^{O} > \mu_{g}^{O}
        \end{cases}
    \end{align*}
    
    Case 2: $B^{group}_{g} > 0$ and $B^{group}_{g} + \mu_{g}^{O} - \mu^{O} < 0$\\
    \begin{align*}
        \begin{cases}
            B^{group}_{g} > 0\\
            B^{group}_{g} + \mu_{g}^{O} - \mu^{O} < 0\\
            B^{group}_{g} > -B^{group}_{g} - \mu_{g}^{O} + \mu^{O}
        \end{cases}
        \Leftrightarrow\begin{cases}
            \rho_g < 0\\
            \mu^{O} > \mu_{g}^{O}\\
            \rho_g \cdot \sqrt{\frac{1}{\alpha_g (1 - \alpha_g)}} \in [\frac{\mu_g^O - \mu^O}{\sigma_{X | G}}, \frac{\mu_g^O - \mu^O}{2\sigma_{X | G}}]
        \end{cases}
    \end{align*}
    
    Case 3: $B^{group}_{g} < 0$ and $B^{group}_{g} + \mu_{g}^{O} - \mu^{O} > 0$\\
        \begin{align*}
        \begin{cases}
            B^{group}_{g} < 0\\
            B^{group}_{g} + \mu_{g}^{O} - \mu^{O} > 0\\
            -B^{group}_{g} > B^{group}_{g} + \mu_{g}^{O} - \mu^{O}
        \end{cases}
        \Leftrightarrow\begin{cases}
            \rho_g > 0\\
            \mu^{O} < \mu_{g}^{O}\\
            \rho_g \cdot \sqrt{\frac{1}{\alpha_g (1 - \alpha_g)}} \in [\frac{\mu_g^O-\mu^O}{2\sigma_{X | G}}, \frac{\mu_g^O-\mu^O}{\sigma_{X | G}}]
        \end{cases}
    \end{align*}

    Case 4: $B^{group}_{g} < 0$ and $B^{group}_{g} + \mu_{g}^{O} - \mu^{O} < 0$\\
    \begin{align*}
        \begin{cases}
            B^{group}_{g} < 0\\
            B^{group}_{g} + \mu_{g}^{O} - \mu^{O} < 0\\
            -B^{group}_{g} > -B^{group}_{g} - \mu_{g}^{O} + \mu^{O}
        \end{cases}
        \Leftrightarrow\begin{cases}
            \rho_g > 0\\
            \mu^{O} < \mu_{g}^{O}\\
            \rho_g \cdot \sqrt{\frac{1}{\alpha_g (1 - \alpha_g)}} > \frac{\mu_g^O - \mu^O}{\sigma_{X | G}}
        \end{cases}
    \end{align*}
    
    Combining cases 1 and 2, and cases 3 and 4 lead to:
    \begin{align*}
        \begin{cases}
            \rho_g < 0\\
            \mu^{O} > \mu_{g}^{O}\\
            \rho_g \cdot \sqrt{\frac{1}{\alpha_g (1 - \alpha_g)}} < \frac{\mu_g^O - \mu^O}{2\sigma_{X | G}}
        \end{cases}
    	\cup\begin{cases}
            \rho_g > 0\\
            \mu^{O} < \mu_{g}^{O}\\
            \rho_g \cdot \sqrt{\frac{1}{\alpha_g (1 - \alpha_g)}} > \frac{\mu_g^O - \mu^O}{2\sigma_{X | G}}
        \end{cases}
    \end{align*}
\endproof

\subsection{Theorem \ref{theorem:lessunfair}}
\label{proof:lessunfair}

\paragraph{Theorem.}
     Under the simplifying assumptions $\sigma^2_{X \mid  \neg O, G} = \sigma^2_{X \mid  \neg O, \neg G}$, and $\mu^O_{g} > \mu^O$, both imputation strategies penalise the marginalised group and the reconstruction gap is larger for the group imputation than the population one (i.e., $\Delta^{group}_g > \Delta^{pop}_g > 0$) iff: 
    \begin{align*}
        &\begin{cases}
            \rho_g \cdot \sigma_{X | G} \cdot f(\alpha_g, r_g, \alpha_{\neg g}) + \rho_{\neg g} \cdot \sigma_{X | \neg G} \cdot f(\alpha_{\neg g}, 1 - r_g, \alpha_{g}) > ((1 - r_g) \alpha_{\neg g} - r_g \alpha_g) (\mu_{g} - \mu_{\neg g})\\
            \rho_g \cdot \sigma_{X | G} \cdot e(\alpha_g) - \rho_{\neg g} \cdot \sigma_{X | \neg G} \cdot e(\alpha_{\neg g}) > \mu_{g} - \mu_{\neg g}\\
            \rho_g \cdot \sigma_{X | G} \cdot h(\alpha_g, r_g, \alpha_{\neg g}) + \rho_{\neg g} \cdot \sigma_{X | \neg G} \cdot h(\alpha_{\neg g}, 1 - r_g, \alpha_{g}) > ((1 - r_g) \alpha_{\neg g} - r_g \alpha_g) (\mu_{g} - \mu_{\neg g})\\
        \end{cases}\\  
        \text{or}&\\
        &\begin{cases}
            \rho_g \cdot \sigma_{X | G} \cdot f(\alpha_g, r_g, \alpha_{\neg g}) + \rho_{\neg g} \cdot \sigma_{X | \neg G} \cdot f(\alpha_{\neg g}, 1 - r_g, \alpha_{g}) > ((1 - r_g) \alpha_{\neg g} - r_g \alpha_g) (\mu_{g} - \mu_{\neg g})\\
            \rho_g \cdot \sigma_{X | G} \cdot e(\alpha_g) - \rho_{\neg g} \cdot \sigma_{X | \neg G} \cdot e(\alpha_{\neg g}) < \mu_{g} - \mu_{\neg g}\\
            \rho_g \cdot \sigma_{X | G} \cdot h(\alpha_g, r_g, \alpha_{\neg g}) + \rho_{\neg g} \cdot \sigma_{X | \neg G} \cdot h(\alpha_{\neg g}, 1 - r_g, \alpha_{g}) < ((1 - r_g) \alpha_{\neg g} - r_g \alpha_g) (\mu_{g} - \mu_{\neg g})
        \end{cases} 
    \end{align*}        
    with $r_g = \mathbb{P}[G = g]$, the ratio of the population belonging to group $g$, $f(\alpha_g, r_g, \alpha_{\neg g}) = \frac{2\alpha_{\neg g} (1-r_g)}{\sqrt{\alpha_{g}(1-\alpha_{g})}}  - \sqrt{\frac{1 - \alpha_{g}}{\alpha_{g}}} \cdot (\alpha_{\neg g} (1 - r_{g}) - \alpha_g r_g)$, $e(\alpha_g) = \sqrt{\frac{\alpha_{g}}{1-\alpha_{g}}}$, and $h(\alpha_g, r_g, \alpha_{\neg g}) = \frac{\alpha_{g}r_g + \alpha_{\neg g} (1 - r_g)}{\sqrt{\alpha_{g}(1-\alpha_{g})}} - \sqrt{\frac{1 - \alpha_{g}}{\alpha_{g}}} \cdot (\alpha_{\neg g} (1 - r_{g}) - \alpha_g r_g)$.
    \vspace{1em}

\proof{Proof.}
   Let consider these following two expressions $\Delta^{group}_g > \Delta^{pop}_g$ and $\Delta^{pop}_g > 0$ separately. First, consider the expression $\Delta^{group}_g > \Delta^{pop}_g$, this can further be expressed as follows:
    \begin{align*}
        \Delta^{group}_g > \Delta^{pop}_g 
        \Leftrightarrow& B^{group^2}_{g} - B^{group^2}_{\neg g} >  B^{pop^2}_{g} - B^{pop^2}_{\neg g}\\
        \Leftrightarrow& \left(B^{pop}_{\neg g} - B^{group}_{\neg g}\right) \left(B^{pop}_{\neg g} + B^{group}_{\neg g}\right) > \left(B^{pop}_{g} - B^{group}_{g}\right) \left(B^{pop}_{g} + B^{group}_{g}\right)\\
        \Leftrightarrow& \left(\mu^O_{\neg g} - \mu^O\right) \left(B^{pop}_{\neg g} + B^{group}_{\neg g}\right) > \left(\mu^O_{g} - \mu^O\right) \left(B^{pop}_{g} + B^{group}_{g}\right)
    \end{align*}
    Using Equation~\eqref{eq:extendedpop}, this corresponds to:
    \begin{align*}
        \alpha_g r_g \gamma \bigg(2\rho_{\neg g} \cdot \sqrt{\frac{1}{\alpha_{\neg g} (1 - \alpha_{\neg g})}} \cdot \sigma_{X | \neg G} &+ \frac{\alpha_g r_g}{\alpha} \cdot \gamma \bigg) > \\ 
        &\alpha_{\neg g}  (1 - r_{g}) \gamma \left(-2\rho_g \cdot \sqrt{\frac{1}{\alpha_{g} (1 - \alpha_{g})}} \cdot \sigma_{X | G} + \frac{\alpha_{\neg g} (1 - r_{g})}{\alpha} \cdot \gamma \right)
    \end{align*}
    Assuming $\mu^O_{g} > \mu^O$ results in $\gamma > 0$:
    \begin{align*}   
        \alpha_g r_g 2\rho_{\neg g} \cdot \sqrt{\frac{1}{\alpha_{\neg g} (1 - \alpha_{\neg g})}} \cdot \sigma_{X | \neg G} + \alpha_{\neg g} (1 - r_{g}) 2\rho_g \cdot &\sqrt{\frac{1}{\alpha_{g} (1 - \alpha_{g})}} \cdot \sigma_{X | G} >\frac{-(\alpha_g r_g)^2 + (\alpha_{\neg g} (1 - r_{g}))^2}{\alpha_g r_g + \alpha_{\neg g} (1 - r_{g})} \gamma \\
        \Leftrightarrow \rho_g \cdot \sigma_{X | G} \cdot f(\alpha_g, r_g, \alpha_{\neg g}) + \rho_{\neg g} \cdot \sigma_{X | \neg G} \cdot f(\alpha_{\neg g}, 1 - r_g&, \alpha_{g}) > ((1 - r_g) \alpha_{\neg g} - r_g \alpha_g) (\mu_{g} - \mu_{\neg g})
    \end{align*}
    with $\gamma = \rho_g \cdot \sqrt{\frac{1 - \alpha_g}{\alpha_g}} \cdot \sigma_{X \mid G} + \mu_g - \mu_{\neg g} - \rho_{\neg g} \cdot \sqrt{\frac{1 - \alpha_{\neg g}}{\alpha_{\neg g}}} \cdot \sigma_{X \mid \neg G}$,
    \\and $f(\alpha_g, r_g, \alpha_{\neg g}) = \frac{2\alpha_{\neg g} (1-r_g)}{\sqrt{\alpha_{g}(1-\alpha_{g})}}  - \sqrt{\frac{1 - \alpha_{g}}{\alpha_{g}}} \cdot (\alpha_{\neg g} (1 - r_{g}) - \alpha_g r_g)$, $f: [0, 1]^3 \longrightarrow \mathbbm{R}^+$.\\

    Second, consider $\Delta^{pop}_g > 0$, we focus on the case in which both components are positive. The equivalence is obtained by considering the complementary case in which both components are negative.
    \begin{align*}
            &\begin{cases}
                B^{group}_{g} - B^{group}_{\neg g} + \mu_g^O - \mu_{\neg g}^O > 0\\
                B^{group}_{g} + B^{group}_{\neg g} + \mu_g^O + \mu_{\neg g}^O - 2\mu^O> 0
            \end{cases} \tag*{\textit{(Assuming $\sigma^2_{X \mid  \neg O, G} = \sigma^2_{X \mid  \neg O, \neg G}$)}} \\
        \Leftrightarrow&
            \begin{cases}
                -\rho_g \cdot \sqrt{\frac{1}{\alpha_{g} (1 - \alpha_{g})}} \cdot \sigma_{X | G} + \rho_{\neg g} \cdot \sqrt{\frac{1}{\alpha_{\neg g} (1 - \alpha_{\neg g})}} \cdot \sigma_{X | \neg G}  > - \gamma\\
                -\rho_g \cdot \sqrt{\frac{1}{\alpha_{g} (1 - \alpha_{g})}} \cdot \sigma_{X | G} - \rho_{\neg g} \cdot \sqrt{\frac{1}{\alpha_{\neg g} (1 - \alpha_{\neg g})}} \cdot \sigma_{X | \neg G}  > - \frac{\alpha_{\neg g} (1 - r_{g}) - \alpha_g r_g}{\alpha_g r_g + \alpha_{\neg g} (1 - r_{g})} \gamma\\
            \end{cases}\\
        \Leftrightarrow&
            \begin{cases}
                 \rho_g \cdot \sigma_{X | G} \cdot e(\alpha_g) - \rho_{\neg g} \cdot \sigma_{X | \neg G} \cdot e(\alpha_{\neg g}) < \mu_{g} - \mu_{\neg g}\\
                 \rho_g \cdot \sigma_{X | G} \cdot h(\alpha_g, r_g, \alpha_{\neg g}) + \rho_{\neg g} \cdot \sigma_{X | \neg G} \cdot h(\alpha_{\neg g}, 1 - r_g, \alpha_{g}) < ((1 - r_g) \alpha_{\neg g} - r_g \alpha_g) (\mu_{g} - \mu_{\neg g})\\
            \end{cases}\\
    \end{align*}
    with $e(\alpha_g) = \sqrt{\frac{\alpha_{g}}{1-\alpha_{g}}}$ and $h(\alpha_g, r_g, \alpha_{\neg g}) = \frac{\alpha_{g}r_g + \alpha_{\neg g} (1 - r_g)}{\sqrt{\alpha_{g}(1-\alpha_{g})}} - \sqrt{\frac{1 - \alpha_{g}}{\alpha_{g}}} \cdot (\alpha_{\neg g} (1 - r_{g}) - \alpha_g r_g)$.

    Therefore $\Delta^{group}_g > \Delta^{pop}_g > 0$ is equivalent to satisfy the following set of equations:
    \begin{align*}
        &\begin{cases}
            \rho_g \cdot \sigma_{X | G} \cdot f(\alpha_g, r_g, \alpha_{\neg g}) + \rho_{\neg g} \cdot \sigma_{X | \neg G} \cdot f(\alpha_{\neg g}, 1 - r_g, \alpha_{g}) > ((1 - r_g) \alpha_{\neg g} - r_g \alpha_g) (\mu_{g} - \mu_{\neg g})\\
            \rho_g \cdot \sigma_{X | G} \cdot e(\alpha_g) - \rho_{\neg g} \cdot \sigma_{X | \neg G} \cdot e(\alpha_{\neg g}) > \mu_{g} - \mu_{\neg g}\\
            \rho_g \cdot \sigma_{X | G} \cdot h(\alpha_g, r_g, \alpha_{\neg g}) + \rho_{\neg g} \cdot \sigma_{X | \neg G} \cdot h(\alpha_{\neg g}, 1 - r_g, \alpha_{g}) > ((1 - r_g) \alpha_{\neg g} - r_g \alpha_g) (\mu_{g} - \mu_{\neg g})\\
        \end{cases}\\  
        \text{or}&\\
        &\begin{cases}
            \rho_g \cdot \sigma_{X | G} \cdot f(\alpha_g, r_g, \alpha_{\neg g}) + \rho_{\neg g} \cdot \sigma_{X | \neg G} \cdot f(\alpha_{\neg g}, 1 - r_g, \alpha_{g}) > ((1 - r_g) \alpha_{\neg g} - r_g \alpha_g) (\mu_{g} - \mu_{\neg g})\\
            \rho_g \cdot \sigma_{X | G} \cdot e(\alpha_g) - \rho_{\neg g} \cdot \sigma_{X | \neg G} \cdot e(\alpha_{\neg g}) < \mu_{g} - \mu_{\neg g}\\
            \rho_g \cdot \sigma_{X | G} \cdot h(\alpha_g, r_g, \alpha_{\neg g}) + \rho_{\neg g} \cdot \sigma_{X | \neg G} \cdot h(\alpha_{\neg g}, 1 - r_g, \alpha_{g}) < ((1 - r_g) \alpha_{\neg g} - r_g \alpha_g) (\mu_{g} - \mu_{\neg g})
        \end{cases} 
    \end{align*}         
\endproof

\begin{figure*}[ht!]
    \centering
     \includegraphics[height=150px]{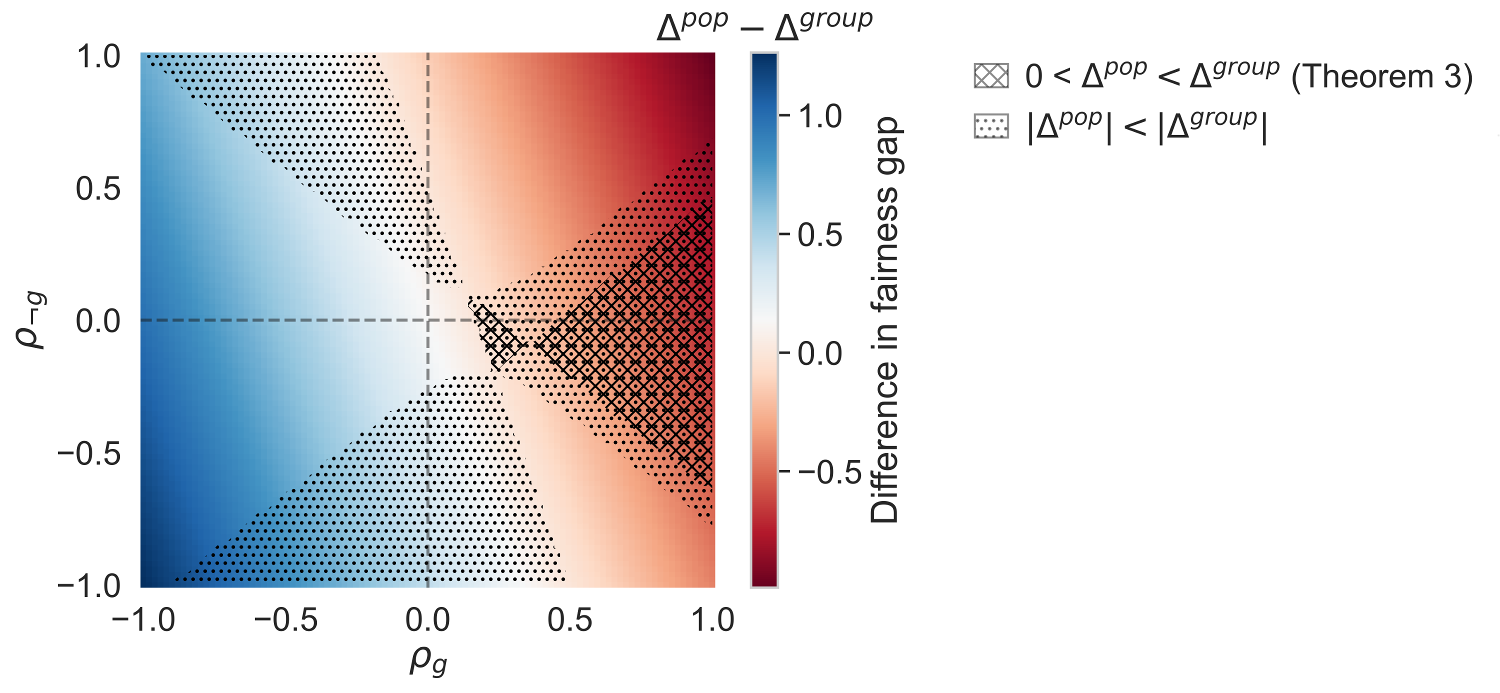}
     \caption{Difference in fairness gap between population imputation and group imputation reconstruction errors. In red, the fairness gap is larger for the group imputation strategy than the population one. In blue, the opposite is true. The crossed area describes settings satisfying Theorem~\ref{theorem:lessunfair}, i.e. when both strategies result in larger reconstruction errors for group $g$ but population mean imputation reduces the fairness gap in comparison to its group imputation variant. The dotted area presents the generalisation of the Theorem when population mean imputation reduces the absolute fairness gap.}
     \label{app:fig:satisfy}
\end{figure*}

\newpage
\paragraph{Example.} Consider a dataset with the following observed characteristics: observed means $\mu^O_{g} = 0.5$ and $\mu^O_{\neg g} = 0$, the marginalised group ratio $r_g = 25\%$, and the observation rates: $\alpha_g = 0.7$ and $\alpha_{\neg g} = 0.8$. Further, we assume the underlying data characteristic $\sigma_{X \mid G} = \sigma_{X \mid \neg G} = 0.5$, and $\sigma_{X \mid \neg O, G} = \sigma_{X \mid \neg O, \neg G}$.
Figure~\ref{app:fig:satisfy} illustrates the theoretical fairness gap difference and the area satisfying the previous theorems under varying missingness characteristics $\rho_g$ and $\rho_{\neg g}$. 

This example provides evidence that for a set of observed characteristics, the problem of the optimal imputation strategy from a reconstruction error point of view is under-determined. Specifically, two missingness processes could lead to the same observed data characteristics but impact which imputation to choose.

\newpage
\section{Simulation study}
\label{annex:synthetic}
This section provides additional details on the experimental design and sensitivity analyses to explore the robustness of our empirical results. While the observed gaps and performances vary across these additional settings, these analyses result in the same conclusions questioning common imputation practices. This evidence strengthens our conclusions: the presented simulations are not an edge case but a common occurrence under group-specific missingness patterns.

\subsection{Data generation}
\label{annex:data_simulation}
\begin{figure}[ht!]
    \centering
    \includegraphics[width =.6\linewidth]{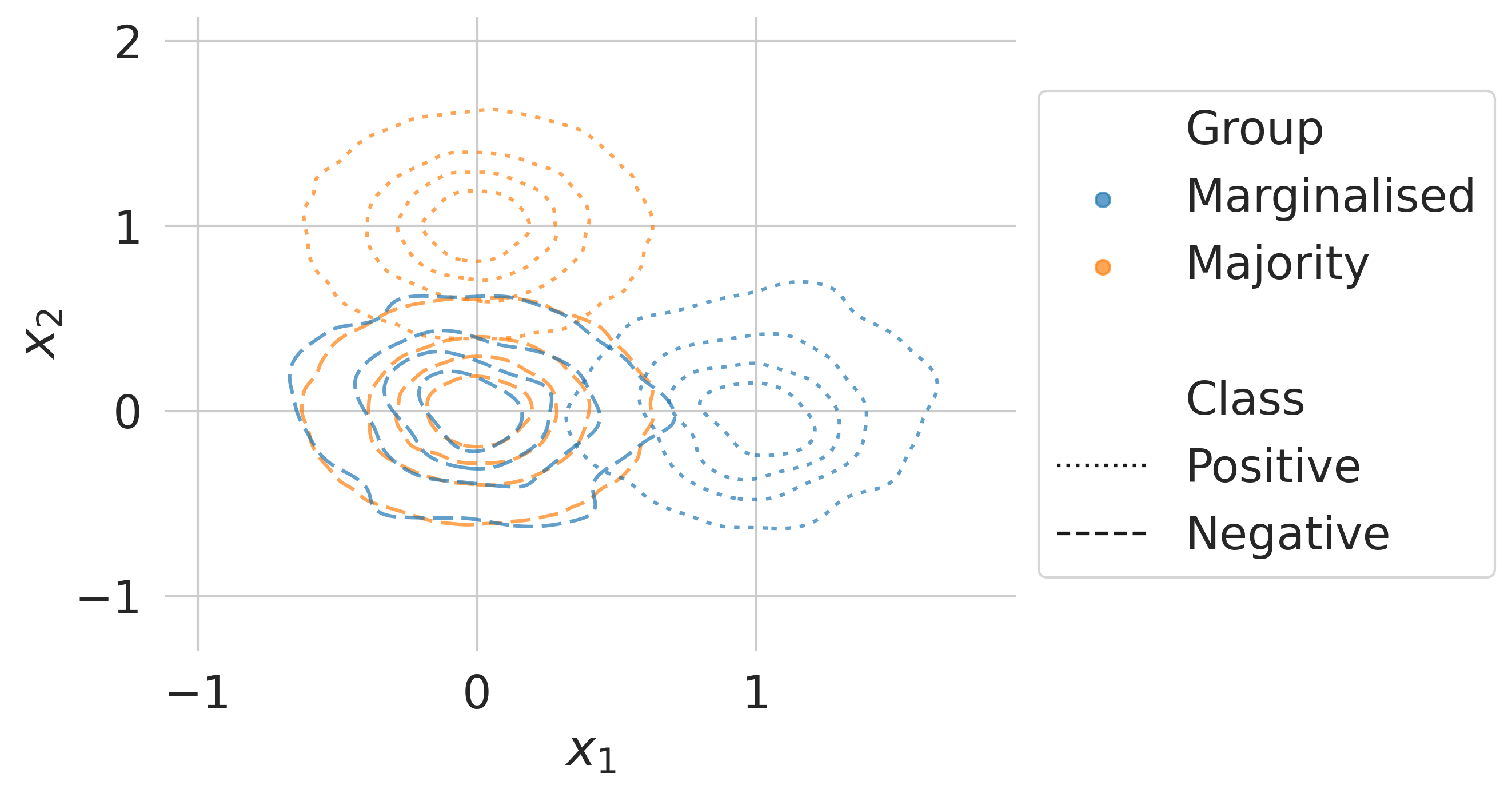}
    \caption{Density distributions of the generated population.}
    \label{fig:annex:distribution}%
\end{figure}

\paragraph{Covariates and outcomes.}
The proposed synthetic population consists of 100,000 points for the majority group and 1,000 for the marginalised group resulting in a sample size of $N = 101,000$ with a ratio of 100:1. Each individual is represented in this dataset as a pair of covariates, i.e., $X \in \mathbb{R}^2$. For each group, 2/3 presents the condition, i.e., $\mathbb{P}(Y_i = 1) = 0.66$. Negatives are drawn from the normal distribution $\mathcal{N}((0, 0), 0.25)$. The condition characterisation, i.e., the boundary between positive and negatives, differs between groups with positive from the majority (resp. the marginalised group) sampled from $\mathcal{N}((1, 0), 0.25)$ (resp. $\mathcal{N}((0, 1), 0.25)$). Figure~\ref{fig:annex:distribution} shows the density distribution of the generated population.

\paragraph{Missingness.}
In this synthetic population, 50\% of the dimension $X_2$ is removed in a given subgroup to enforce the three clinical presence scenarios. We enforce the following clinical missingness:
\begin{itemize}
    \item Limited access to quality care (S1):  $O_2^{(S1)} \mid [G = 1] \sim \text{Bernoulli}(0.5)$
    \item (Mis)-informed collection (S2): $O_2^{(S2)} \mid [X_1 > 0.5] \sim \text{Bernoulli}(0.5)$
    \item Confirmation bias (S3): $O_2^{(S3)} \mid [X_2 > 0.5] \sim \text{Bernoulli}(0.5)$
    \item Mixture (M): $O_2^{(M)} = O_2^{(S1)} \land O_2^{(S2)} \land O_2^{(S3)}$
\end{itemize}
With $O_2$, the observation indicator associated with $X_2$ and $G$, the group membership ($G = 1$ indicates a member of the marginalised group). 

\subsection{Modelling}
We generate 100 datasets and enforce the different missingness patterns before running a logistic regression with an l2 penalty ($\lambda = 1$). Results are computed on the 20\% test set and averaged over the 100 iterations with 95\% confidence bounds reported.

\clearpage
\newpage
\subsection{Sensitivity analyses}
\label{annex:sensitivity}
The simulations relying on the previous data generating process resulted in two core conclusions: (i) consider both population and group-specific imputation variants, because group-specific can harm marginalised groups despite the aim to improve performance for them, and (ii) to improve algorithmic fairness, compare properties of downstream predictive performance.

\subsubsection{Difference in condition manifestation.}
In Section~\ref{sec:simulation}, the proposed simulations have focused on a given underlying distribution of data in which the marginalised group presents the condition differently than the majority. In this section, we propose to study when the condition manifestation is the same across groups but the condition \emph{prevalence} differs. As discussed in Section~\ref{sec:missingness}, this group difference may result in different group-specific missingness processes.

\begin{figure*}[ht!]
    \centering
    \includegraphics[width=0.9\linewidth]{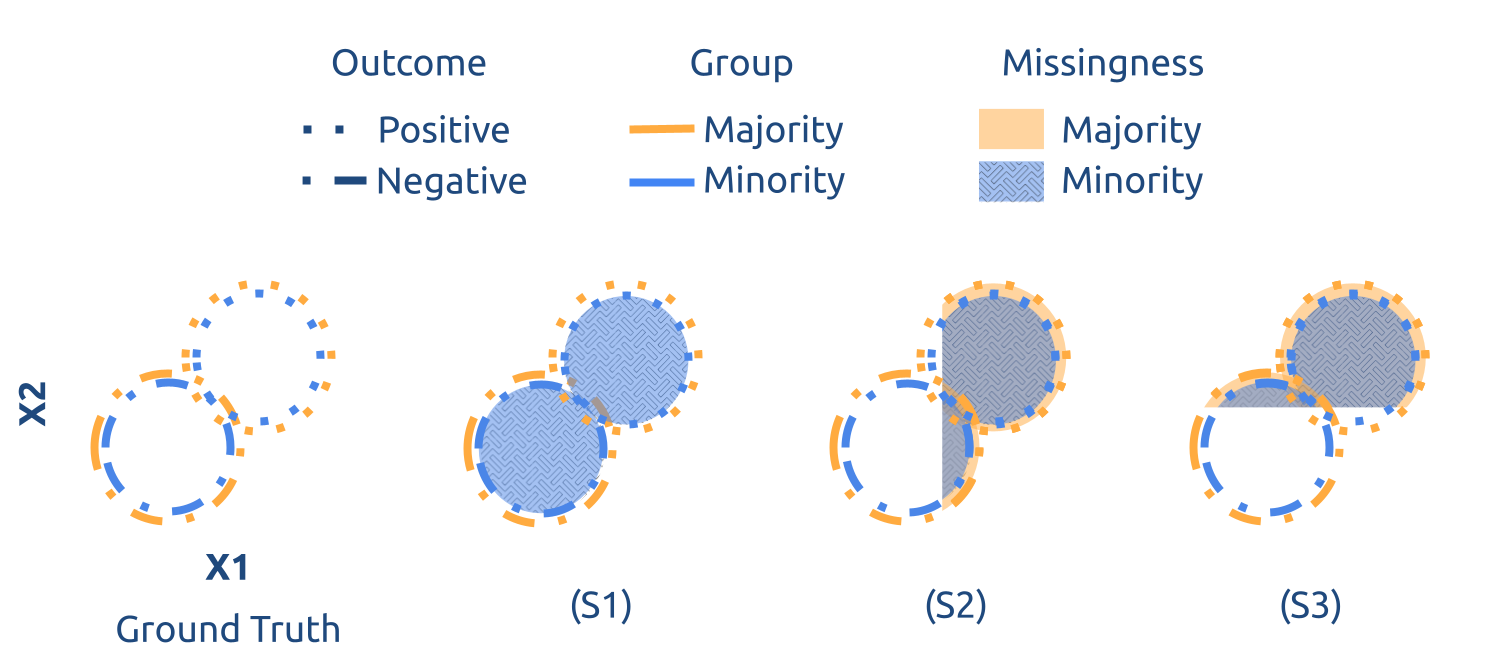}
    \caption{Graphical summary of clinical missingness in the simulation experiments with identical condition manifestation but different prevalence across groups. Missingness is enforced on $X_2$, affecting 50\% of the shaded regions for the indicated group.}
    \label{fig:summary:same}%
\end{figure*}

Consider a population of $N = 101,000$ patients with a ratio of 100:1 for the majority. Patients without the condition are drawn from the normal distribution $\mathcal{N}((0, 0), 0.25)$ and positives are sampled from $\mathcal{N}((1, 1), 0.25)$. Contrarily to the previous simulations, the marginalised group has a prevalence of 50\% while the rest of the population, 10\%. We then enforce the three previously described missingness processes. Figure~\ref{fig:summary:same} illustrates how the same three proposed missingness processes would be expressed in this population. Importantly, due to differences in the prevalence, the missingness processes still differentially affect the two groups. For instance, the proposed (S2) affects the positive cases of both groups, representing 10\% resp. 50\% of these groups. 

\begin{figure*}[ht!]
    \centering
    \includegraphics[height =140px]{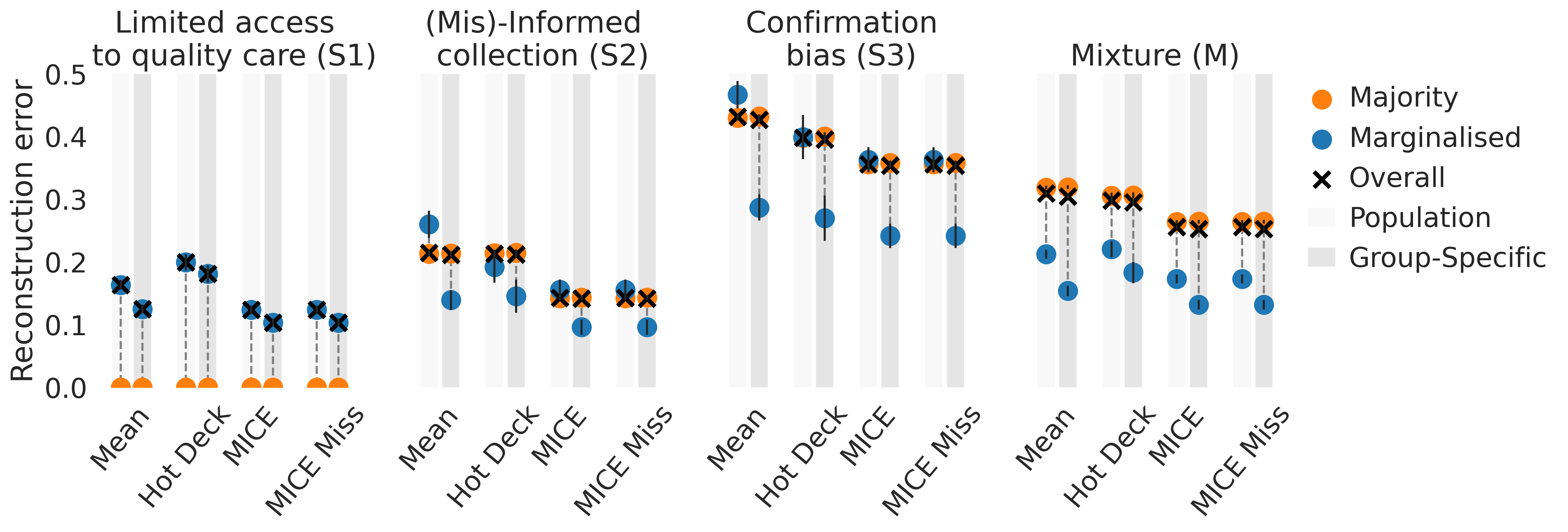}
    \caption{Group-specific reconstruction errors across scenarios on 100 synthetic experiments with the same condition manifestation across groups. Lower reconstruction error is better.}
    \label{fig:summary:rec:same}
\end{figure*}

\begin{figure*}[ht!]
    \centering
    \includegraphics[height =140px]{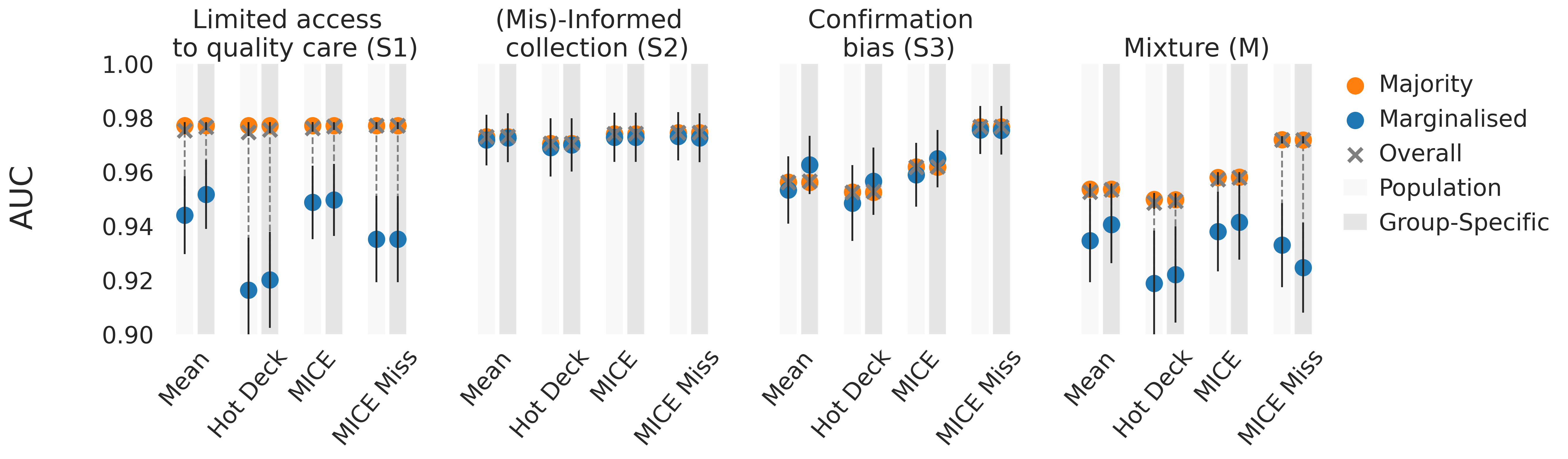}
    \caption{Group-specific AUC performance across scenarios on 100 synthetic experiments with the same condition manifestation across groups.}
    \label{fig:summary:auc:same}
\end{figure*}

Figures~\ref{fig:summary:rec:same} and~\ref{fig:summary:auc:same} present the associated reconstruction error and AUC performance differentiated by scenario and imputation strategies. This set of experiments shows that group-specific condition prevalence would lead to similar discrepancies in reconstruction errors and performance under the different missingness scenarios. First, note how group-specific imputations reduce the reconstruction error gap in (S1) but consistently increase this gap in all other scenarios compared to the population alternatives. While not significant, group MICE Miss also presents a larger gap than its population alternative in terms of downstream performance. Second, while the population and group-specific imputation present different gaps in reconstruction error, this is not reflected in the downstream performance, e.g. MICE alternatives perform the same under (S3) despite presenting opposite reconstruction error gaps.
Echoing the main text's conclusions, these results stand, despite no difference in condition manifestation, highlighting the need for a thorough empirical evaluation of different imputation strategies even when subgroups do not differ in their covariates distributions.

\clearpage
\newpage
\subsubsection{Lower condition prevalence.} 
In Section~\ref{sec:simulation}, the proposed simulations rely on a condition prevalence of 2/3. In this sensitivity analysis, we analyse the same condition expression for a lower prevalence of 10\%, all other factors being kept the same as presented in Appendix~\ref{annex:data_simulation}. 

\begin{figure*}[ht!]
    \centering
    \includegraphics[height =140px]{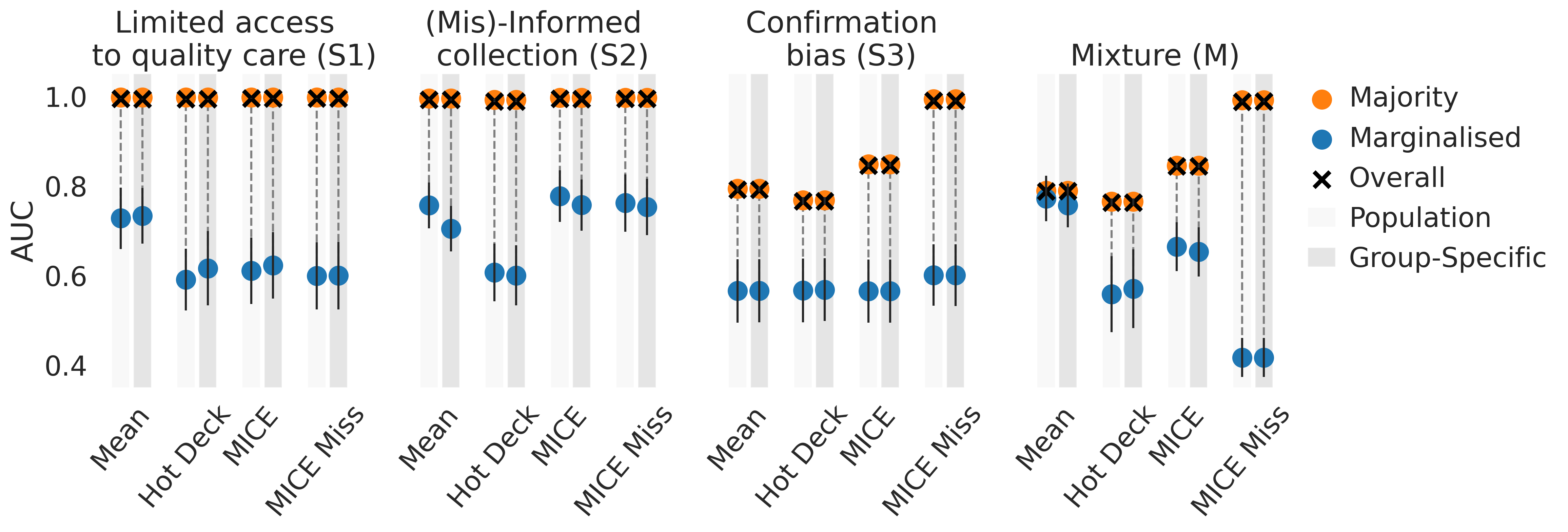}
    \caption{Group-specific AUC performance across scenarios on 100 synthetic experiments with 10\% condition prevalence.}
    \label{fig:summary:auc:prev}
\end{figure*}

Figures~\ref{fig:summary:rec:prev} and~\ref{fig:summary:auc:prev} present the group-specific reconstruction errors and predictive performance when using this alternative condition prevalence. While presenting different performances, these analyses are consistent with Section~\ref{sec:simulation}. In particular, we see that current imputation practices favouring group-specific imputation can be suboptimal regarding both reconstruction error (e.g. Mean imputation in S3) and downstream predictive performance (e.g. Mean imputation in S2). Further, reliance on a single imputation strategy can unnecessarily discriminate against the marginalised group, as a different choice of imputation could avoid this unfairness., e.g. while Mean imputation would improve predictive performance for both groups under (S1) and (S2), relying on this choice would lead to worse performance for both groups in comparison to Mean Miss under (S3). Finally, reducing the reconstruction gap does not always improve downstream predictive fairness, as shown in (M), where all strategies present similar reconstruction gaps but quite different downstream ones. These results further validate our conclusions under more realistic condition prevalence.

\begin{figure*}[ht!]
    \centering
    \includegraphics[height =140px]{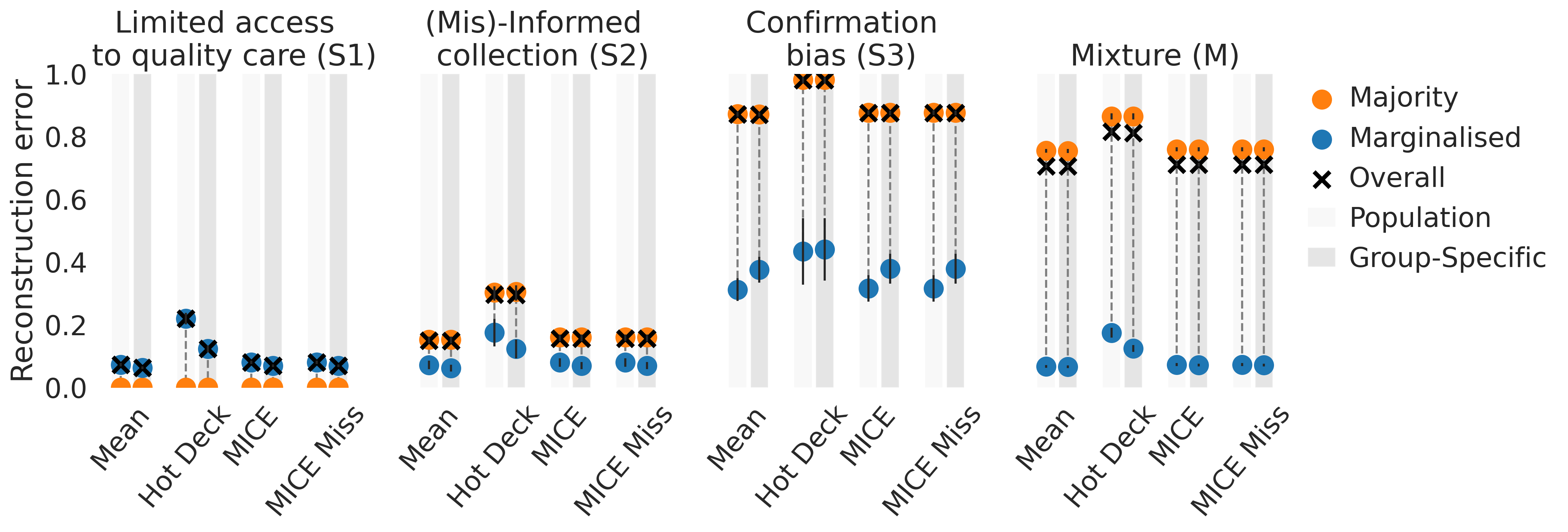}
    \caption{Group-specific reconstruction errors across scenarios on 100 synthetic experiments with 10\% condition prevalence. Lower reconstruction error is better.}
    \label{fig:summary:rec:prev}
\end{figure*}

\newpage
\newpage
\subsubsection{Correlated covariates.}
\begin{figure*}[ht!]
    \centering
    \includegraphics[width=0.9\linewidth]{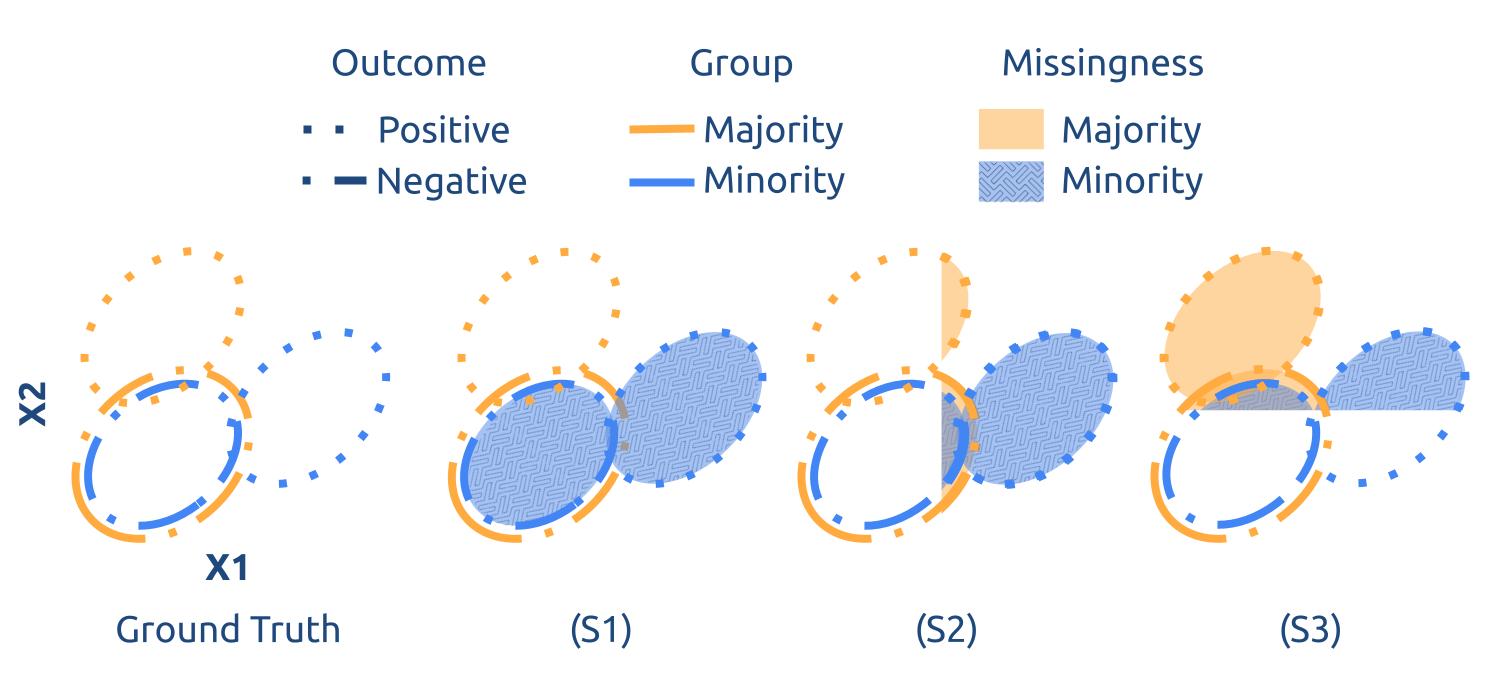}
    \caption{Graphical summary of clinical missingness in the simulation experiments with correlated covariates. Missingness is enforced on $X_2$, affecting 50\% of the shaded regions for the indicated group.}
    \label{fig:summary:correlation}%
\end{figure*}

The covariate distributions introduced in Appendix~\ref{annex:data_simulation} do not present correlations between the covariates at the group level. While the missingness process can introduce informative correlation structures that MICE imputation would leverage for improved imputation, this setting may not reflect the strength of MICE imputation strategies. In this analysis, we enforce correlation between $X_1'$ and $X_2'$ using the same settings as previously introduced, except that the first covariate is defined as the sum of the previous two covariates, i.e. $(X_1', X_2')= (X_1 + X_2, X_2)$. This data generation enforces a correlation structure between covariates as schematised in Figure~\ref{fig:summary:correlation}. Using this data, we then enforce the same missing scenarios.

\begin{figure*}[ht!]
    \centering
    \includegraphics[height =140px]{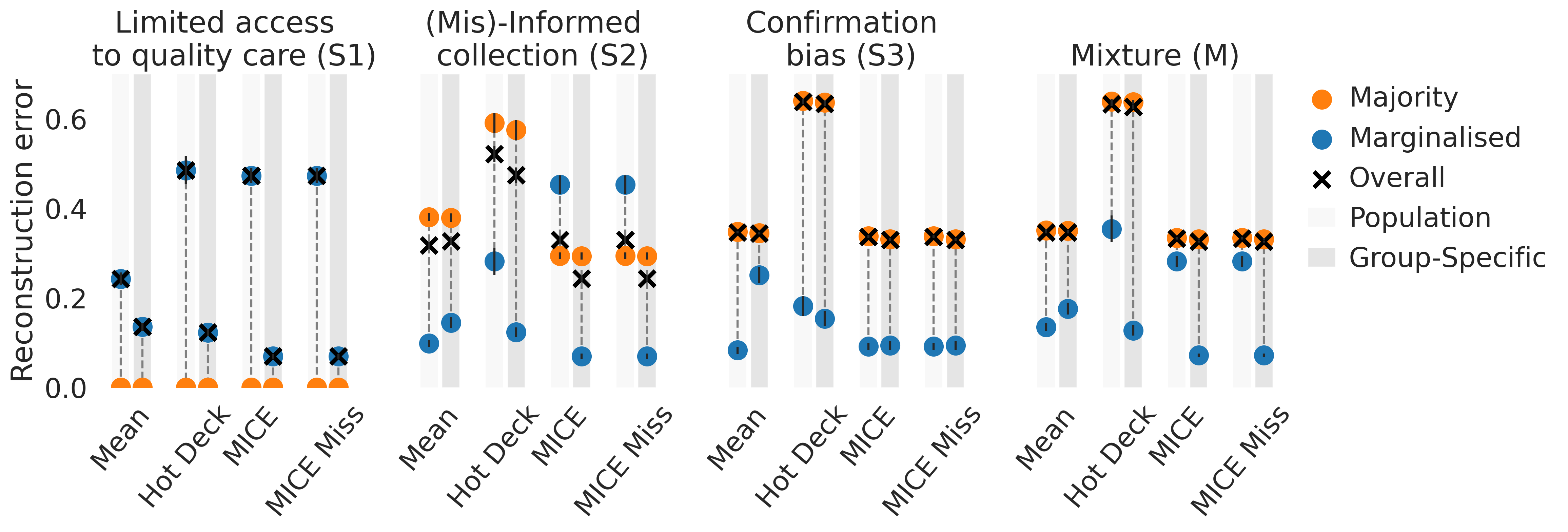}
    \caption{Group-specific reconstruction errors across scenarios on 100 synthetic experiments with correlated covariates. Lower reconstruction error is better.}
    \label{fig:summary:rec:corr}
\end{figure*}

\begin{figure*}[ht!]
    \centering
    \includegraphics[height =140px]{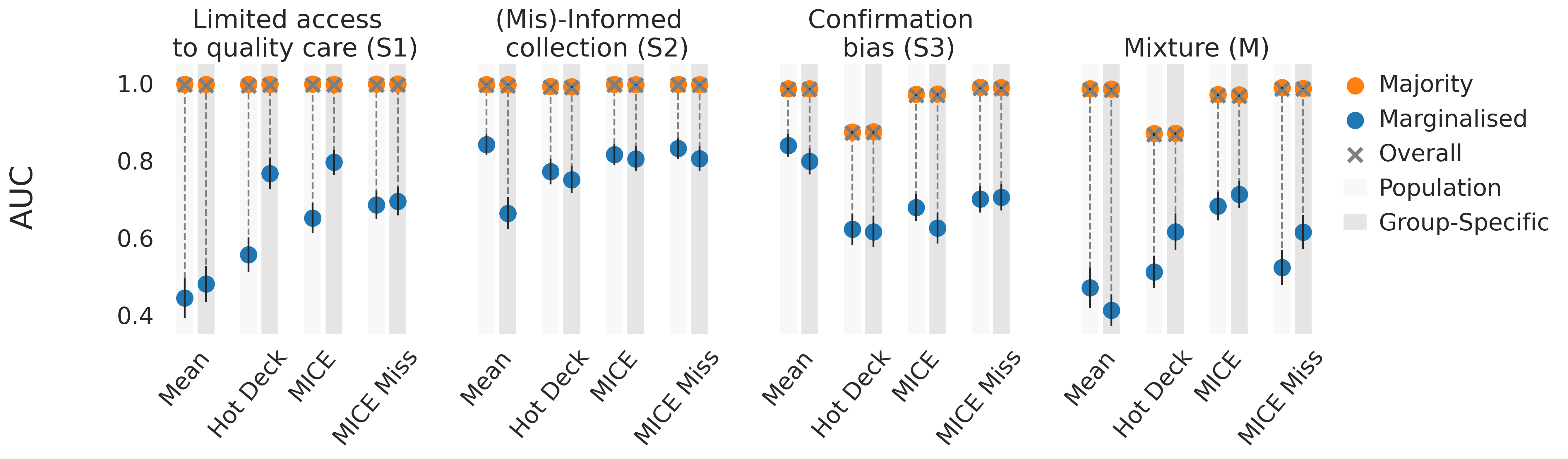}
    \caption{Group-specific AUC performance across scenarios on 100 synthetic experiments when covariates are correlated.}
    \label{fig:summary:auc:corr}
\end{figure*}

Figures~\ref{fig:summary:rec:corr} and~\ref{fig:summary:auc:corr} present the associated reconstruction error and AUC performance differentiated by scenario and imputation strategies. Note how group imputation can increase the fairness gap both in reconstruction error --- as shown in (S2) with Hot Deck, and in downstream performance --- as shown by Mean imputation in (S2). Focusing on downstream performance associated with MICE, the recommendation for group-specific is beneficial for MICE in (S1), but negatively impacts performance for the marginalised group in (S3), further demonstrating the inadequacy of this recommendation. Finally, the minimisation of the reconstruction gap does not translate at the level of the predictive performance as shown in (S3), where group MICE Miss presents one of the smallest reconstruction gaps but the largest downstream performance gap.

\clearpage
\subsubsection{A larger marginalised group.} 
In Section~\ref{sec:synthetic}, the simulations consider a marginalised group, equivalent to 1\% of the majority group. In this section, we propose to analyse the impact on downstream performance when considering a larger marginalised group. Specifically, we consider a marginalised group equivalent to 10\% and 50\% of the majority group of size 10,000 while maintaining the rest of the simulation setting as in the main text. 

\begin{figure*}[ht!]
    \centering
    \includegraphics[height =140px]{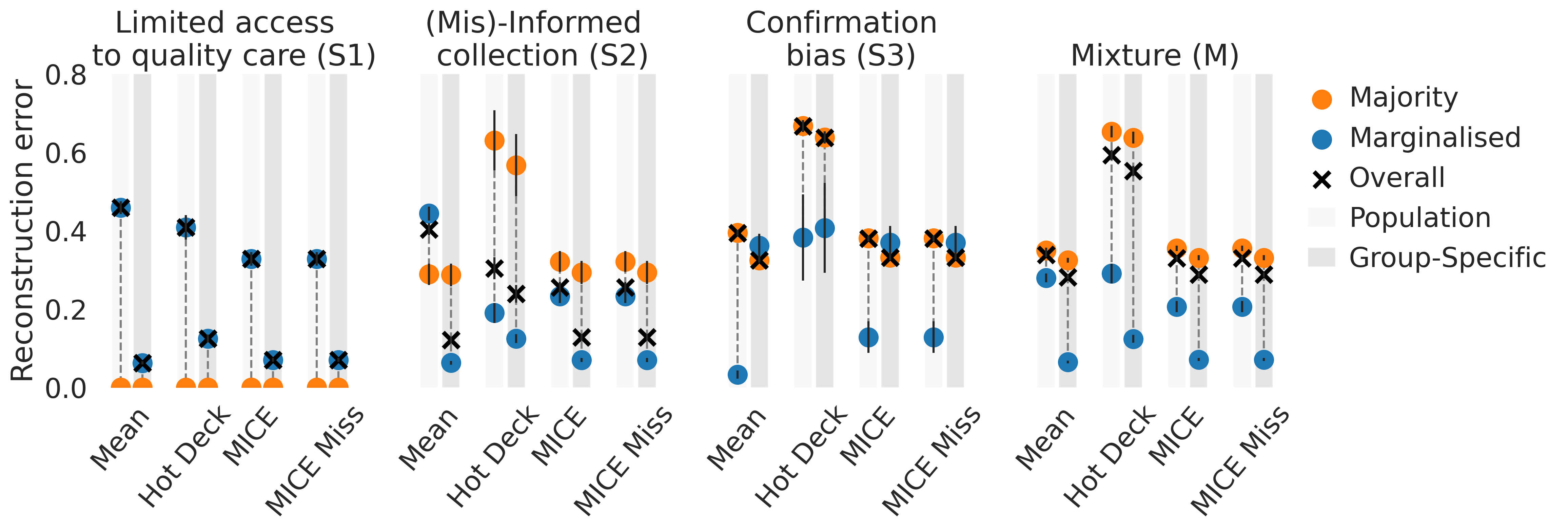}
    \caption{Group-specific reconstruction error across scenarios on 100 synthetic experiments with a marginalised group equivalent to 10\% of the majority one.}
    \label{fig:synthetic:rec10}
\end{figure*}

\begin{figure*}[ht!]
    \centering
    \includegraphics[height =140px]{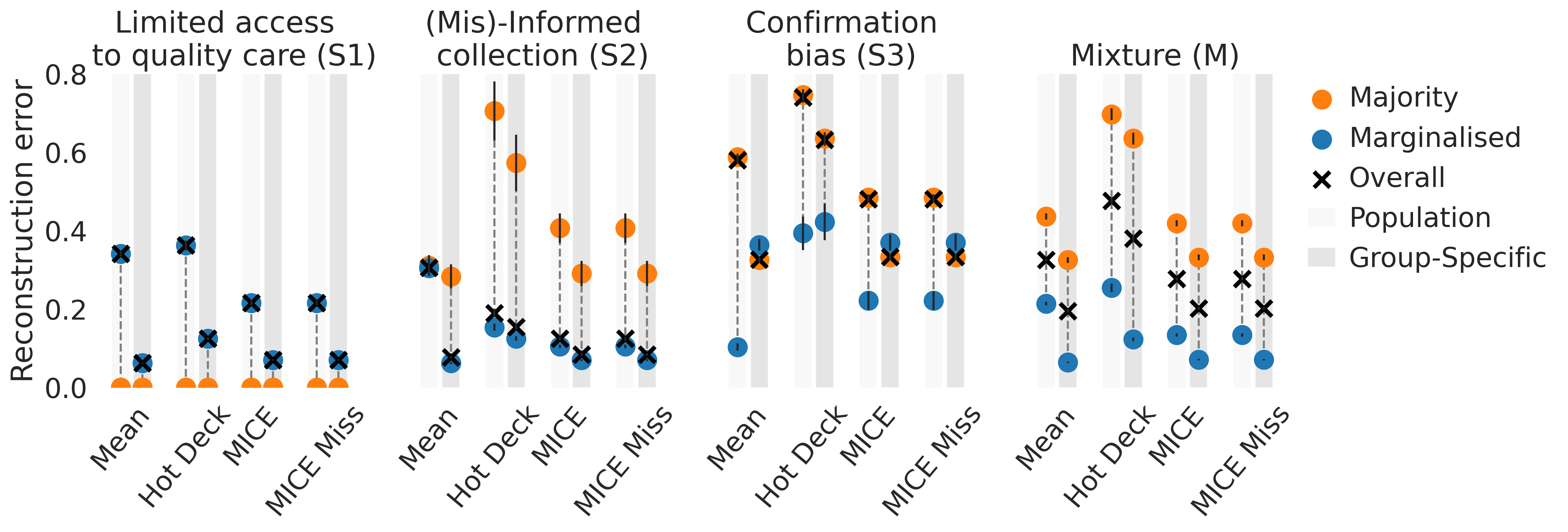}
    \caption{Group-specific reconstruction error across scenarios on 100 synthetic experiments with a marginalised group equivalent to 50\% of the majority one.}
    \label{fig:synthetic:rec50}
\end{figure*}

\begin{figure*}[ht!]
    \centering
    \includegraphics[height =140px]{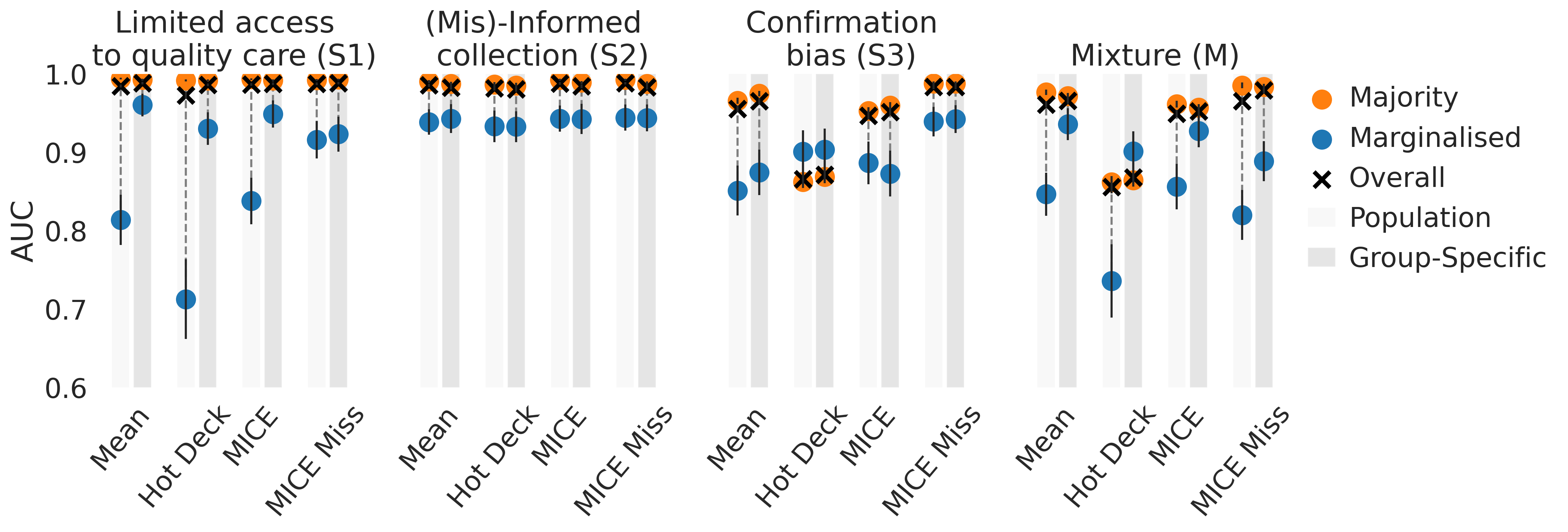}
    \caption{Group-specific AUC performance across scenarios on 100 synthetic experiments with a marginalised group equivalent to 10\% of the majority one.}
    \label{fig:synthetic:10}
\end{figure*}

\begin{figure*}[ht!]
    \centering
    \includegraphics[height =140px]{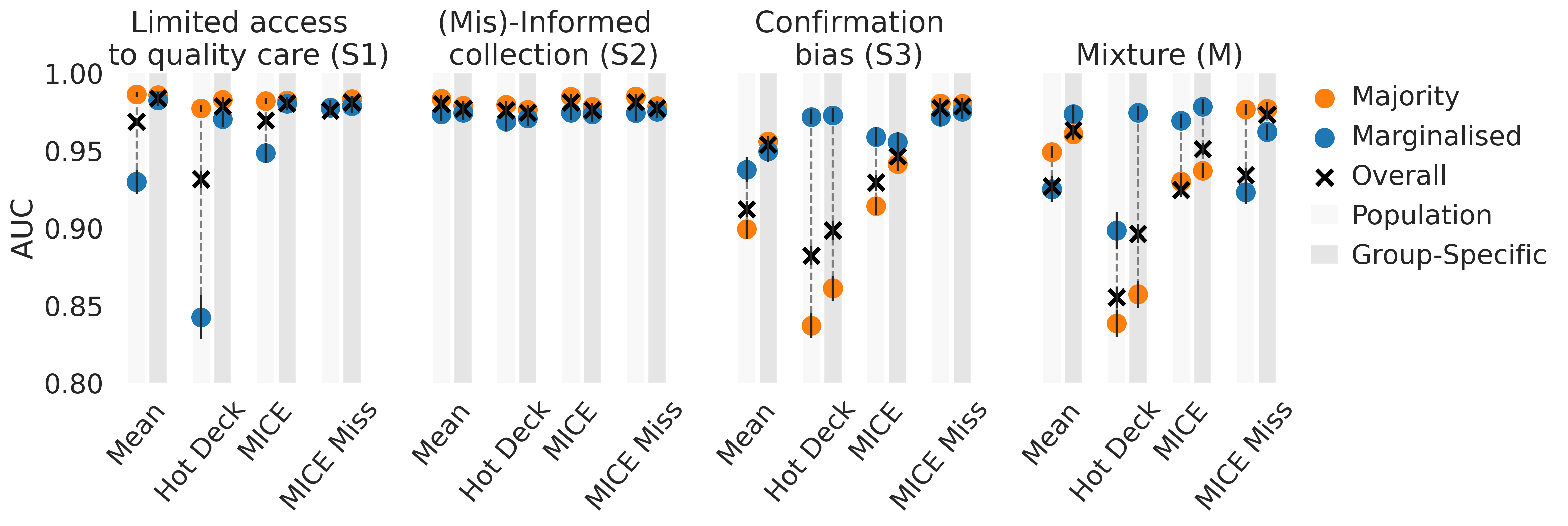}
    \caption{Group-specific AUC performance across scenarios on 100 synthetic experiments with a marginalised group equivalent to 50\% of the majority one.}
    \label{fig:synthetic:50}
\end{figure*}

Figures~\ref{fig:synthetic:rec10} and~\ref{fig:synthetic:rec50} present the associated reconstruction errors and Figures~\ref{fig:synthetic:10} and~\ref{fig:synthetic:50} present downstream predictive performances. Despite similar overall performances, two imputation strategies can have drastically different impacts on the minority group, as indicated by Mean in (S1). Particularly, the group-specific recommendation can increase algorithmic fairness gaps. Under both prevalences, Group Mean imputation presents larger reconstruction gaps than its population alternative. Note that group-specific strategies are beneficial in these simulations, but the reliance on any one imputation could unnecessarily lower performances. For instance, in Figure~\ref{fig:synthetic:rec50}, the direction of the gap is inverted between MICE and MICE Miss, with better performance for the majority group when including the missingness indicator, whereas better performance for the marginalised group under MICE. Finally, while both alternatives of MICE Miss present different gaps in reconstruction errors, the two methodologies have the same predictive performance, further validating the disconnect between reducing reconstruction and performance gaps.

\newpage
\subsubsection{Increased covariate noise.} In Section~\ref{sec:synthetic}, the covariate distributions are generated from normal distributions with a standard deviation of $0.25$. In this additional setting, we consider a larger standard deviation resulting in overlapping covariate distributions between positive and negative cases. This additional noise in the data generation renders the distributions harder to separate by a classification model. Specifically, we consider a standard deviation of $0.5$, with a marginalised group equivalent to 10\% of the majority group of size 10,000 while maintaining the rest of the simulation setting as in the main text. 

\begin{figure*}[ht!]
    \centering
    \includegraphics[height =140px]{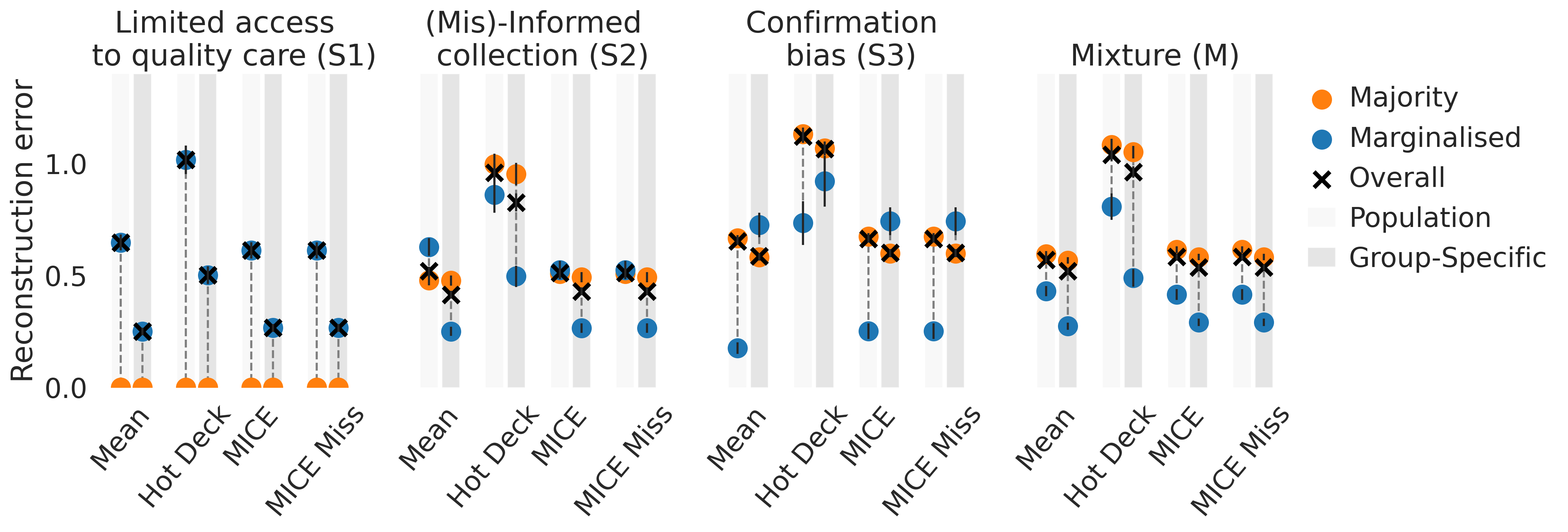}
    \caption{Group-specific reconstruction error across scenarios on 100 synthetic experiments with additional covariate noise.}
    \label{fig:synthetic:noiserec}
\end{figure*}

\begin{figure*}[ht!]
    \centering
    \includegraphics[height =140px]{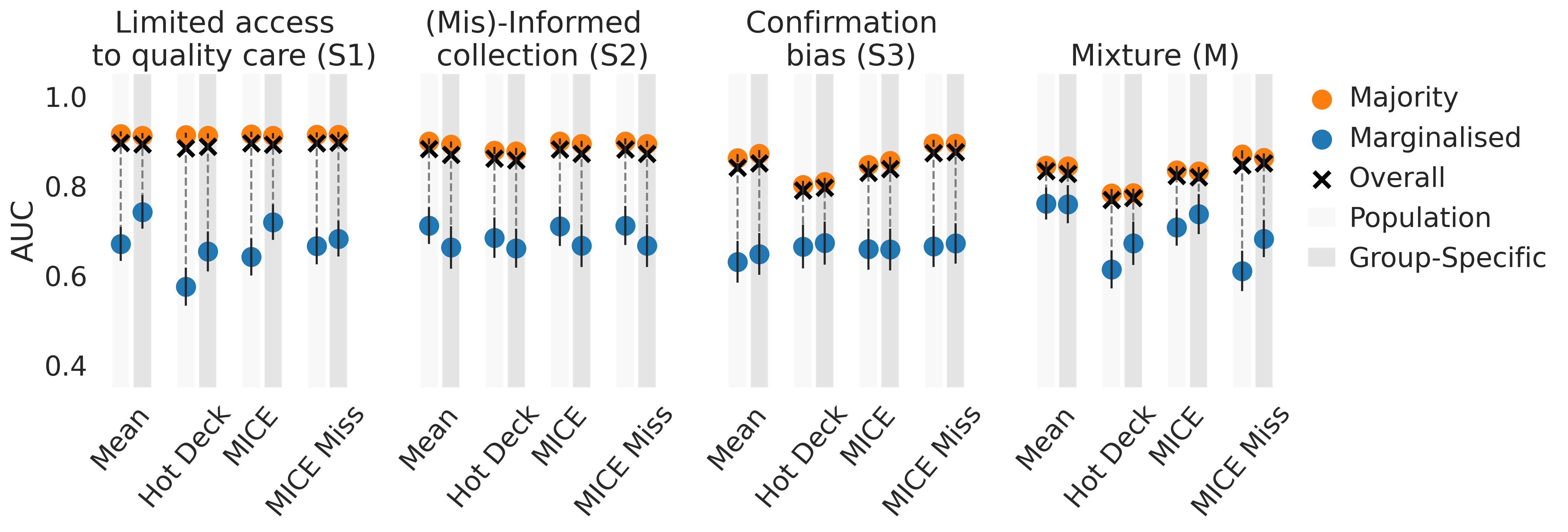}
    \caption{Group-specific AUC performance across scenarios on 100 synthetic experiments with additional covariate noise.}
    \label{fig:synthetic:noiseauc}
\end{figure*}

Figure~\ref{fig:synthetic:noiserec} presents the reconstruction errors, and Figure~\ref{fig:synthetic:noiseauc} the downstream predictive performances. This increased covariate noise does lower the AUC of the synthetic experiments. While impacting performance, these additional results show that adding noise does not alter the main text conclusions. First, the use of group-specific imputation can increase reconstruction errors, as shown in (S3), where Group Mean results in a larger reconstruction error for the marginalised group than its population alternative. Further, while Group Mean reduces reconstruction error in (S2), it does not improve downstream performances, with the population mean presenting the best predictive performances for all groups. Similarly, while group and population imputation present large discrepancies in reconstruction errors under (S3), no difference appears in their predictive performances. These observations confirm the disconnect between the fairness of reconstruction errors and downstream performances. Further, scenario (S2) continues to show slight improvements in predictive performance and the fairness gap from population imputation strategies compared to group-specific imputations. 

\clearpage
\clearpage
\section{MIMIC III}
\label{annex:mimic}

\subsection{Dataset}
\begin{figure}[ht!]
    \centering
    \includegraphics[width=.5\linewidth]{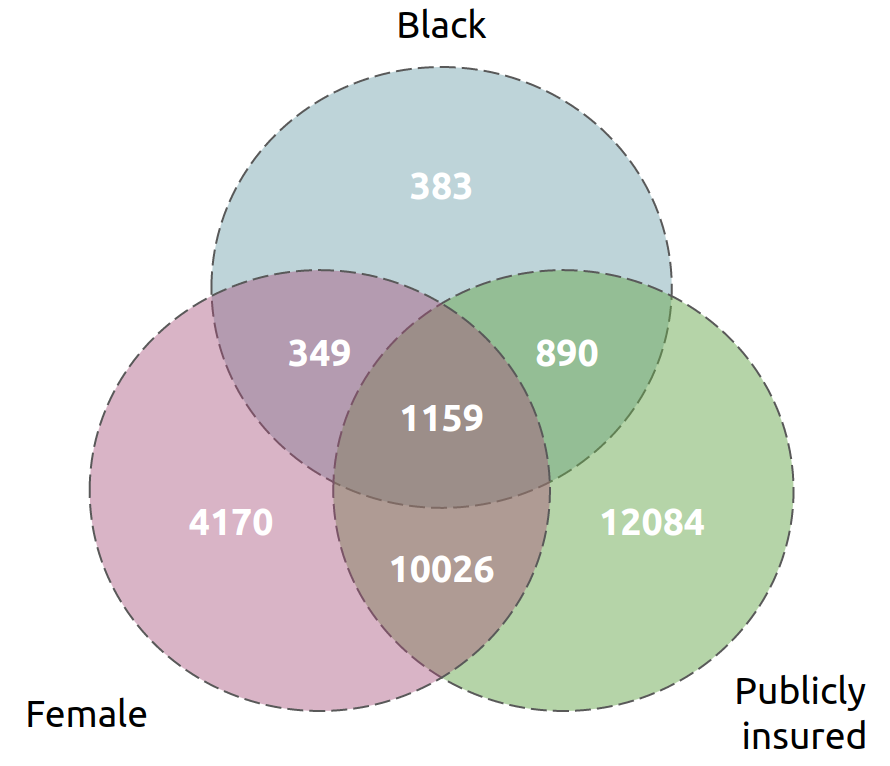}
    \caption{Venn diagram of the population distribution in the three marginalised groups.}
    \label{fig:annex:mimic: marginalised}%
\end{figure}

After preprocessing~\cite{wang2020mimic} and standardisation, the MIMIC III dataset consists of $36,296$ patients with 67 different laboratory tests. Focusing on the three marginalised groups of interest, the population can be further divided into marginalised subgroups as presented in Figure~\ref{fig:annex:mimic: marginalised}. This representation underlines the importance of identifying subgroups at risk in the studied population. 

\subsection{Experimental design}
For this real-world dataset, patients are split into three groups: 80\% for training, 10\% for validation and 10\% for hyperparameters selection. The hyperparameter search consisted of the l2 penalty selection for the logistic regression among $\lambda \in [0.1, 1., 10., 100.]$.

We bootstrap the test set 100 times and report the mean and 95\% confidence bounds.

\newpage
\subsection{Controlling for one group at a time} 
\label{annex:mimic:onegroup} 
\begin{figure*}[ht!]
    \centering
        \includegraphics[height =140px]{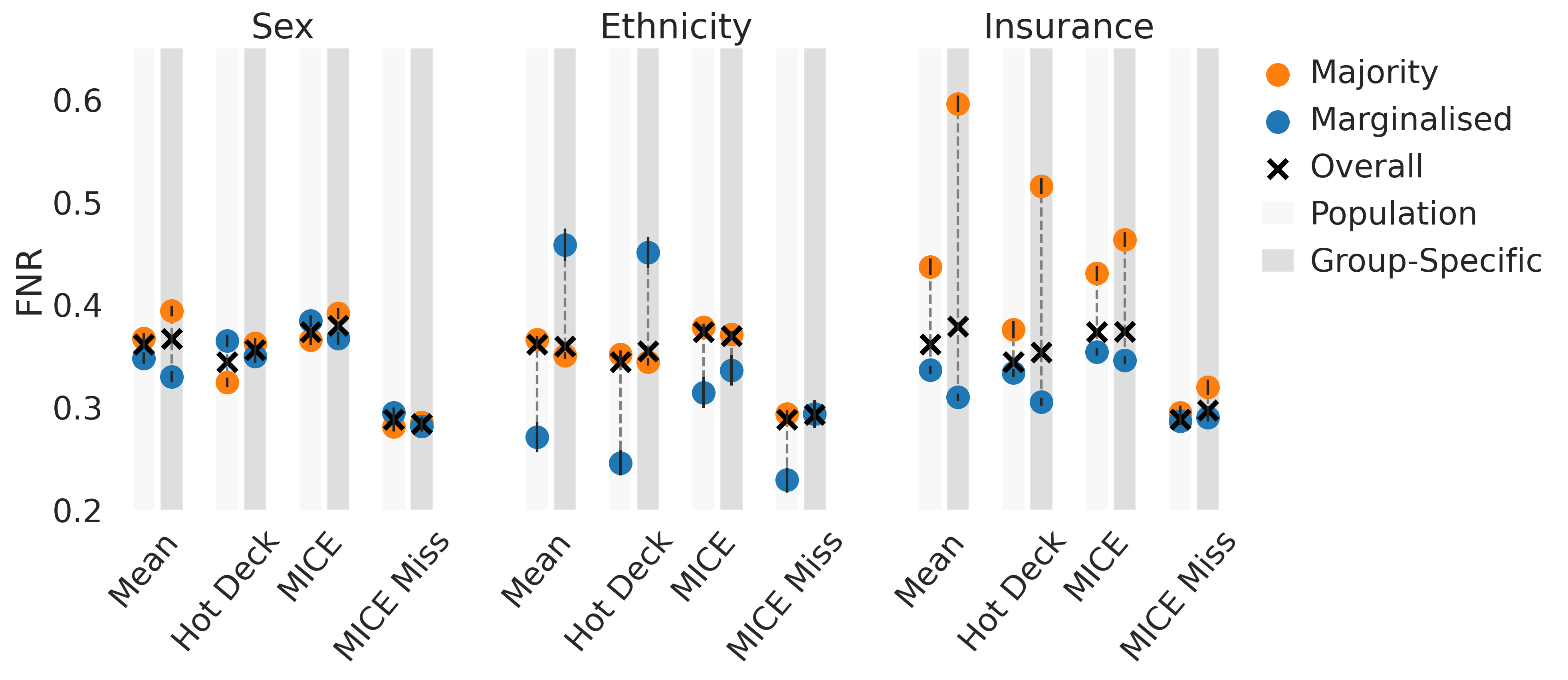}
        \caption{False negative rates across marginalised groups in MIMIC III experiment when controlling on each respective group.}
    \label{fig:mimic:all}
\end{figure*}

In Section~\ref{sec:mimic}, the group-specific variants correspond to strategies controlling for all groups of interest. Alternatively, one could consider each group individually, using a group-specific imputation strategy for each respective group. The use of these imputation strategies results in the updated Figure~\ref{fig:mimic:all}. These experiments lead to the same conclusions in which practitioners should prefer the population MICE Miss imputation to improve downstream predictive performance. However, note that in these figures, each group-specific strategy refers to a different imputation.

\subsection{Metric sensitivity}
\label{annex:mimic:threshold}
In Section~\ref{sec:mimic}, we present results for a policy of 30\% additional care. As we fix this threshold using the observed prioritisation rate, we propose to measure how the results would vary under increased and decreased thresholds: 5\% and 50\%. 

\begin{figure*}[ht!]
    \centering
    \includegraphics[height =140px]{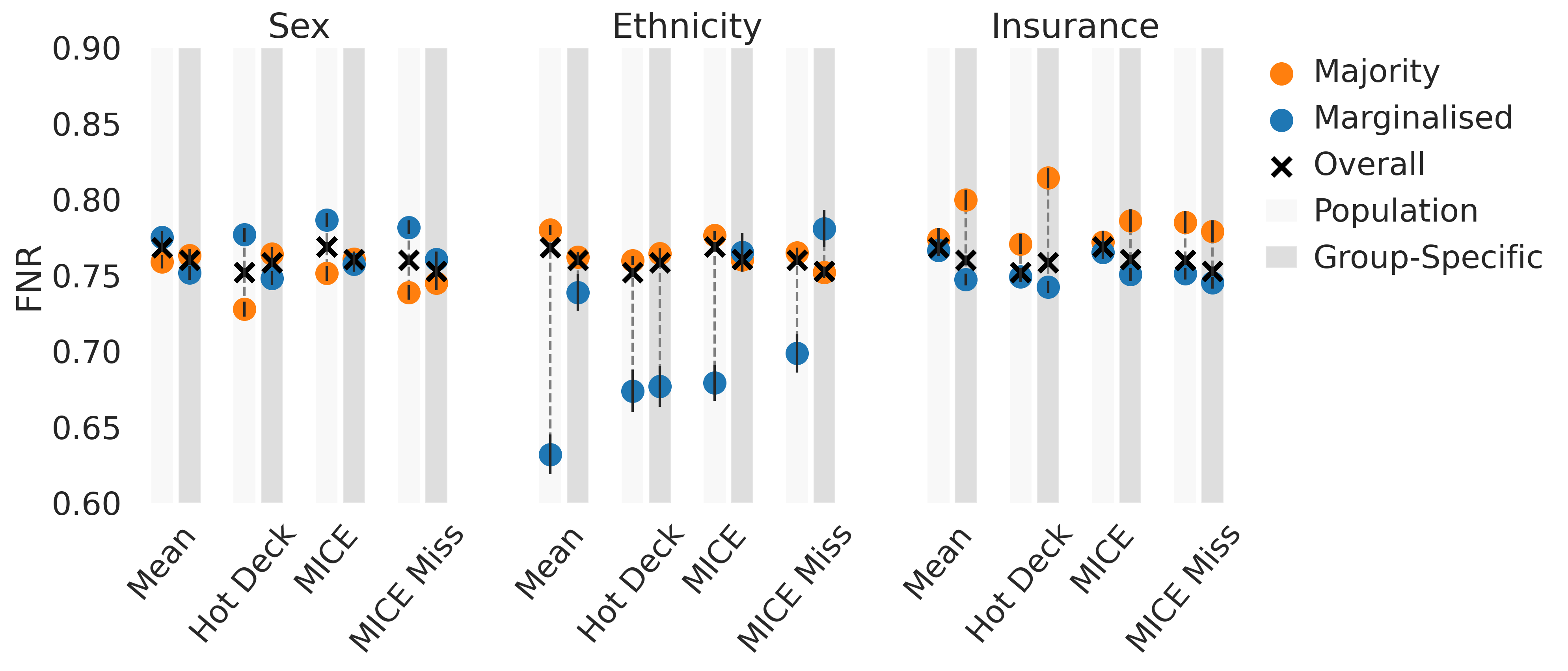}
    \caption{False negative rates across marginalised groups in MIMIC III experiment for 5\% additional care.}
    \label{fig:mimic:5}
\end{figure*}

\begin{figure*}[ht!]
    \centering
    \includegraphics[height =140px]{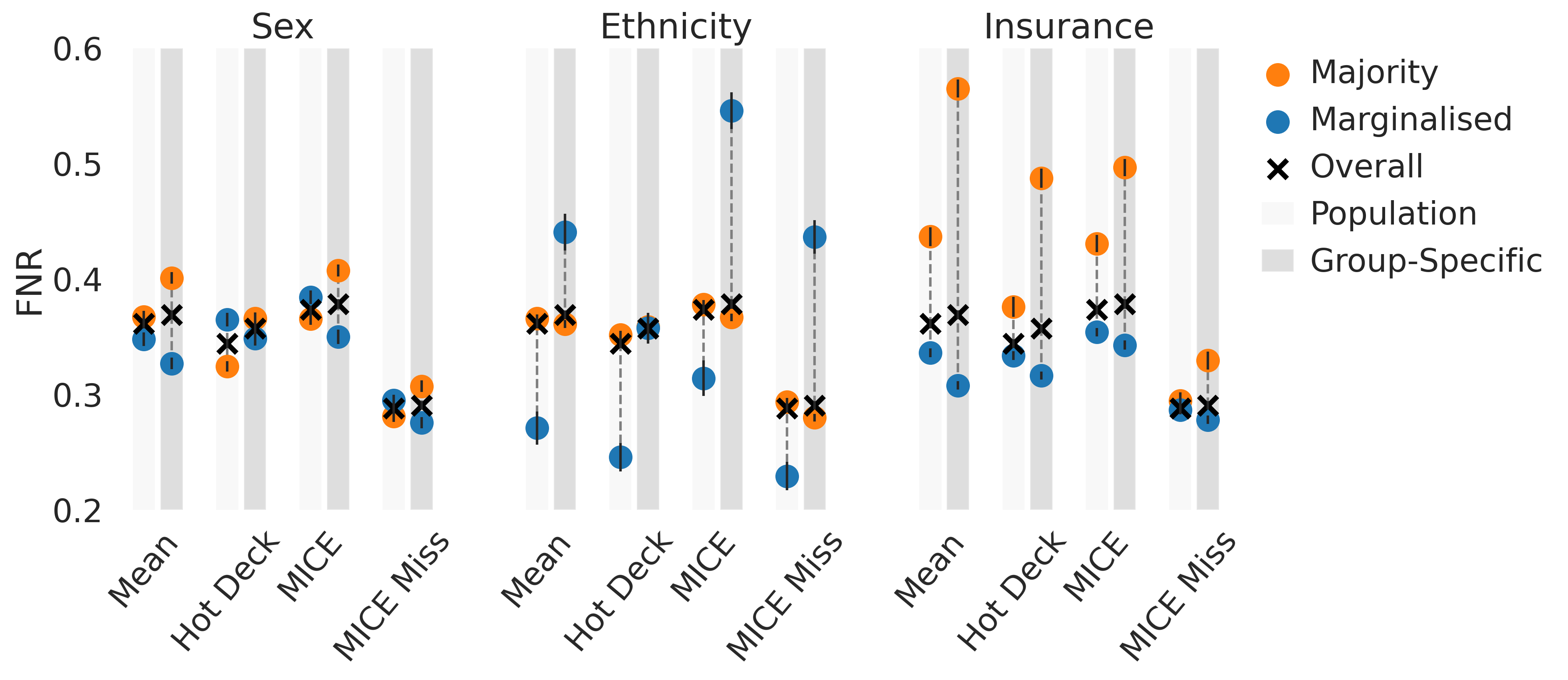}
    \caption{False negative rates across marginalised groups in MIMIC III experiment for 30\% additional care.}

    \label{fig:mimic:30}
\end{figure*}

\begin{figure*}[ht!]
    \centering
    \includegraphics[height =140px]{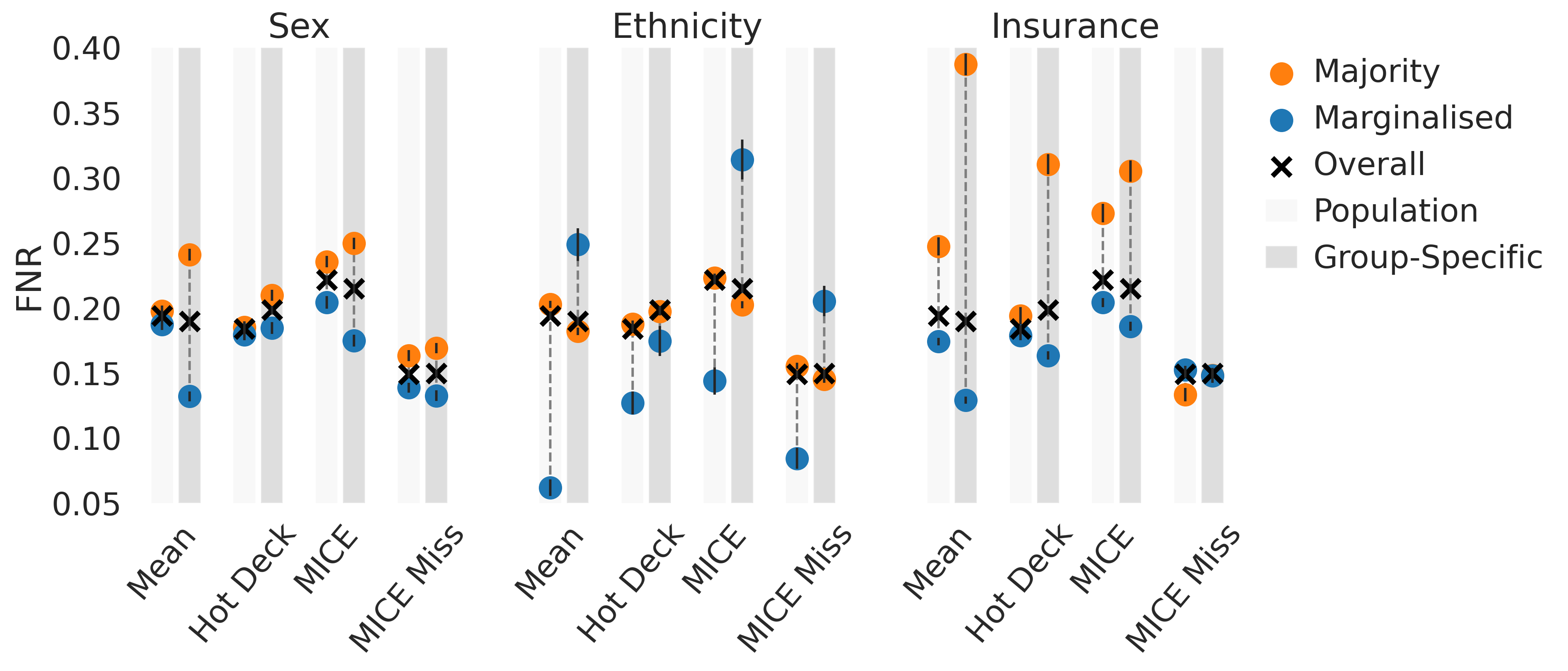}
    \caption{False negative rates across marginalised groups in MIMIC III experiment for 50\% additional care.}
    \label{fig:mimic:50}
\end{figure*}

Figures~\ref{fig:mimic:5},~\ref{fig:mimic:30} and~\ref{fig:mimic:50} present the results at 5\%, 30\% (as in the main text) and 50\% thresholds. First, note that the magnitude of the FNR increases with smaller thresholds, as less patients are prioritised, more high-risk patients are missed. Second, group-specific performances depend upon imputation choice, echoing the same results as in Section~\ref{sec:mimic}. Critically, while MICE should be preferred at higher prioritisation rates, the choice of the optimal imputation is more complex for the 5\% prioritisation rate. This additional set of experiments demonstrates that the target task may also affect which imputation strategy best serves the task at hand. 

\newpage
\subsection{Alternative modelling strategies}
\label{annex:mimic:models} 
\begin{figure*}[ht!]
    \centering
    \includegraphics[height =140px]{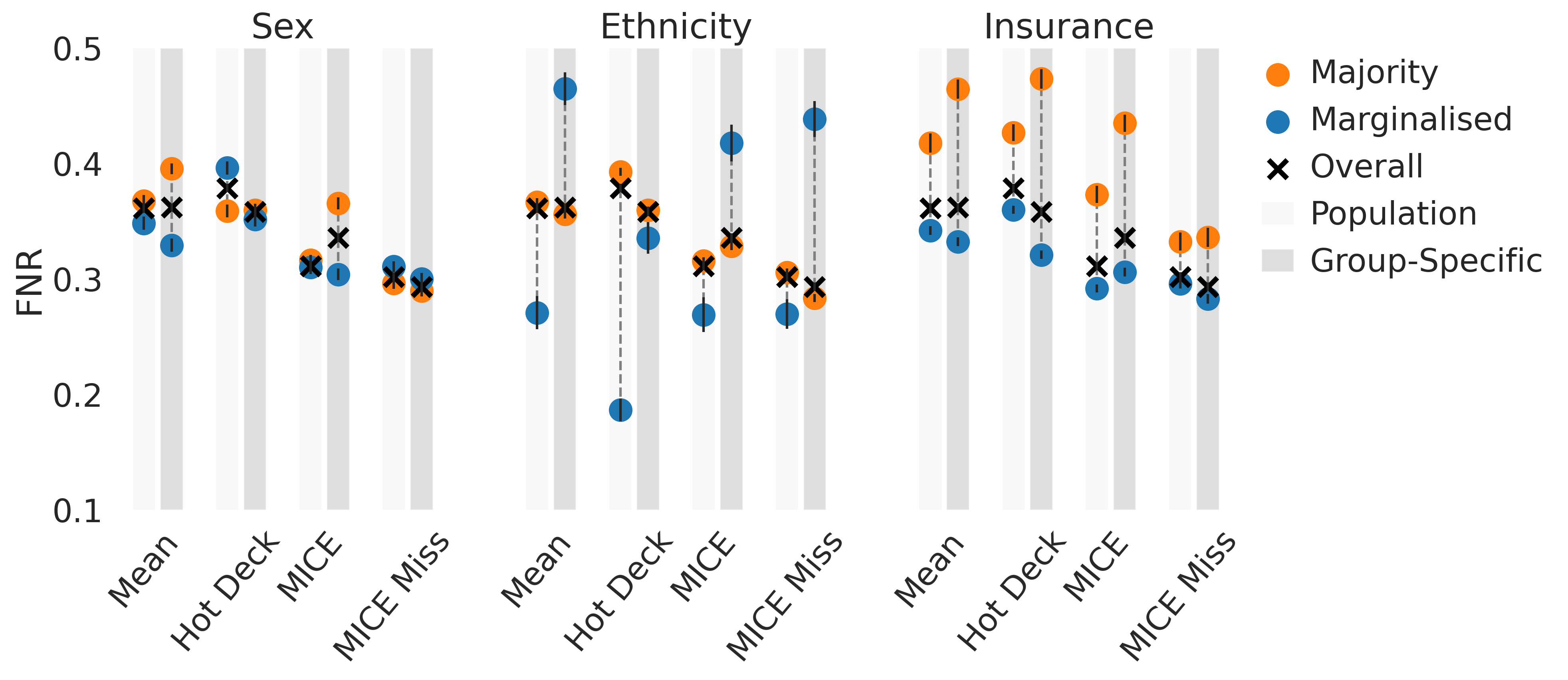}
    \caption{False negative rates across marginalised groups in MIMIC III experiment with a neural network modelling the outcome of interest.}
    \label{fig:mimic:nn}
\end{figure*}

\begin{figure*}[ht!]
    \centering
    \includegraphics[height =140px]{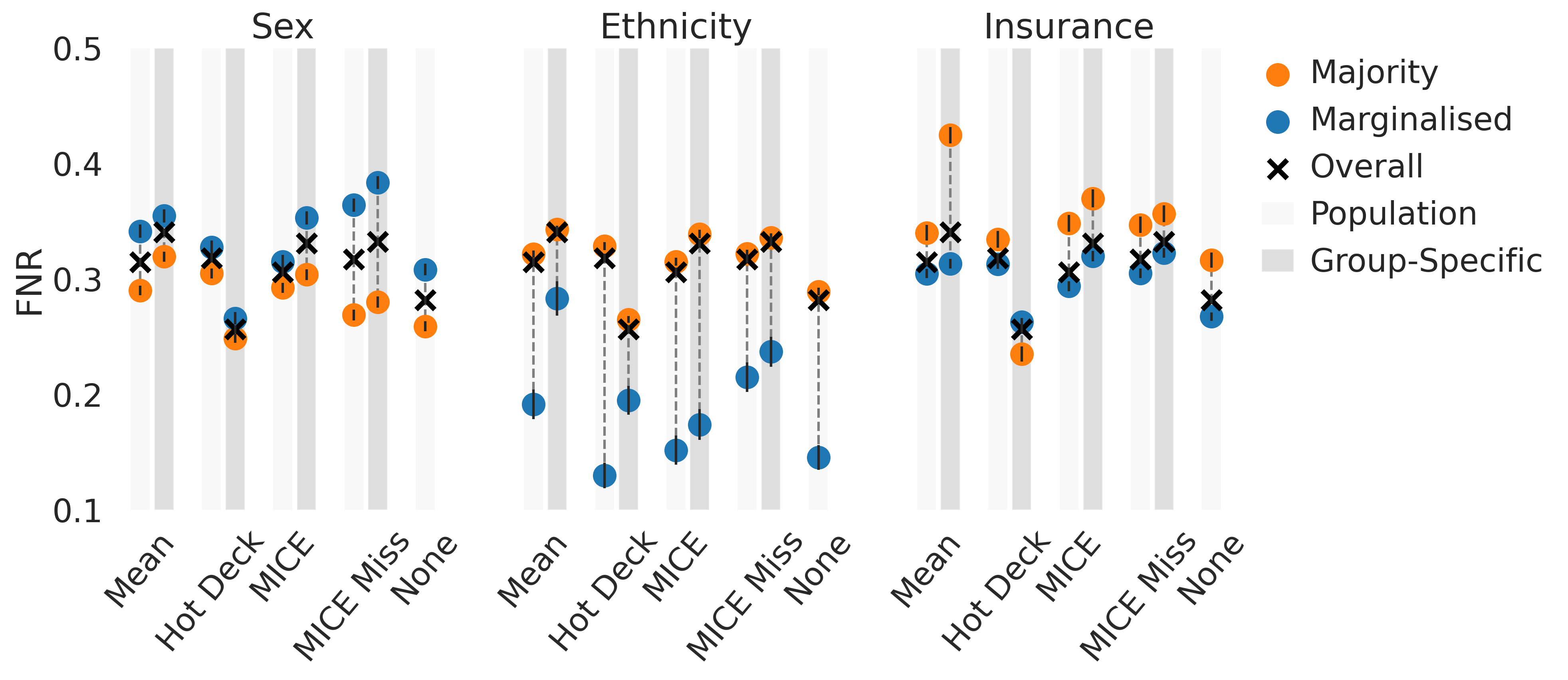}
    \caption{False negative rates across marginalised groups in MIMIC III experiment with a decision tree modelling the outcome of interest.}
    \label{fig:mimic:decision}
\end{figure*}

In Section~\ref{sec:mimic}, we use a logistic regression model to regress the outcomes on the imputed covariates as often observed in medical analyses. In these additional experiments, we explore if one would observe a similar impact of the choice of imputation strategy on algorithmic fairness when considering alternative modelling strategies. Particularly, we focus on advanced predictive models, namely neural networks and decision trees. Figures~\ref{fig:mimic:nn} and~\ref{fig:mimic:decision} present the observed group-specific false negative rates for both predictive models, similarly echoing the main text conclusions. Scikit-learn~\cite{scikit-learn}'s implementation of decision trees can directly handle missingness. During training, the optimisation criterion is computed for missing data assigned to either descendant node at each tree split. Missing data are then assigned to the node that maximises the criterion. This strategy using missing data without imputation, denoted by None in Figure~\ref{fig:mimic:decision} due to the absence of imputation, does not improve performance or reduce the gap between groups.

These results echo the results presented in Section~\ref{sec:mimic}: the choice of imputation impacts downstream performance even under more flexible strategies. Our results first highlight that using different predictive models impacts performance, with decision trees most improving performance. Further, the imputation choice impacts these modelling strategies differently. Using a neural network, practitioners would favour the MICE Miss strategy to achieve the best performance across the different considered groups; whereas the best performances for all groups are achieved for the group-specific Hot Deck when considering decision trees. Together, these results confirm the critical role of imputation choice that more flexible predictive models do not circumvent.

\clearpage
\section{Case study: In-Hospital Mortality Prediction in SUPPORT}
\label{annex:support}
In Section~\ref{sec:mimic}, we study how to inform the choice of imputation strategy when predicting mortality in the ICU using the MIMIC dataset. In this section, we present an alternative application of the framework when modelling in-hospital mortality using the Study to Understand Prognoses Preferences Outcomes and Risks of Treatment (SUPPORT) dataset~\cite{connors1995controlled}. This additional analysis results in a similar imputation card, despite different impacts of imputation on downstream performances, further confirming the need for the proposed framework.

\subsection{Task description.}
The task associated with this controlled study is the prediction of in-hospital mortality given different laboratory tests measured on the third day after admission and indicators of pre-existing conditions. The cohort consists of 9,105 patients admitted to one of five US hospitals.

\subsection{Informing imputation choice.}
\begin{figure}[hp]
    \centering
    \captionsetup{font=small, labelformat=empty}
    \begin{boxedminipage}{\textwidth}
        \centerline{\textbf{Short-term survival prediction}}
        \vspace{5mm}
        \begin{minipage}[t]{0.475\textwidth}
            \paragraph{\textbf{Key Factors}: }
            \begin{itemize}
                \item Demographic groups: Sex (43.8\% female), ethnicity (Black (16.0\%) vs non-Black) and income (71.6\% low).
                \item Data collection setting: All data from 5 teaching hospitals in the USA.
            \end{itemize}
            \paragraph{\textbf{Missingness Process}: }
            \begin{itemize}
                \item Known mechanisms: Standardised procedures following SUPPORT procedures.
                \item Potential influences: Prioritisation of care over observational study requirements.
            \end{itemize}
            \paragraph{\textbf{Descriptive Statistics}: } 
            \begin{itemize}
                \item Range of missingness rate at the end of 24 hours of observation across the different covariates [0.00 - 55.03] with average 11.26\%. The following table presents the results stratified for the different groups.
                \item Percentage of patients with more than 50\% of tests observed: 48.25\%.
            \end{itemize}
            \vspace{5mm}
            \footnotesize
            \centerline{
                \begin{tabular}{c | c | c }
                   Groups & Marginalised & Majority \\ \toprule
                   Sex& 11.14 [0.00 - 57.58]& 11.36 [0.00 - 53.04]\\ 
                   Ethnicity& 10.70 [0.00 - 55.25] & 11.37 [0.00 - 54.99] \\
                   Income& 11.09 [0.00 - 54.92]& 11.71 [0.00 - 55.31]\\ 
                \end{tabular}}
            \captionof*{table}{Missingness percentage (mean [min - max]) stratified per groups. }
        \end{minipage}
        \hfill
        \begin{minipage}[t]{0.475\textwidth}
            \paragraph{\textbf{Methods and Metrics}: }
            \begin{itemize}
                \item Imputation: Mean imputation, Hot Deck, MICE, MICE Missing (using a missingness indicator as input to the model); and their group-specific variants.
                \item Modelling: Logistic regression with l2 penalty on the imputed data.
                \item Metrics: Use of False Negative Rate (FNR) at a 30\% capacity (current threshold of prioritisation) to reflect the percentage of patients that would not be prioritised despite being at risk, both at the population level and stratified by groups.
            \end{itemize}
            \paragraph{\textbf{Empirical Evaluation of Imputation}: } 
            The following figure describes the performance stratified by groups. Overall performance ranges from FNR, highlighting a large impact of imputation on performance.

            {\footnotesize
            \centerline{
                \begin{tabular}{c | c | c }
                   Groups & Gap Range & Best \\ \toprule
                   Sex&  [-12.72 - -3.85]& -3.85\\ 
                   Ethnicity& [-2.49 - 38.14]& -0.66\\ 
                   Income& [-2.08 - 6.09] & -1.32\\  
                \end{tabular}}
            \captionof*{table}{Range FNR performance gaps (in percent) stratified per group.}}
            
            \paragraph{\textbf{Recommendations and Caveats}:} 
            Assuming a stable missingness process and population distribution at deployment, Group MICE with missingness indicator minimises the number of patients missed  across and within each group.
        \end{minipage}
        \vspace{5mm}
        \begin{minipage}{\textwidth}
        \centering
            \includegraphics[height =140px]{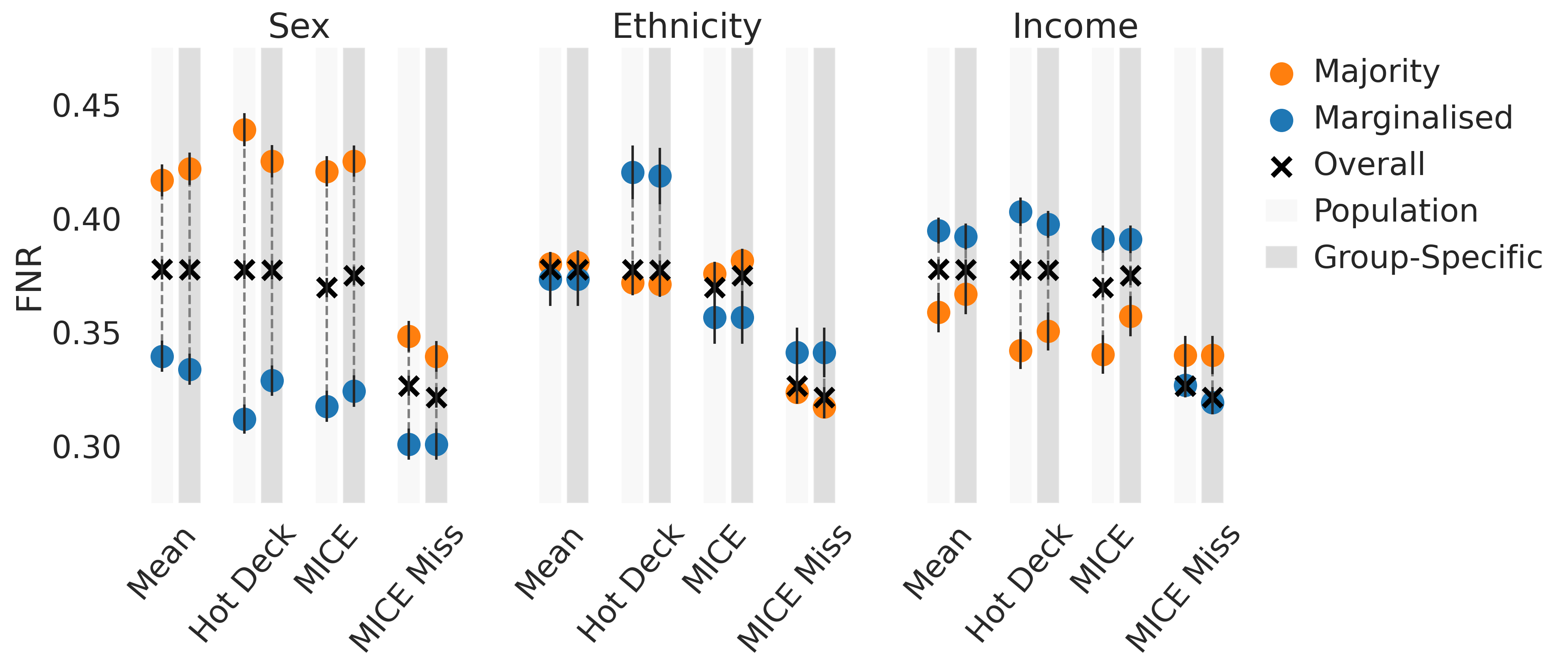}
            \captionof*{figure}{Model performance stratified per group and imputation strategies.}
        \end{minipage}
    \end{boxedminipage}
    \captionsetup{font=normalsize, labelformat=default}
    \caption{Imputation card for in-hospital mortality prediction in the SUPPORT dataset.}
    \label{fig:card:support}
\end{figure}

We apply the framework introduced in Section~\ref{sec:framework} to inform the imputation choice, reported in the following and summarised in Figure~\ref{fig:card:support} using the Imputation Card reporting framework.

\paragraph{Key factors.} Following the algorithmic fairness literature and the available demographic attributes in the SUPPORT dataset, we study groups defined by: ethnicity (Black vs non-Black), sex (female vs male) and income ($\geq$ 25k or less)\footnote{SUPPORT distinguishes 4 income brackets, we groups them into upper and lower income bracket with similar sizes.} as these different groups may interact differently with the healthcare system. Our analysis focuses on patients with observed group membership, resulting in a subset of 6,122 patients.

\paragraph{Missingness process.} The data was collected from five teaching hospitals adhering to standardised medical guidelines. The medical records were meticulously abstracted to ensure consistency across patients and validated to assess agreement between extractions. However, group membership may influence the care provided in hospitals, meaning that the missingness patterns could reflect variations in medical interactions.

\paragraph{Descriptive statistics.}
There is 11.26\% missing data in the considered patient cohort. Missingness is primarily affecting laboratory tests, whereas pre-existing conditions are fully observed. Table~\ref{table:support:missingness} details the percentage of missing laboratory tests disaggregated by the considered groups and considered outcomes. We disaggregate the results between patients who died during their stay and those who survived, finding that patients who died had a significantly higher number of orders compared to those who survived.

When examining missingness across the groups of interest, we observe higher rates of missing data for male, non-Black, and higher-income patients. 
Although the exact cause of missingness rate differences cannot be determined from observational data, these observations highlight the relationship between group membership, missingness, and outcomes. 

\begin{table}[!ht]
    \centering
    \caption{Missingness rate (mean [min - max]) for covariates measured by the third day after admission stratified per groups and outcomes.}
    \label{table:support:missingness}
    \begin{minipage}{.49\textwidth}
        \centering
        \begin{threeparttable}
            \begin{tabular}{rc}
                 & Missingness \\
                \toprule
                Survived: Yes$^+$                & 11.13 [0.00 - 51.60]\\
                Survived: ~No$^+$                 & 11.68 [0.00 - 89.08] \\
                \midrule
                Black                & 10.70 [0.00 - 55.25]\\
                Other            & 11.37 [0.00 - 54.99]\\
            \end{tabular}
            \begin{tablenotes}
              \small
              \item $^+$ In-hospital mortality.
            \end{tablenotes}
        \end{threeparttable}
    \end{minipage}
    \begin{minipage}{.49\textwidth}
        \centering
        \begin{threeparttable}
            \begin{tabular}{rc}
             & Orders\\
            \toprule
            Female               & 11.14 [0.00 - 57.58]\\
            Male                 & 11.36 [0.00 - 53.04]\\
            \midrule
            Low income            & 11.09 [0.00 - 54.92]\\
            High income              & 11.71 [0.00 - 55.31]
            \end{tabular}
            \begin{tablenotes}
              \small
              \item
            \end{tablenotes}
        \end{threeparttable}
    \end{minipage}
\end{table}

\paragraph{Methods and metrics.} 
Similarly to the MIMIC case-study, we consider the same imputation and predictive models as presented in Section~\ref{sec:mimic}, i.e. Mean, Hot-Deck, MICE, MICE Miss and their group variants, followed by a logistic regression modelling.

\begin{figure*}[!ht]
\centering
    \includegraphics[height = 140px]{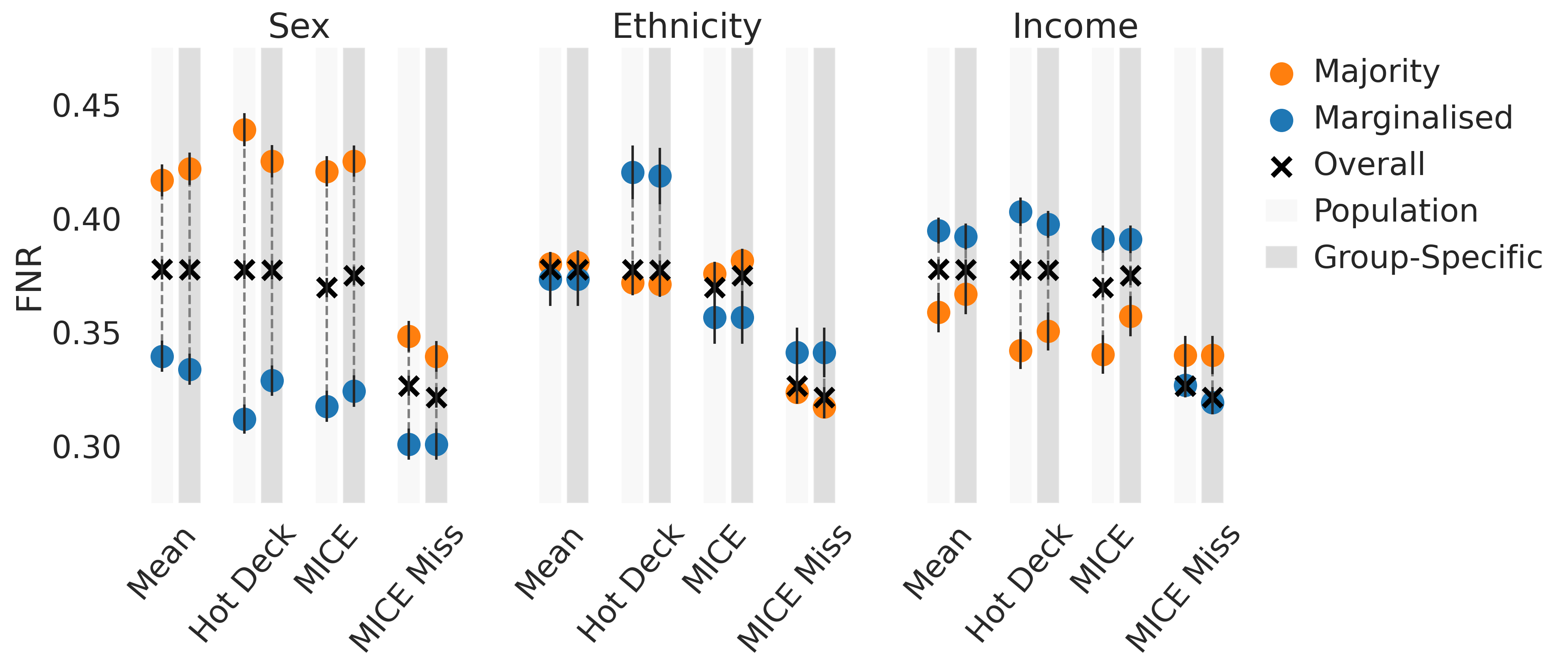}
    \caption{False negative rates across marginalised groups in SUPPORT experiment, bootstrapped on the test set over 100 iterations.}
    \label{fig:support:fnr}
\end{figure*}

\paragraph{Empirical comparison of imputation.} 
Figure~\ref{fig:support:fnr} summarises the impact of each imputation strategy on downstream predictive performance. From this figure, MICE Miss variants present the best overall performance, but the subgroups are impacted differently. Specifically, the smallest performance gaps are achieved by the population variant for both ethnicity and income splits, however, the group variant improves group performance for all subgroups in each split.

\paragraph{Recommendations and caveats.} 
Building on the previous comparison of performance across different imputation strategies, and assuming a similar data-generative and missingness process at deployment, Group MICE Miss appears to benefit all considered groups by minimising the number of critically ill patients who would otherwise not be prioritised. However, note that preferring the group-variant over its population alternative slightly increases the performance gap despite benefiting both groups. In the considered application, maximising subgroup performance may be preferred over reducing the fairness gap.

\end{document}